\documentclass[preprint,12pt]{elsarticle}

\graphicspath{{images/}} 

\usepackage{amssymb}
\usepackage{amsmath}

\usepackage{hyperref}

\usepackage{multirow}

\usepackage{geometry}
\usepackage{longtable}
\usepackage[table]{xcolor}        
\usepackage{subcaption}
\usepackage{threeparttable}
\usepackage{booktabs}       
\usepackage{placeins}
\usepackage{algorithm}
\usepackage{algpseudocode}

\journal{Expert Systems with Applications}

\begin{document}

\begin{frontmatter}



    \title{\textbf{DART}: An Automated End-to-End Object Detection Pipeline with Data \textbf{D}iversification, Open-Vocabulary Bounding Box \textbf{A}nnotation, Pseudo-Label \textbf{R}eview, and Model \textbf{T}raining}


    \author[tum,liebherr]{Chen Xin\corref{ca}}
    \cortext[ca]{Corresponding author}
    \ead{chen.xin@tum.de}

    \author[liebherr]{Andreas Hartel}
    \ead{andreas.hartel@liebherr.com}

    \author[tum]{Enkelejda Kasneci}
    \ead{enkelejda.kasneci@tum.de}

    \affiliation[tum]{organization={Technical University of Munich},
        addressline={Arcisstraße 21},
        city={München},
        postcode={80333},
        country={Germany}}

    \affiliation[liebherr]{organization={Liebherr-Electronics and Drives GmbH},
        addressline={Peter-Dornier-Straße 11},
        city={Lindau (Bodensee)},
        postcode={88131},
        country={Germany}}

    \begin{abstract}
        Accurate real-time object detection is vital across numerous industrial applications, from safety monitoring to quality control. Traditional approaches, however, are hindered by arduous manual annotation and data collection, struggling to adapt to ever-changing environments and novel target objects. To address these limitations, this paper presents DART, an innovative automated end-to-end pipeline that revolutionizes object detection workflows from data collection to model evaluation. It eliminates the need for laborious human labeling and extensive data collection while achieving outstanding accuracy across diverse scenarios. \textbf{DART} encompasses four key stages: (1) Data \textbf{D}iversification using subject-driven image generation (DreamBooth with SDXL), (2) \textbf{A}nnotation via open-vocabulary object detection (Grounding DINO) to generate bounding box and class labels, (3) \textbf{R}eview of generated images and pseudo-labels by large multimodal models (InternVL-1.5 and GPT-4o) to guarantee credibility, and (4) \textbf{T}raining of real-time object detectors (YOLOv8 and YOLOv10) using the verified data. We apply DART to a self-collected dataset of construction machines named Liebherr Product, which contains over 15K high-quality images across 23 categories. The current instantiation of DART significantly increases average precision (AP) from 0.064 to 0.832. Its modular design ensures easy exchangeability and extensibility, allowing for future algorithm upgrades, seamless integration of new object categories, and adaptability to customized environments without manual labeling and additional data collection. The code and dataset are released at \url{https://github.com/chen-xin-94/DART}.
    \end{abstract}




\end{frontmatter}



\section{Introduction}
\label{sec:intro}

\begin{figure}[!h]
    \centering
    \includegraphics[width=1\textwidth]{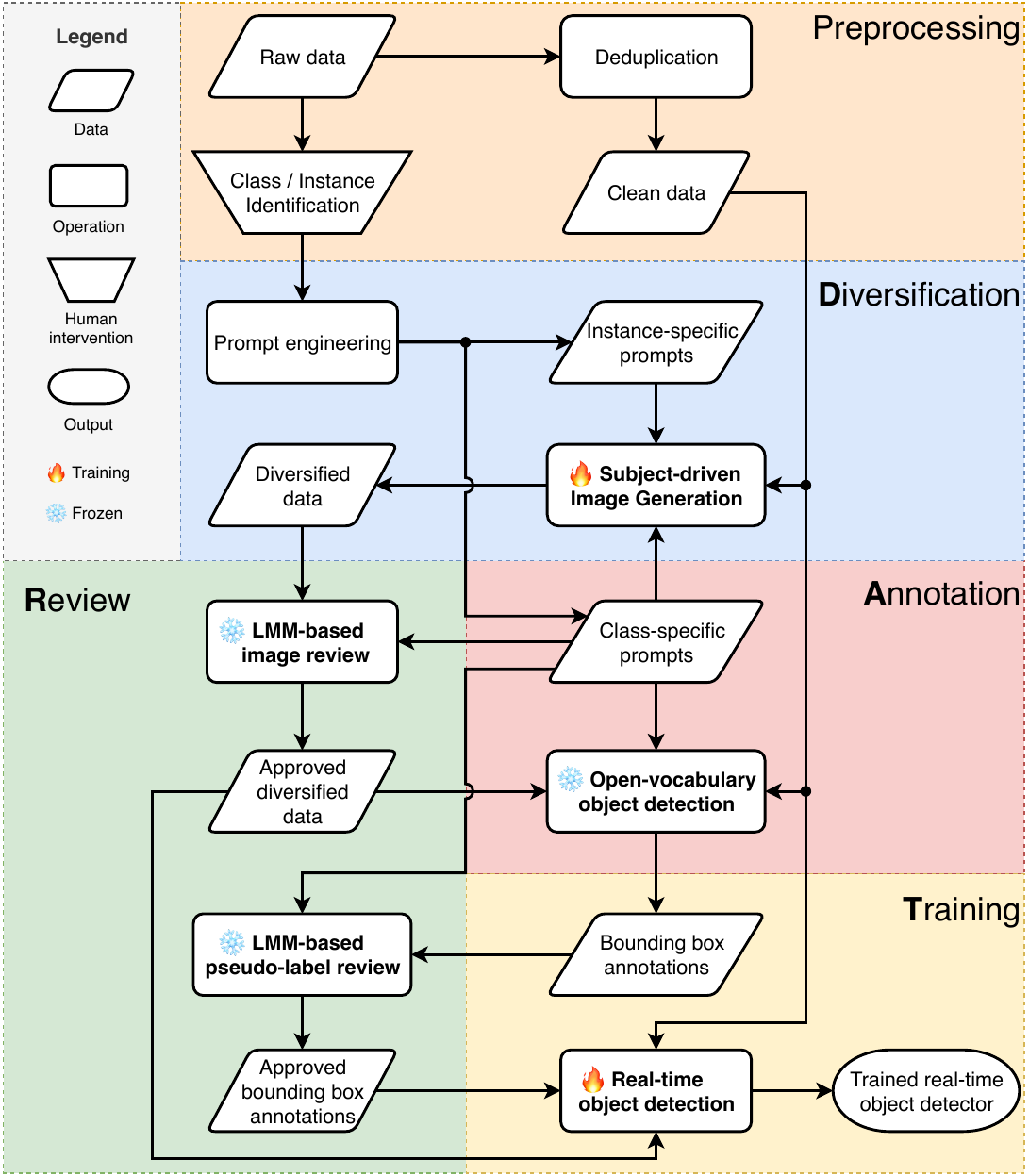}
    \caption{\textbf{DART}, an automated end-to-end pipeline spanning the entire workflow of an object detection application from data collection to model evaluation. After preprocessing, \textbf{DART} enters the data \textbf{D}iversification stage, where a subject-driven image generation module (Dreambooth with SDXL) creates diversified images. The \textbf{A}nnotation stage (red) utilizes both approved diversified data and original clean data to produce bounding boxes through open-vocabulary object detection (Grounding DINO). These pseudo-labels are then \textbf{R}eviewed by another large multimodal model (GPT-4o) and subsequently serve as ground truth for the \textbf{T}raining of real-time object detectors (YOLO).}
    \label{fig:dart}
\end{figure}

In recent years, computer vision has witnessed rapid advancements, revolutionizing various applications through innovative methodologies. A significant development is the emergence of generative AI techniques, exemplified by text-to-image Diffusion models \cite{rombachHighResolutionImageSynthesis2022,rameshHierarchicalTextConditionalImage2022,nicholGLIDEPhotorealisticImage2022,sahariaPhotorealisticTexttoImageDiffusion2022, podellSDXLImprovingLatent2023}. They establish a robust backbone for generating high-fidelity, contextually rich images, which in turn enables other specialized frameworks \cite{mouT2IAdapterLearningAdapters2023, yeIPAdapterTextCompatible2023,galImageWorthOne2022}, including DreamBooth\cite{ruizDreamBoothFineTuning2023}. These frameworks build upon this foundation to enhance the personalization and precision of the generated images, providing more nuanced and versatile controls. Another prominent advancement in generative AI is the large multimodal model (LMM), notably GPT-4 series \cite{openaiGPT4TechnicalReport2023,openaiGPT4o2024}. LMM leverages the synergy between textual and visual data, significantly enhancing the capabilities of comprehensive semantic understanding and interaction in vision-language tasks. Concurrently, traditional computer vision areas like object detection have also experienced expedited breakthroughs. Open-vocabulary object detection (OVD) models like Grounding DINO \cite{liuGroundingDINOMarrying2023} have expanded the scope of object recognition, allowing systems to detect and classify objects beyond the constraints of predefined categories. Meanwhile, the YOLO (You Only Look Once) families \cite{redmonYouOnlyLook2016, redmonYOLO9000BetterFaster2016, redmonYOLOv3IncrementalImprovement2018, bochkovskiyYOLOv4OptimalSpeed2020, YOLOv5Ultralytics2020, liYOLOv6SingleStageObject2022, wangYOLOv7TrainableBagoffreebies2022, liYOLOv6V3FullScale2023, jocherUltralyticsYOLOv82023,wangYOLOv9LearningWhat2024, wangYOLOv10RealTimeEndtoEnd2024} of real-time object detection systems continue to push the Pareto frontier of speed and accuracy, achieving significant advancement through fast and continuous iterations.

Despite the above-mentioned remarkable progress in image generation, LMM,  OVD, and real-time object detection, each technology still has its prominent strengths and weaknesses. While the state-of-the-art image generation models produce high-quality outputs, the results may not always exhibit full photorealism. LMMs excel in the semantic understanding of images but struggle with directly handling numerical data. OVD can detect a wide range of object categories, but it is often too slow. In contrast, real-time object detection is fast but limited to predefined categories. If the strengths of these four technologies can be effectively combined and integrated, it will undoubtedly result in a significantly improved outcome. The proposed DART pipeline (\autoref{fig:dart}) aims to achieve this complementary combination of image generation, LMM,  OVD, and real-time object detection, leveraging their strengths and compensating for their weaknesses to create a comprehensive and efficient object detection solution from data collection to model evaluation.

Ideally, an object detection application should be capable of detecting any target object both quickly and accurately in diverse environments. However, current technologies do not allow for the simultaneous achievement of open-vocabulary detection, real-time capability, high accuracy, and robustness. The most common approach in the industry is to use real-time object detection to meet the essential requirement for speed while manually collecting and annotating as much data as possible to approach the other three goals \cite{xuMultiscaleObjectDetection2024,dongELNetEfficientLightweight2024,gaoPETransformerPathEnhanced2024,choDetectionMovingObjects2023}. Despite the availability of numerous public datasets, they offer limited assistance in this regard because a majority of specific objects are simply not covered by public datasets. For instance, LVIS \cite{guptaLVISDatasetLarge2019}, one of the most diversified object detection datasets, only includes a little more than 1000 entry-level object categories, which is inadequate for many industrial applications. OVD models, on the contrary, are designed to alleviate the object coverage and manual labeling issue. However, they are excessively demanding in terms of computation and memory, making them unsuitable for real-time and on-device applications. On the other hand, the latest image generation tools, such as Stable Diffusion models, are promising alternatives to the inefficient and costly manual collection of more images of the same target object from different scenarios due to their ability to generate diverse images. However, these models lack the ability to mimic the appearance of a specific subject given by visual prompts. Customized frameworks such as DreamBooth could be applied on top for subject-driven image generation instead to address this issue. The generated images can then serve as inputs for OVD to produce bounding boxes, acting as pseudo-labels for training downstream real-time object detection models. However, the results from generative AI can sometimes be overly fanciful, and the quality of bounding boxes produced by OVD can be difficult to guarantee. This necessitates a filtering mechanism to ensure the quality of both generated images and pseudo-labels for the downstream training. This task was previously thought to be exclusive to humans, but now LMMs are well-suited for this task. Although falling short in performing number-related tasks, such as generating bounding box coordinates directly from an input image, they excel in understanding the semantic context of images. As a result, LMM can review the quality of generated images and annotations by solving a designed vision-language task in a Visual Question Answering(VQA) manner. To summarize, the foundational advancements in four areas of computer vision collectively form the backbone of the proposed automated end-to-end object detection pipeline \textbf{DART} to fulfill the four objectives of an ideal object detection application. Specifically, \textbf{DART}  utilizes subject-driven image generation frameworks for data \textbf{D}iversification to enhance robustness, OVD for ground truth bounding box \textbf{A}nnotation to extend object categories without human efforts, LMM for \textbf{R}eviewing generated images and bounding boxes to ensure accuracy and explainability, and \textbf{T}rained real-time object detector as the final output model.

To prove the effectiveness of the DART pipeline for industrial object detection applications and further exploit its potential, we collect a dataset from the internal database of Liebherr, a German-Swiss multinational equipment manufacturer. The collected dataset comprises images of Liebherr products covering a wide variety of heavy machinery such as construction machines, earthmoving equipment, deep foundation machines, mining machines, various types of cranes, material handling machines, and more. We apply dataset-specific preprocessing techniques to the raw images to obtain a clean and refined dataset named Liebherr Products (LP).

In this paper, we instantiate the DART using the current best algorithms for core modules and utilize the pipeline on the LP dataset. Specifically, we use DreamBooth \cite{ruizDreamBoothFineTuning2023} with Stable Diffusion XL (SDXL) \cite{podellSDXLImprovingLatent2023} for data diversification, Grounding DINO \cite{liuGroundingDINOMarrying2023} for bounding box annotation, GPT-4o \cite{openaiGPT4o2024} for pseudo-label review, InternVL-1.5 \cite{chenHowFarAre2024} for generated image review, and finally YOLOv8 \cite{jocherUltralyticsYOLOv82023} and YOLOv10 \cite{wangYOLOv10RealTimeEndtoEnd2024} families as the final object detector to be deployed. This instantiation results in an exceptional real-time object detector capable of accurately identifying and localizing all object categories with an average precision (AP) of 0.832. Without using DART, the result is a disastrous 0.064. The dramatic difference between object detection with and without DART is vividly illustrated in \autoref{fig:predictions_1} and \autoref{fig:predictions_2} through qualitative comparisons. Quantitatively, we also demonstrate the effectiveness of all four major components of DART by showing the continuous improvement in detection performance with the incremental addition of each block. Notably, all four major blocks in the DART pipeline are fully automated, completely eliminating the need for human intervention. This starkly contrasts with traditional supervised training methods that require extensive human efforts in labeling and data collection. Additionally, we design the pipeline to be easily interchangeable and extendable, facilitating the adoption of more advanced algorithms in the future and enabling the seamless integration of new data and object categories with minimal human effort.

In summary, our major contributions are as follows:
\begin{enumerate}
    \item We propose an automated end-to-end object detection pipeline named \textbf{DART} (\autoref{fig:dart}). It covers the entire workflow of an object detection application, from data collection to model evaluation. Specifically, it utilizes subject-driven image generation for data \textbf{D}iversification, open-vocabulary object detection (OVD) for ground truth bounding box \textbf{A}nnotation, large multimodal model (LMM) for \textbf{R}eviewing generated images and pseudo-labels, and \textbf{T}rained real-time object detector as the final output.
    \item We instantiate the core module (bold text in \autoref{fig:dart}) of the four major stages in the DART pipeline with the current best options available. Specifically, we
          \begin{itemize}
              \item customize DreamBooth with Stable Diffusion XL (SDXL) to generate images of the same instances but with different postures and backgrounds for data diversification,
              \item apply Grounding DINO to both collected and generated images for bounding box annotation as pseudo-labels,
              \item adopt GPT-4o and InternVL-1.5 to review the quality of pseudo-labels and photorealism of generated images to determine their suitability as training data.
              \item and finally train models from YOLOv8 and YOLOv10 families as the final real-time object detector.
          \end{itemize}
    \item We collect a dataset, named Liebherr Products (LP), of more than 15K images covering various heavy machinery from Liebherr products. We apply the instantiated DART pipeline on the LP dataset and obtain a highly capable YOLOv8n that can accurately identify and localize all object categories with an AP of 0.832 and a latency of 2.47 ms on a T4 GPU without any effort of human labeling. In contrast, an AP of only 0.064 is achieved without the DART pipeline.
    \item The four key phases in the DART pipeline are fully automated, eliminating the need for human intervention, especially in annotation and extra data collection. Additionally, we employ a modular design in DART to ensure easy exchangeability and extensibility. This facilitates a smooth transition to more advanced future algorithms, seamless integration of new object categories, and adaptability to environments with customized requirements with minimal human intervention.
    \item We conduct numerous experiments to demonstrate the effectiveness of DART by showing the continuous improvement in detection performance with the incremental addition of each of the four major components and finding the optimal tradeoff within each block.
\end{enumerate}

\section{Related work}
\label{sec:related}

In the related work section, we will focus on four areas corresponding to the DART pipeline's core stages in order: subject-driven image generation, open-vocabulary object detection, large multimodal models, and real-time object detection. By discussing progress in these fields, we aim to justify our current implementation of state-of-the-art methods and explore the potential for further enhancements by incorporating more advanced modules in the future.

\subsection{Subject-driven image generation}
\label{sec:r-gen}
Text-to-image diffusion models \cite{rombachHighResolutionImageSynthesis2022,rameshHierarchicalTextConditionalImage2022,nicholGLIDEPhotorealisticImage2022,sahariaPhotorealisticTexttoImageDiffusion2022, podellSDXLImprovingLatent2023} have swiftly advanced in recent years, establishing themselves as powerful tools in generative AI. These models generate high-quality images from textual descriptions through iterative denoising processes, forming the foundation for subject-driven text-to-image generation. Given only a few (typically 3-5) casually captured images of a specific subject, subject-driven image generation aims to produce customized images with high detail fidelity and variations guided by text prompts. One typical approach is to fine-tune a special prompt token to describe the concept of the specific subject. Textual Inversion \cite{galImageWorthOne2022} shows evidence that a single-word embedding is sufficient for capturing unique concepts. Recent methods introduce additional modules to the subject-driven image generation framework and only update the newly added modules during training while keeping the pre-trained text-to-image frozen. T2I-Adapter \cite{mouT2IAdapterLearningAdapters2023} aligns pre-trained T2I models with external control signals extracted from trainable CNNs to enhance controllability in text-to-image generation without greatly affecting the original generation ability. IP-Adapter \cite{yeIPAdapterTextCompatible2023} decouples the cross-attention mechanism by separating the cross-attention layers for text features and image features to achieve image prompt capability for the pre-trained text-to-image diffusion models. InstantID \cite{wangInstantIDZeroshotIdentityPreserving2024} follows the path of decoupled cross-attention and leverages a pre-trained face model to detect and extract face ID embedding from the reference facial image to generate ID-preserving images. Although these fine-tuning methods are quick and simple, the generated images often fail to completely resemble the given subject. Therefore, we adopt a full fine-tuning framework, DreamBooth, which directly fine-tunes both the text-to-image backbone with a few images of a specific subject and the token embedding, enabling the generation of diverse high-fidelity images of the specific subjects. To maximize the diversity and fidelity of the generations, we select the powerful SDXL as the backbone for the DreamBooth framework. We also improve the training process to maximize the potential of Dreambooth with SDXL and design diversified prompts to generate new images with diversified contexts.

\subsection{Open-vocabulary object detection}
\label{sec:r-ovd}
Open-vocabulary object detection (OVD) has become a cutting-edge approach in modern object detection. It aims to detect any objects given their corresponding text prompts beyond predefined categories. Recent groundbreaking works in OVD include ViLD \cite{guOpenvocabularyObjectDetection2022} which distills CLIP \cite{radfordLearningTransferableVisual2021} embeddings to cropped image embeddings from region proposal network (RPN) \cite{renFasterRCNNRealTime2016}, as well as GLIP \cite{liGroundedLanguageImagePretraining2022} where object detection is reformulated as a grounding task by aligning bounding boxes to phrases using early fusion. Moreover, Grounding DINO \cite{liuGroundingDINOMarrying2023} extends the idea of cross-modality fusion to both feature extraction and decoder layers. It excels in zero-shot scenarios, demonstrating an impressive generalization ability to detect unseen objects. Nevertheless, one major issue of OVDs is that they are too computation- and memory-intensive to meet the needs of real-time and on-device applications. This perspective holds true even for the latest lightweight OVDs such as YOLO-world \cite{chengYOLOWorldRealTimeOpenVocabulary2024}. Therefore, instead of directly using OVD to solve our tasks, the proposed DART pipeline leverages OVD to generate ground truth as training data for downstream real-time object detectors. Combined with the LMM-based bounding box review (see \autoref{sec:r-lmm} and \autoref{sec:lmm}), our approach results in a faster downstream detector and alleviates the speed limitations inherent in OVD.

\subsection{Large multimodal model}
\label{sec:r-lmm}
Recent advancements in large multimodal models (LMMs) have been characterized by the release of several groundbreaking proprietary systems, including GPT-4v \cite{openaiGPT4TechnicalReport2023}, Claude 3 \cite{anthropicClaudeFamily2024}, and Gemini \cite{googleGeminiFamilyHighly2024}. These models have collectively set a new standard for vision-language understanding, demonstrating unprecedented capabilities in combining visual inputs with advanced language processing. Recently, GPT-4o \cite{openaiGPT4o2024} further extends these capabilities with enhanced contextual awareness and multimodal reasoning, pushing the boundaries of LMM by achieving state-of-the-art performance across a wide range of vision-language tasks. Meanwhile, the open-source community has also been actively advancing in this domain. Flamingo \cite{alayracFlamingoVisualLanguage2022} utilizes visual and language inputs as prompts, demonstrating exceptional few-shot performance in VQA. LLaVA series \cite{liuVisualInstructionTuning2023, liuImprovedBaselinesVisual2024, xuLLaVAUHDLMMPerceiving2024} and MiniGPT-4 \cite{zhuMiniGPT4EnhancingVisionLanguage2023} introduce visual instruction tuning to improve the instruction-following ability of LMMs. InternVL family \cite{chenInternVLScalingVision2024,chenHowFarAre2024} scales up vision foundation models and trains on a curated high-quality dataset to achieve comparable results to GPT-4v among various multimodal benchmarks. Although open-source models have proven adequate for basic semantic understanding tasks such as image review \cite{liLLaVAMedTrainingLarge2023, leeNERIFGPT4vAutomatic2023}, proprietary models, especially GPT-4o, still demonstrate superior performance in handling intricate visual data and sophisticated multimodal interactions, especially when additional visual hints are applied \cite{yangSetofMarkPromptingUnleashes2023}. However, they both struggle with tasks requiring numerical outputs, such as directly generating bounding box coordinates \cite{yangDawnLMMsPreliminary2023}. Therefore, our DART pipeline only leverages GPT-4o/InternVL-1.5 to judge whether a model-generated output is suitable as pseudo-labels/input data for the downstream real-time object detection training, while leaving the specialized task of bounding box annotation to OVD. Additionally, to better handle the review task, we craft specialized prompts tailored to the characteristics of the two LMMs, resulting in excellent outcomes.

\subsection{Real-time object detection}
\label{sec:r-od}
Real-time object detection aims to classify and locate objects in real-time with low latency, which is essential for numerous industrial applications.  This area has witnessed significant advancements in recent years, primarily driven by the iterative development of state-of-the-art YOLO models. Early development of YOLO from v1 to v3 \cite{redmonYouOnlyLook2016, redmonYOLO9000BetterFaster2016, redmonYOLOv3IncrementalImprovement2018} determined the optimal segmentation of the real-time object detection network structure: backbone, neck, and head. YOLOv4 \cite{bochkovskiyYOLOv4OptimalSpeed2020} and YOLOv5 \cite{YOLOv5Ultralytics2020} changed the backbone to CSPNet \cite{wangCSPNetNewBackbone2019} and introduced a series of best practices in computer vision regarding training strategy, post-processing, as well as plugin modules. Based on these improvements, YOLOv6 \cite{liYOLOv6V3FullScale2023} and YOLOv7 \cite{wangYOLOv7TrainableBagoffreebies2022} are further fine-tuned in label assignment strategy\cite{fengTOODTaskalignedOnestage2021} and loss functions \cite{zhengDistanceIoULossFaster2019, gevorgyanSIoULossMore2022,rezatofighiGeneralizedIntersectionUnion2019}, and they utilize Repblock \cite{dingRepVGGMakingVGGstyle2021} and E-ELAN \cite{wangDesigningNetworkDesign2022} as the core of their backbone respectively. YOLOv8 \cite{jocherUltralyticsYOLOv82023} introduces the C2f building block for effective feature extraction and fusion, along with a comprehensive set of integrated tools, solidifying its position as the go-to choice for real-time object detection. YOLOv9 \cite{wangYOLOv9LearningWhat2024} proposes programmable gradient information (PGI) to refine the training process and a new lightweight network architecture – Generalized Efficient Layer Aggregation Network (GELAN). Finally, the concurrent work YOLOv10 \cite{wangYOLOv10RealTimeEndtoEnd2024} presents an NMS-free training strategy and polishes the model architecture with a holistic efficiency-accuracy-driven design, contributing to its state-of-the-art real-time object detection capabilities. Considering YOLOv8's ease of use and YOLOv10's optimal tradeoff between speed and accuracy, this work employs both the YOLOv8 and YOLOv10 families as the real-time object detectors of choice. Although we have selected the best YOLO models, their reliance on predefined object categories restricts their applicability in open scenarios. Therefore, real-time object detection only serves as the last stage in DART, depending on the other three stages to provide high-quality training data.

\section{Methods}
\label{sec:methods}

The proposed method, DART, is an automated end-to-end object detection pipeline that spans the entire workflow of an object detection application from data collection to model evaluation. As illustrated in \autoref{fig:dart}, \textbf{DART} starts with dataset preprocessing, followed by four main stages: data \textbf{D}iversification through subject-driven image generation, bounding box \textbf{A}nnotation via OVD, LMM-based \textbf{R}eview of generated images and pseudo-labels, as well as real-time object detector \textbf{T}raining. The core module (bold text in \autoref{fig:dart}) of each phase is instantiated with state-of-the-art techniques, including the framework DreamBooth with SDXL for subject-driven image generation, the application of Grounding DINO for OVD, the adoption of GPT-4o and InternVL-1.5 for generation quality review, and the training of YOLOv8 and YOLOv10 models for real-time object detection. The following sections provide a detailed breakdown of each component of the proposed DART pipeline.

\subsection{Dataset preprocessing}
\label{sec:preprocessing}

The data preprocessing phase, illustrated by the orange block in \autoref{fig:dart}, involves identifying classes and instances from raw data as input for prompt engineering and deduplication to obtain clean data.

\textbf{Class identification.} The raw dataset is sourced from Liebherr’s internal digital system, where images of Liebherr's products are initially roughly categorized in the database. By refining these preliminary classifications with detailed product descriptions, we establish 23 accurately defined classes, as detailed in \autoref{sec:dataset}. We find the precise definition of the class name essential because it highly affects the performance of the following data diversification phase, since an ambiguous class definition leads to inferior quality of the generated data of that class. As a result, each image in the raw dataset is assigned at least one precise class name based on its primary object, and this broad assignment will be used as one of the text prompts for bounding box annotation, as described in \autoref{sec:gdino}.

\textbf{Instance identification.} For images containing a single object, we tag them at the instance level using distinctive metadata information, such as product numbers. For each class, we randomly selected 2-5 instances and manually curated approximately 10 images for each instance (see \autoref{fig:instance}). These images will be further utilized for data diversification (\autoref{sec:dreambooth}). Note that tagging with instance labels is a straightforward and easy task since we only need 20-50 images in total for each class.

\textbf{Prompt engineering.} Class names for images are the cornerstone for prompt construction for image generation (\autoref{sec:dreambooth}), OVD (\autoref{sec:gdino}), and LMM-based review (\autoref{sec:lmm}). The constructed class-specific prompts vary slightly for different tasks. Instance tags, however, will only serve as the backbone for instance-specific prompts for training SDXL models under the DreamBooth framework. Specific prompt construction rules will be introduced in the corresponding sections mentioned above and further detailed in \autoref{app:prompt}. The precise text prompts for data diversification, annotation, and review stage are listed in \autoref{tab:prompts_dreambooth}, \autoref{tab:prompt_gpt}, and \autoref{tab:prompt_internvl}, respectively.

\textbf{Deduplication.} The initial raw dataset contains many duplicated images. We employ perceptual hashing (pHash) \cite{klingerPHashOrgHome2008} for preliminary deduplication. However, pHash tends to misidentify similar images as duplicates, for example, consecutive shots of the same scene from slightly different angles. These near duplicates are manually retained in the dataset but are ensured to appear only in the training set.

Upon completing these steps, we obtain a clean dataset, dubbed Liebherr Products (LP). We also acquire instance-specific and class-specific prompts for the following phases. It should be noted that all human interventions within the DART framework conclude here, and the subsequent four phases of DART are fully automated.

\subsection{Data diversification based on DreamBooth with Stable Diffusion XL}
\label{sec:dreambooth}

\begin{figure}[!h]
    \centering
    \includegraphics[width=1\textwidth]{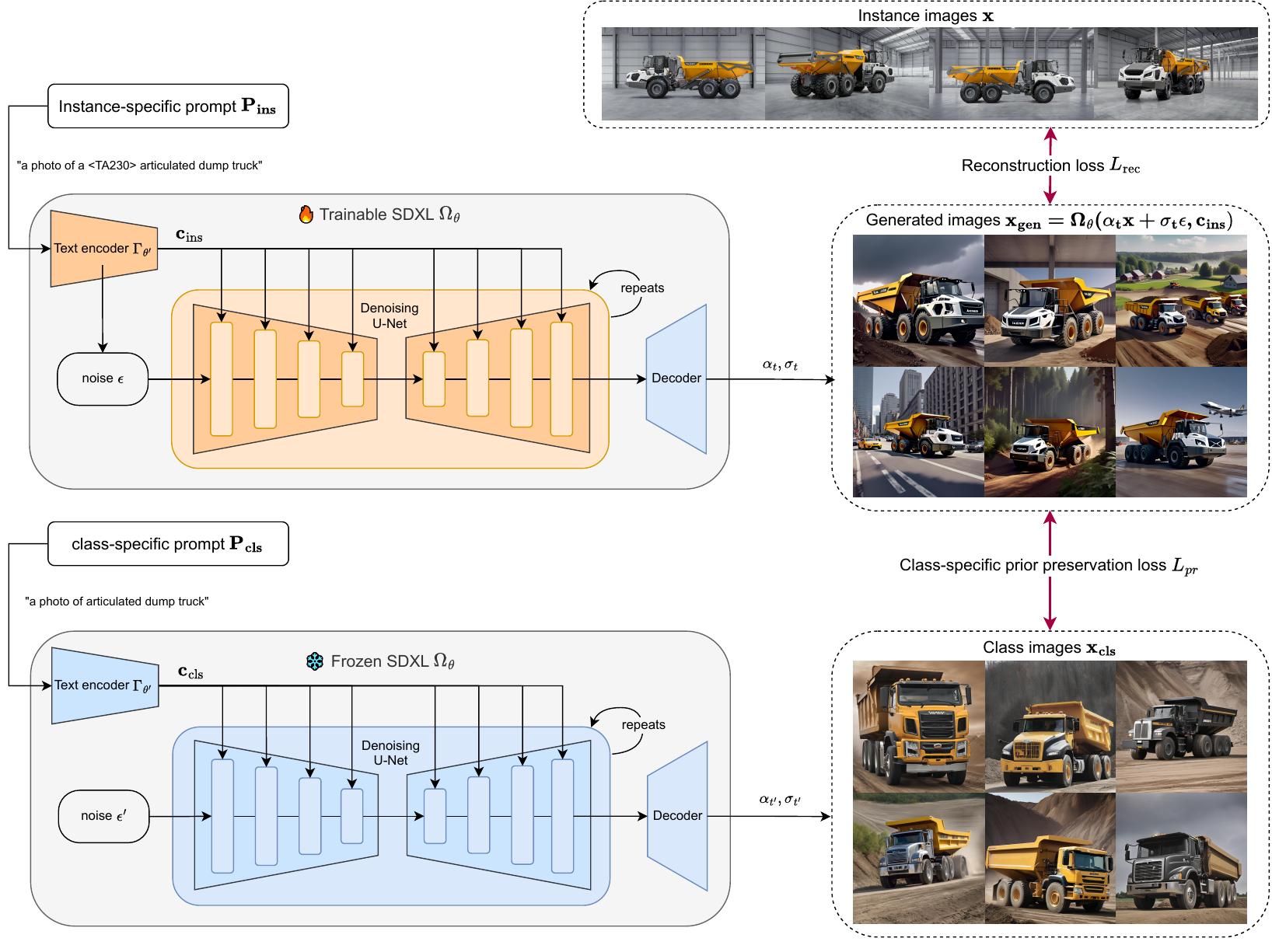}
    \caption{DreamBooth with SDXL. Modified from \cite{ruizDreamBoothFineTuning2023}.}
    \label{fig:dreambooth}
\end{figure}

Data diversification and augmentation in DART are deeply rooted in the latest image-generation techniques. One of the most popular ones is Stable Diffusion, which represents a series of latent diffusion models that iteratively denoise random Gaussian noise in the latent space of a pre-trained autoencoder using a U-Net architecture \cite{ronnebergerUNetConvolutionalNetworks2015} conditioned on textual prompts. Stable Diffusion XL (SDXL) introduces several architectural improvements over previous versions of Stable Diffusion, including a larger U-Net backbone, more powerful text encoders, and novel conditioning schemes to enhance the visual fidelity of generated samples, resulting in significantly improved performance. However, SDXL takes only text prompts and, therefore, cannot mimic the appearance of a specific subject given by visual prompts, which is essentially our goal for the data diversification stage. DreamBooth addresses this issue by fine-tuning a pre-trained SDXL model using a few images of a specific subject as ground truth. The key idea is to bind a unique textual identifier with the visual subject, allowing the model to synthesize novel photorealistic images of the subject in diverse scenes and poses with the help of diversified text prompts. The framework of DreamBooth with SDXL is illustrated in \autoref{fig:dreambooth}. Without loss of generality, we showcase the class "articulated dump truck" and one of its instances, "TA230".

During training, a pre-trained SDXL model $\Omega_\mathbf{\theta}$ with frozen weights $\mathbf{\theta}$ is first applied to generate a bunch of class images $\mathbf{x_{\text{cls}}}=\Omega_\mathbf{\theta}(\mathbf{\epsilon'},\mathbf{c_{\text{cls}}})$, given initial noises $\mathbf{\epsilon'} \sim \mathcal{N}(0, I)$ and a conditioning vector $\mathbf{c_{\text{cls}}}=\Gamma_{\mathbf{\theta'}}(\mathbf{P_{\text{cls}}})$ generated by a text encoder $\Gamma$ with frozen weights $\mathbf{\theta'}$ using a class-specific text prompt $\mathbf{P_{\text{cls}}}$, which takes the form "a photo of a \{class\_name\}".

Then, the same SDXL model is fine-tuned by setting $\mathbf{\theta}$ and $\mathbf{\theta'}$ to be trainable. In practice, we fine-tune all layers of the SDXL model, including the text encoder, using LoRA (Low-Rank Adaptation) \cite{huLoRALowRankAdaptation2021}. LoRA freezes the pre-trained weights and inserts trainable low-rank matrices into the transformer layers’ weight matrices, enabling efficient adaptation with minimal additional parameters. Here, we slightly abuse the notation of $\mathbf{\theta}$ and $\mathbf{\theta'}$ to also represent LoRA parameters in what follows.

The loss of the fine-tuning process is composed of two parts. The first part is the standard reconstruction loss $L_{\text{rec}}$ for diffusion model training. Specifically, it's a squared error (SE) loss between a ground truth instance image $\mathbf{x}$ and a denoised version of it, predicted by the model $\Omega_\mathbf{\theta}$ at a specific diffusion timestep $t \sim \mathcal{U}[0, 1]$:
\begin{equation}
    L_{\text{rec}} = w_t \left(\Omega_\mathbf{\theta}(\alpha_t \mathbf{x} + \sigma_t \mathbf{\epsilon}, \mathbf{c}_{\text{ins}}) - \mathbf{x}\right)^2
    \label{eq:loss_rec}
\end{equation}
where $\alpha_t$, and $\sigma_t$ are functions of timestep $t$ that control the noise schedule of the diffusion process. As for $w_t$ in \autoref{eq:loss_rec}, we adopt the Min-SNR weighting strategy \cite{hangEfficientDiffusionTraining2024} to avoid the inefficiency and instability caused by conflicts among timesteps. Specifically, $w_t=\min(\alpha_t^2/\sigma_t^2,\gamma)$, where $\gamma$ is a hyperparameter to avoid the model focusing too much on small noise levels. Note that the encoded instance-specific prompt $\mathbf{c}_{\text{ins}}$ is used for the prediction of the ground truth instance. In contrast to class-specific prompts $\mathbf{P_{\text{cls}}}$ that only provide a general description of a class, instance-specific prompts $\mathbf{P_{\text{ins}}}$ should include both a general class description for better image quality and a unique identifier to build the correspondence to the given visual instance. Therefore,  $\mathbf{P_{\text{ins}}}$ utilizes the extracted instance name (usually the product number) wrapped in angle brackets as the identifier (e.g., "<TA230>" in \autoref{fig:dreambooth}) followed by the original class name.

The second component of the overall loss is the class-specific prior preservation loss $L_{pr}$, which is essentially the same SE loss but treats the class images $\mathbf{x_{\text{cls}}}$ as ground truth instead. It is introduced to mitigate the overfitting issue of the standard fine-tuning. With both reconstruction loss $L_{\text{rec}}$ and class-specific prior preservation loss $L_{pr}$, DreamBooth encourages diversity in the generated images while preserving the key visual features of the subject. The total loss $L$ can be expressed as follows:
\begin{equation}
    \begin{split}
        L &= L_{\text{rec}} + \lambda L_{pr} \\
        &= w_t \left(\Omega_\mathbf{\theta}(\alpha_t \mathbf{x} + \sigma_t \mathbf{\epsilon}, \mathbf{c}_{\text{ins}}) - \mathbf{x}\right)^2 \\
        &+ \lambda w_{t'} \left(\Omega_\mathbf{\theta}(\alpha_{t'} \mathbf{x_{\text{cls}}} + \sigma_{t'} \mathbf{\epsilon'}, \mathbf{c}_{\text{cls}}) - \mathbf{x_{\text{cls}}}\right)^2
    \end{split}
    \label{eq:loss}
\end{equation}
We fine-tune a specific SDXL model with the DreamBooth framework for each instance. Diversified text prompts can then guide the trained SDXL models to depict each specific instance in novel contexts. We carefully design 67 text prompts in total that cover a wide and diverse range of scenarios, weather, poses, and quantities. Please refer to \autoref{tab:prompts_dreambooth} for the list of prompts used during inference for data generation. Note that to achieve our goal of diversification, we sampled a wide range of prompts covering various scenarios. As we will see in \autoref{sec:vis} and \autoref{app:vis}, DART's data diversification stage effectively follows the scenarios and requirements mentioned in the text, demonstrating excellent adaptability to customized environments with minimal human intervention. At the end of the data diversification block, we obtain diversified data ready to be annotated.

\subsection{Open-vocabulary bounding box annotation via GroundingDINO}
\label{sec:gdino}

Grounding DINO is an open-vocabulary object detector that marries a transformer-based detector DINO \cite{podellSDXLImprovingLatent2023} with grounded pre-training, which can draw bounding boxes for arbitrary objects given inputs as plain text. The key to Grounding DINO's open-set capability is the tight fusion of language and vision modalities, achieved through three main components: a feature enhancer, a language-guided query selection, and a cross-modality decoder. The feature enhancer fuses language and visual features by interleaving text-to-image and image-to-text cross-attention based on features extracted from text and image backbones in the early stage. The language-guided query selection uses language information (text features) to guide the selection of detection queries from image features, and the cross-modality decoder extends standard cross-attention to two layers using image and text features as keys and values right before model outputs.

In this study, we leverage Grounding DINO for the bounding box annotation of target classes. An intuitive approach is to concatenate class names with delimiters ('.' for Grounding DINO) into one text prompt as input to the model. However, we find that the model performance decreases exponentially as the length of the text grows. Thus, we turn to a strategy of incrementally adding and replacing text prompts to ensure that we ensure higher scores for ground truth accuracy.

\begin{algorithm}[ht!]
\caption{Bounding Box Annotation}\label{alg:gdino}
\begin{algorithmic}
\Require Pre-trained Grounding DINO model $\mathcal{G}$ which takes given text prompt $\mathcal{T}$ and image $\mathcal{I}$ to generate bounding boxes $\mathcal{B}$ with confidence scores $\mathcal{S}$ and corresponding phrases $\mathcal{R}$, i.e. $\mathcal{G}: \mathcal{T}, \mathcal{I} \to \mathcal{B}, \mathcal{S}, \mathcal{R}$, 
\Require A dataset of $N$ images $\mathcal{D} = \{\mathcal{I}_n\}_{n=1}^{N}$, a mapping of each image to its pre-defined class $\mathcal{M}_{\text{cls}}: \mathcal{I} \to \mathcal{C}$, a mapping of each class to its list of synonyms $\mathcal{M}_{\text{syn}}: \mathcal{C} \to \mathcal{L}_{\text{syn}}$, and a mapping of each class to its list of common co-occurring classes $\mathcal{M}_{\text{co}}: \mathcal{C} \to \mathcal{L}_{\text{co}}$, a prompt construction rule that takes class names and provides specific text pormpt $\mathcal{P}:\mathcal{C} \to \mathcal{T}$
\Ensure Bounding box annotations $\mathcal{B}$ for all images in $\mathcal{D}$
\State$\mathcal{B} \gets [\ ]$
\For{$I\in \mathcal{D}$} 
    \State$B_I\gets [\ ]$ \Comment{a list to store temporary annotation results for Image $I$}
    \State$c \gets \mathcal{M}_{\text{cls}}(I)$ \Comment{extract pre-assigned class name $c$}
    \State$L_{\text{syn}} \gets \mathcal{M}_{\text{syn}}(c)$ \Comment{extract the list of synonyms for class $c$}
    \State$L_{\text{co}} \gets \mathcal{M}_{\text{co}}(c)$ \Comment{extract the list of co-occurring classes for class $c$}
    \State\textbf{Bounding box generation for the original prompt}
    \State$t \gets \mathcal{P}(c)$ \Comment{construct original prompt}
    \State$(b,s,r) \gets \mathcal{G}(t,I)$  \Comment{apply Grounding DINO}
    \State$B_I\gets B_I\cup (b,s,r)$ \Comment{Append results}
    \State\textbf{Bounding box generation for the co-occurring prompt}
    \If{$L_{\text{co}} \neq \emptyset$}  \Comment{if there're co-occurring classes for class $c$}
        \State$t_{\text{co}}  \gets \mathcal{P}(L_{\text{co}},c)$ \Comment{construct co-occurring prompt using original class names}
        \State$(b,s,r)\gets \mathcal{G}(t_{\text{co}},I)$  \Comment{apply Grounding DINO}
        \State$B_I\gets B_I\cup (b,s,r)$ \Comment{Append results}
    \EndIf
    \State\textbf{Bounding box generation for synonym prompts}
    \If{$L_{\text{syn}} \neq \emptyset$}  \Comment{if there're synonyms}
    \For{$c_{\text{syn}} \in L_{\text{syn}}$} \Comment{Loop through all synonyms for class $c$}
        \State$t_{\text{syn}}  \gets \mathcal{P}(c_{\text{syn}} )$ \Comment{construct synonym prompt}
        \State$(b,s,r) \leftarrow \mathcal{G}(t_{\text{syn}},I)$ \Comment{apply Grounding DINO }
        \State$B_I\leftarrow B_I\cup (b,s,r)$ \Comment{Append results}
        \If{$L_{\text{co}} \neq [\ ]$}  \Comment{if there're co-occurring classes for class $c$}
            \State$t'_{\text{co}}  \gets \mathcal{P}(L_{\text{co}},c_{\text{syn}})$ \Comment{construct co-occurring prompt using the synoynm}
            \State$(b,s,r) \gets \mathcal{G}(t'_{\text{co}}, I)$  \Comment{apply Grounding DINO}
            \State$B_I\gets B_I\cup (b,s,r)$ \Comment{Append results}
        \EndIf
    \EndFor
    \EndIf
    \State$\mathcal{B} \gets \mathcal{B} \cup B_I$ \Comment{Append all annotation results for Image $I$}
\EndFor
\State\Return$\mathcal{B}$
\end{algorithmic}
\end{algorithm}

As described in \autoref{alg:gdino}, we start by using the class name that is roughly assigned during data collection (see \autoref{sec:preprocessing}) as the original prompt. To avoid misunderstanding caused by unclear semantics of the text, we design the synonym prompt using synonyms for the categories with ambiguous meanings (e.g., a handler can mean either the machine or the human operator), common aliases (e.g., dozer for bulldozer) or explicit superclasses (e.g., crane for crawler crane). A list of all synonyms can be found in \autoref{tab:synonyms}. Note that we input one synonym at a time to the model to generate bounding boxes for all synonym prompts to trade redundancy for high accuracy. The class-agnostic NMS (the last step of \autoref{alg:nms}) will finally address the redundant annotations.

To address the common phenomenon of multiple object categories in a single image and avoid the subpar performance of concatenating all classes as text prompts, we utilize the information (collected during data preprocessing) on frequently co-occurring object combinations to create another type of prompt dubbed "co-occurring prompt". Unlike the original and synonym prompts, where one phrase is presented at a time, the co-occurring prompt concatenates the names of objects commonly appearing together with delimiters into a single text. For instance, three classes of the LP dataset, "mining trucks," "mining excavators," and "mining bulldozers," often appear simultaneously but rarely with other classes. Thus, images categorized into these three classes during preprocessing are given a co-occurring prompt that combines the names of the corresponding classes. This construction rule is necessary because Grounding DINO tends to produce identical boxes but different class labels when the separately given text prompt is semantically similar, which is often the case for co-occurring classes. This issue is significantly alleviated when these semantically similar texts are input into Grounding DINO as one text prompt. However, the ultimate solution to this problem is the class-agnostic NMS, which will be discussed in the next paragraph. Additionally, we apply the same rule for constructing co-occurring prompts for classes with synonyms, iterating through all combinations of these synonyms.

\begin{algorithm}[h!]
\caption{Filtering and NMS}\label{alg:nms}
\begin{algorithmic}
\Require A dataset of $N$ images $\mathcal{D} = \{\mathcal{I}_n\}_{n=1}^{N}$, the non-maximum-suppression (NMS) function $\mathcal{F}$, bounding box annotations results consisting of generated box coordinates $b$, the confidence score $s$, and the corresponding phrase $r$ stored as a 3-tuple for all $M$ annotations returned from \autoref{alg:gdino}, i.e. $ \mathcal{B}=\{(b,s,r)_m\}_{m=1}^{M}$
\Ensure Final annotations $\mathcal{B}_{\text{final}}$ for all images in $\mathcal{D}$
\State$\mathcal{B}_{\text{final}} \gets [\ ]$
\For{$I\in \mathcal{D}$} 
    \State$B_I^{\text{raw}} \gets$ a list of each annotation tuple $(b,s,r)$ in $\mathcal{B}$ that belongs to Image $I$
    \State$B_I \gets$ a list of each annotation tuple in $B_I^{\text{raw}}$ but with its original class name $(b,s,c)$ 
    \State\textbf{Filtering}
    \State$B_{\text{filtered}} \gets [\ ]$
    \If{$length(B_I)==1$} \Comment{for images with single annotation}
        \State$B_{\text{filtered}} \gets B_{\text{filtered}} \cup B_I$ \Comment{take the single annotation}
    \Else       
        \For{$(b,s,r) \in B_I$}   \Comment{for images with multiple annotations}
            \If{$s \geq 0.5 $} \Comment{extract only annotations with score $\geq$ 0.5}
                \State$B_{\text{filtered}} \gets B_{\text{filtered}} \cup (b,s,r)$ \Comment{Append results}
            \EndIf
        \EndFor
    \EndIf
    \State\textbf{Non-Maximum Suppression (NMS)}
    \State$B, S \gets$ all $b$ and $s$ in $B_{filtered}$ \Comment{extract boxes with scores}
    \State$B_{\text{nms}} \gets \mathcal{F}(B,S)$    \Comment{apply NMS}
    \State$\mathcal{B}_{\text{final}} \gets \mathcal{B}_{\text{final}} \cup B_{\text{nms}}$ \Comment{Append final annotation results for Image $I$}
\EndFor
\State\Return$\mathcal{B}_{\text{final}}$
\end{algorithmic}
\end{algorithm}

After getting the output bounding boxes and their corresponding confidence scores from the original, synonym, and co-occurring prompts for all images, we apply filtering and non-maximum suppression (NMS) to keep only the best bounding boxes. As illustrated in \autoref{alg:nms}, we first sift out the ones with confidence scores bigger than 0.5 unless there's only one box for an image, in which case we take the one regardless. Then, we transform labels derived from various text prompts back to their original class names from synonyms and finally adopt class-agnostic NMS to address the issue of confusing similar labels and keep only the most feasible annotations. The remaining bounding boxes with their given class labels and scores are the candidate pseudo-labels to be reviewed.

\subsection{LMM-based review of pseudo-labels and image photorealism using InternVL-1.5 and GPT-4o}
\label{sec:lmm}

LMM leverages the synergy between textual and visual data to perform tasks that require cross-modality comprehension, such as visual question answering (VQA). In the VQA setup, questions with hints and requirements are first summarized as text prompts and then further processed by a (preferably transformer-based \cite{vaswaniAttentionAllYou2017}) language model. The extracted text features are combined with encoded image features (usually by a ViT \cite{dosovitskiyImageWorth16x162021}) and then passed through another transformer to generate the text answer.

\begin{figure}[!h]
    \centering
    \begin{tabular}{cc}
        \begin{subfigure}[b]{0.45\textwidth}
            \centering
            \includegraphics[width=\textwidth]{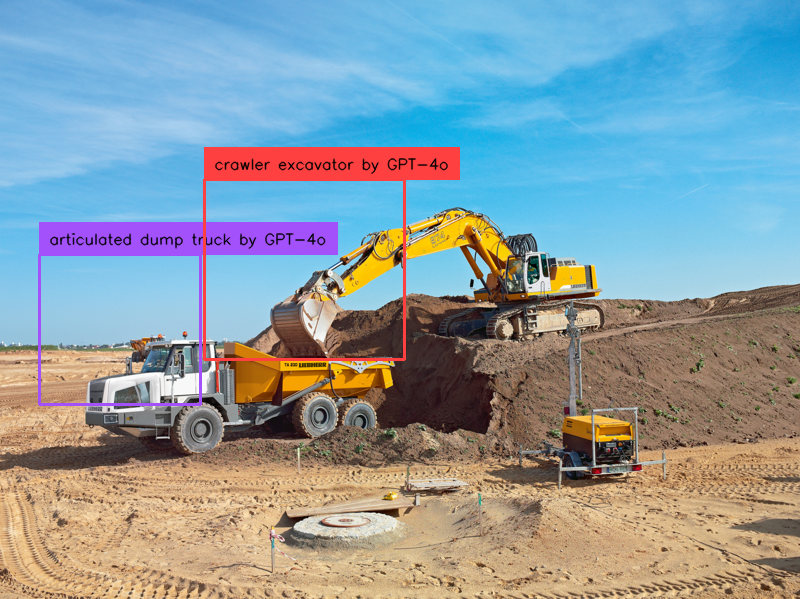}
            \caption{Response: bounding box coordinates}
            \label{subfig:ann_gpt}
        \end{subfigure} &
        \begin{subfigure}[b]{0.45\textwidth}
            \centering
            \includegraphics[width=\textwidth]{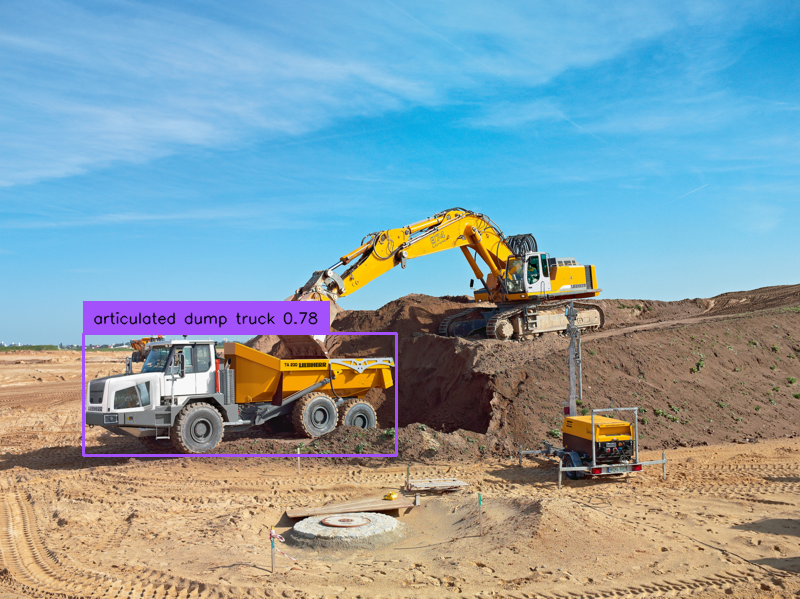}
            \caption{Response: No}
        \end{subfigure}   \\
        \begin{subfigure}[b]{0.45\textwidth}
            \centering
            \includegraphics[width=\textwidth]{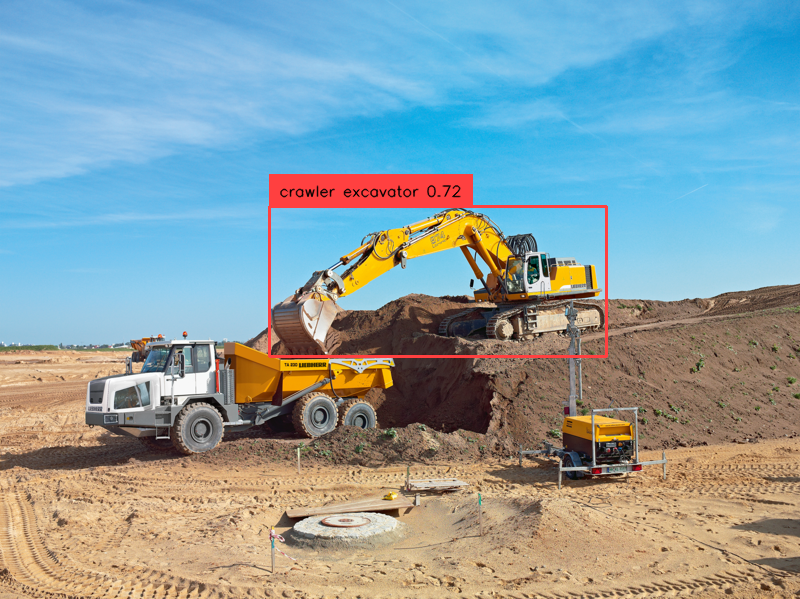}
            \caption{Response: No}
        \end{subfigure}   &
        \begin{subfigure}[b]{0.45\textwidth}
            \centering
            \includegraphics[width=\textwidth]{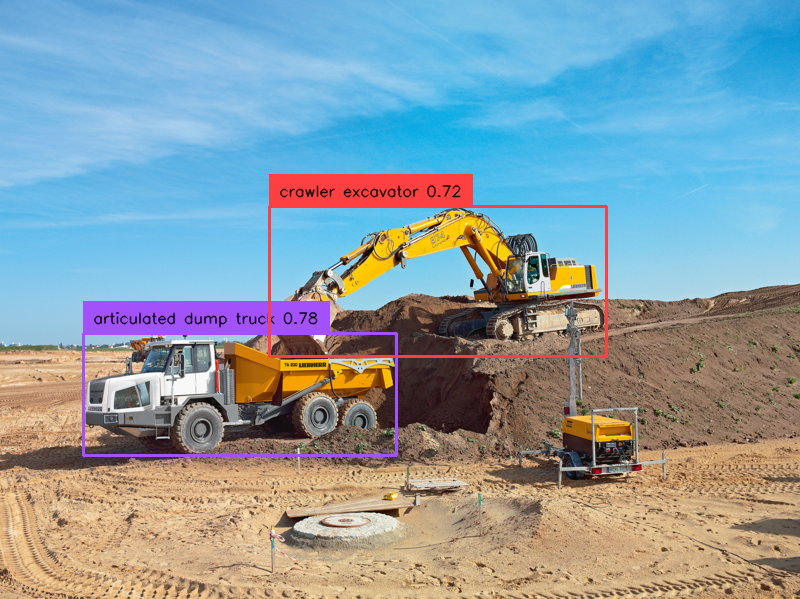}
            \caption{Response: Yes}
        \end{subfigure}
    \end{tabular}
    \caption{Bounding box annotation and review with GPT-4o. For (a), GPT-4o is asked to directly generate the bounding box coordinates given class labels and the raw image. The returned output is drawn on the image. They are nonfunctional, as expected. In other cases, GPT-4o is requested to review the candidate pseudo-labels drawn directly on top of the image and return Yes/No answers indicating whether the labeling process is flawless. The exact text prompt used for the review task can be found in \autoref{app:prompt_lmm}. For (b) and (c), only partial labels are provided, and GPT-4o correctly responds with "No". In (d), complete labels are presented, resulting in the appropriate response of "Yes" from GPT-4o.}
    \label{fig:gpt4_examples}
\end{figure}

While LMMs excel in semantic understanding and comprehension, they still struggle with tasks that require precise numerical reasoning or manipulation. For instance, directly generating bounding boxes can be problematic, as depicted in a failure case of GPT-4o in \autoref{subfig:ann_gpt}. In contrast, GPT-4o can precisely interpret the bounding box and its associated class label drawn directly on the image as visual hints, demonstrated by its correct judgments of whether the given boxes and labels are both accurate and complete, as shown in (b)-(c) in \autoref{fig:gpt4_examples}.

The proposed DART pipeline leverages GPT-4o to review the pseudo-labels generated in the bounding box annotation stage. We draw the predicted bounding boxes together with their corresponding class labels and confidence scores on top of the raw images and use these synthesized images as the visual prompts. In the text prompt, we ask GPT-4o to behave as an expert bounding box reviewer and make judgments based on three criteria: precision, recall, and fit. The corresponding three questions are:

\begin{enumerate}
    \item Precision: Does each bounding box perfectly enclose one single target object?
    \item Recall: Are all target objects localized by a bounding box?
    \item Fit: Is each bounding box neither too loose nor too tight?
\end{enumerate}

Besides the problem statement and requirement specification, the reminder section for the text prompt is also vital for an optimal response. We find that GPT-4 models tend to be overly strict for bounding box edges and inconsistent when dealing with occluded objects. Consequently, we summarize these common issues in text form (see \autoref{tab:prompt_gpt} and other specific requirements for the prompt detailed in \autoref{app:prompt_lmm}) and append them at the end of the original text prompt as a reminder. This operation makes GPT-4o much less likely to make the same mistakes. Furthermore, we also utilize Chain-of-Thought (COT) \cite{weiChainofThoughtPromptingElicits2023} reasoning to enhance the model's performance. The final text prompt used for the pseudo-label review task is detailed in \autoref{app:prompt_lmm}.

GPT-4o's output is instructed to be the responses for the three key questions with brief explanations. We consider an image to be correctly labeled only if all three questions receive positive responses. Otherwise, the image is removed from the training data.

Like bounding boxes, the generated images via Grounding DINO in \autoref{sec:dreambooth} must be validated to ensure they adhere to the requirements of downstream tasks. This verification is crucial as current SDXL models tend to produce overly imaginative images. Compared to the bounding box review based on three criteria, the image photorealism review is an easier task as it only considers photorealism. InternVL-1.5, with its robust visual comprehension, handles this task efficiently. Specifically, it is employed to conduct a preliminary review (with text prompt listed in \autoref{tab:prompt_internvl}) of images generated by SDXL after training with DreamBooth for all selected instances, assessing their photorealism and overall quality. Only the approved images further go through the annotation phase (\autoref{sec:gdino}).

After GPT-4o and InternVL-1.5, only high-quality pseudo-labels and high-fidelity images are kept for real-time object detection training. We show convincing evidence in \autoref{sec:ablation} and \autoref{sec:lmm_exp} that both LMM-based reviews significantly enhance the overall performance and accuracy of the final output models.

\subsection{Real-time object detector training for YOLOv8 and YOLOv10}
\label{sec:yolo}

\begin{figure}[!h]
    \centering
    \includegraphics[width=1\textwidth]{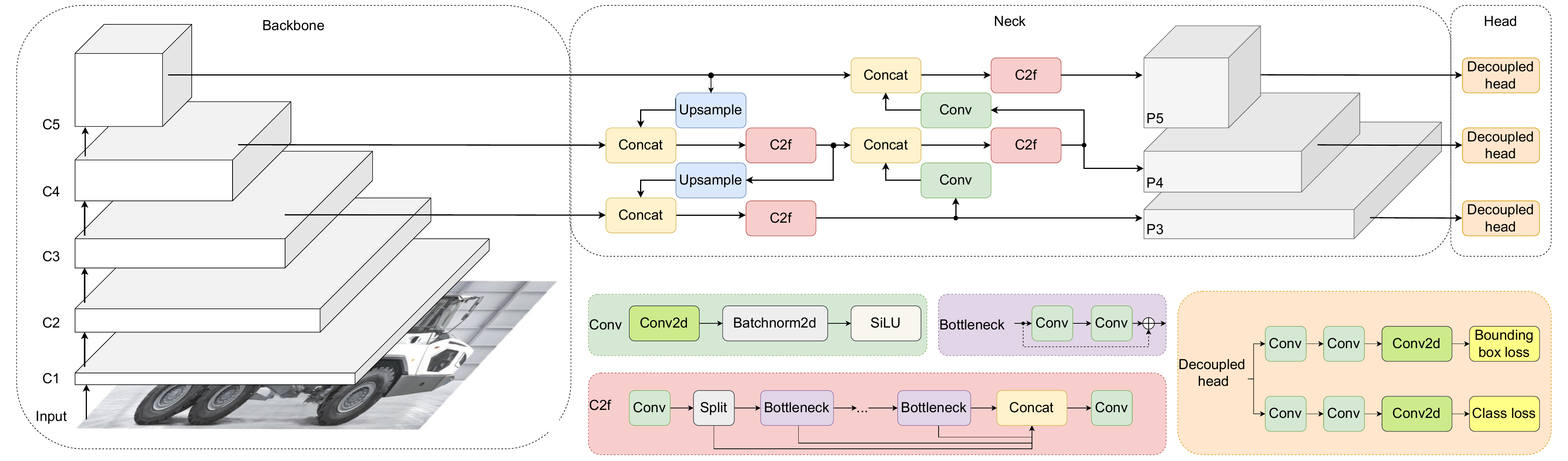}
    \caption{Model architecture of YOLOv8. Modified from \cite{jocherUltralyticsYOLOv82023, rangekingBriefSummaryYOLOv82023}}
    \label{fig:yolov8}
\end{figure}

In the realm of real-time object detection, YOLO frameworks have established themselves as the dominant methodology primarily because they effectively balance computational efficiency with high detection accuracy. Currently, the most popular YOLO variant is YOLOv8. As illustrated in the figure, YOLOv8's architecture (\autoref{fig:yolov8}) inherits the overall framework from previous YOLO iterations, utilizing the Darknet \cite{redmonYOLOv3IncrementalImprovement2018} backbone, SPP (Spatial Pyramid Pooling \cite{heSpatialPyramidPooling2014}) and PAN (Path Aggregation Network \cite{liuPathAggregationNetwork2018}) structure in the neck, and three multi-scale decoupled heads to predict bounding box coordinates and class labels. For the major block in the backbone and neck, YOLOv8 employs the C2f module, which is a faster implementation of the CSPBlock \cite{wangCSPNetNewBackbone2019} featuring two convolution layers at each end, with the CSP Bottlenecks in between (depicted as the red block in \autoref{fig:yolov8}). In terms of training strategy, the model adopts an anchor-free approach and incorporates Task Alignment Learning (TAL) \cite{fengTOODTaskalignedOnestage2021} for ground truth assignment and Distribution Focal Loss (DFL) \cite{liGeneralizedFocalLoss2020a} as the regression loss. These enhancements contribute to improved computational efficiency and detection accuracy. The concurrent YOLOv10 is the latest release in the YOLO family. It proposes a consistent dual assignments strategy for NMS-free training as well as a holistic efficiency-accuracy-driven model design featuring comprehensive adoption of pointwise, depthwise, and large-kernel convolutions with partial self-attention (PSA), which greatly reduces the computational overhead and enhances model capability.

With approved (\autoref{sec:lmm}) diversified (\autoref{sec:dreambooth}) images and their corresponding pseudo-labels (\autoref{sec:gdino}), training a real-time object detector becomes plain and easy. Since DART’s output will be deployed on edge devices, we select and train only the nano (n) and small (s) variants of YOLOv8 and YOLOv10. These models, pre-trained on the COCO dataset \cite{linMicrosoftCOCOCommon2015}, are fine-tuned using approved diversified data and pseudo-labels. Extensive experimentation reveals the best-performing YOLO models, considering the tradeoff between accuracy, speed, and model size (see \autoref{sec:yolo_exp}). By demonstrating the superior performance of the YOLO models under various conditions (see \autoref{sec:ablation} and \autoref{sec:analysis}), our experiments also underline the outstanding performance of the DART pipeline, validate the effectiveness of the DART pipeline, and pinpoint the optimal variant for each stage.

\section{Experiments}
\label{sec:experiments}

\subsection{Dataset}
\label{sec:dataset}

\textbf{Data collection.} We collect a dataset named Liebherr Products (LP) from the internal database of Liebherr, a German-Swiss multinational equipment manufacturer. During data collection, we also employ class and instance identification (in \autoref{sec:preprocessing}) for bounding box annotation (in \autoref{sec:gdino}) and data diversification (in \autoref{sec:dreambooth}), respectively.

\begin{figure}[!h]
    \centering
    \includegraphics[width=0.9\textwidth]{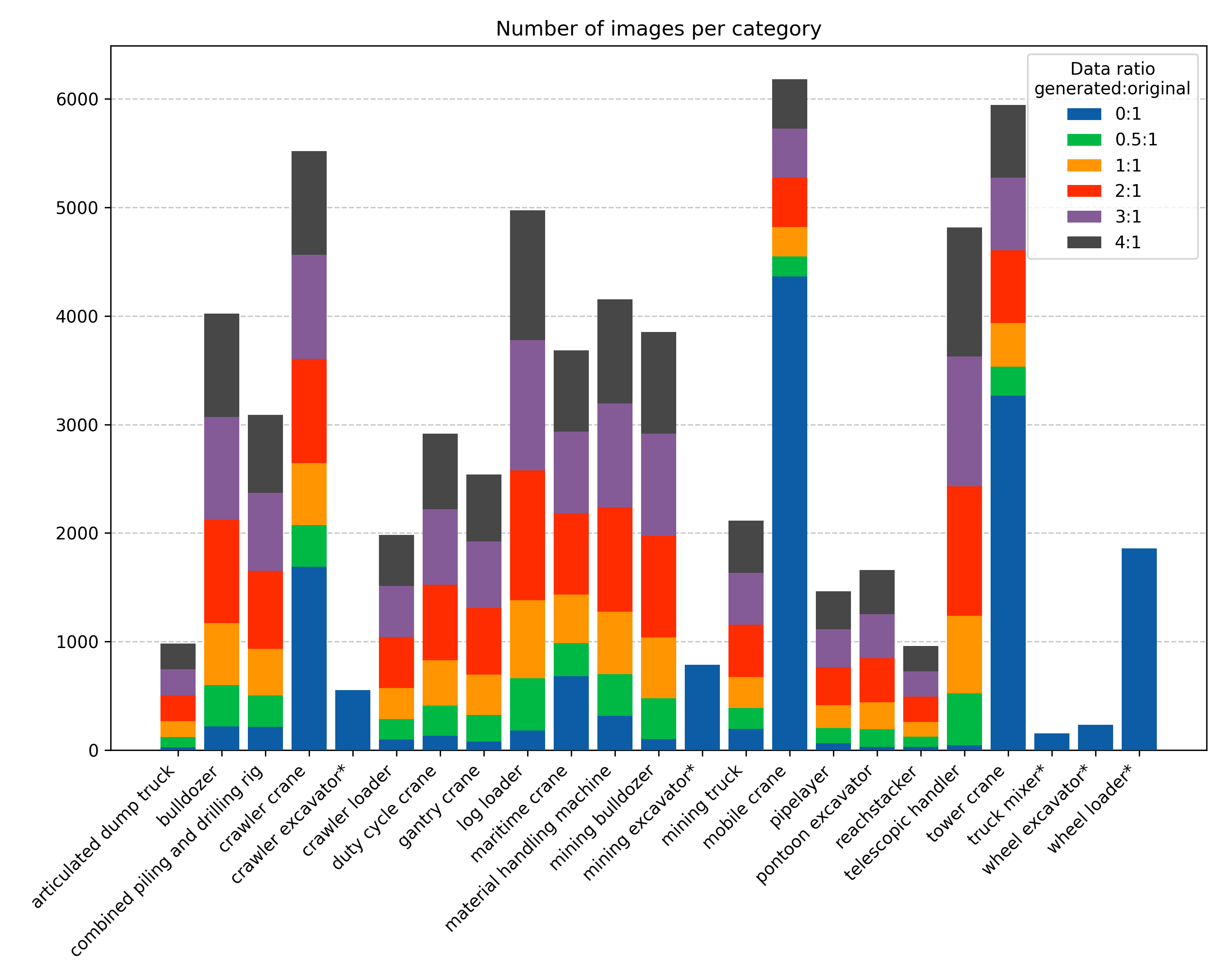}
    \caption{Number of images per category for different ratios of generated to original data. Each color in the color bar represents the incremental data increase from the previous to the current ratio. Note that five classes (marked by *) are not considered during the data diversification phase (\autoref{sec:dreambooth}) given their already strong initial performance. See \autoref{sec:yolo_exp} for more details.}
    \label{fig:bar}
\end{figure}

\textbf{Dataset composition.} The LP dataset (after preprocessing) encompasses more than 15K images of Liebherr’s extensive range of heavy machinery, such as construction machines, earthmoving equipment, deep foundation machines, mining machines, various types of cranes, material handling machines, and so on. A list of the exact 23 categories can be found either as the x-axis tick labels in \autoref{fig:bar} or in the first column of \autoref{tab:synonyms}. \autoref{fig:bar} illustrates the distribution of images in each category, represented by the blue bars labeled "0:1". The unbalanced data distribution in the LP dataset makes the downstream object detection task even more challenging. However, as we will see in \autoref{fig:cm_yolo}, the final YOLO models still achieve consistently good results across various classes, highlighting the effectiveness of DART. Furthermore, \autoref{fig:bar} also shows the data distribution after incorporating varying proportions of generated data, which will be elaborated on in the \autoref{sec:dreambooth_exp}.

\textbf{Dataset split.} The dataset is initially split into a training and a test set with an 80/20 distribution. To address class imbalance, stratified random sampling ensures each class is adequately represented in the test set. We allocate 20\% of the training set to create a validation set, which is used for hyperparameter fine-tuning. After obtaining the optimal hyperparameter set, the model is trained on the original training set with 12K images and evaluated on the 3K images from the test set. The training set could incorporate generated data in specific experiment setups to enhance data diversification. Unless otherwise specified, the generated-to-original data ratio is maintained at the optimal proportion of 3:1, as outlined in \autoref{sec:dreambooth_exp}. Notably, categories that already exhibit strong performance with only the original data are excluded from the augmentation with generated data, as detailed in \autoref{sec:yolo_exp}. All performance metrics reported in the tables and figures of this paper are based on the test set.

\subsection{Evaluation metrics}
\label{sec:metrics}

In this work, we utilize the average precision (AP) of the trained object detection models in the last stage as the evaluation metric for the proposed DART pipeline. AP summarizes the precision-recall (PR) curve into a single value, serving as an approximation of the area under the PR curve. To compute AP, precision $p$ and recall $r$ are first evaluated as follows:
\begin{equation}
    p(t) = \frac{\text{TP}}{\text{TP} + \text{FP}}, r(t) = \frac{\text{TP}}{\text{TP} + \text{FN}},
\end{equation}
where detections are classified as true (T) or false (F) according to their confidence scores and determined as positive (P) or negative (N) based on the Intersection over Union (IoU) threshold $t$.

Then a modified version of PR curve $\tilde{p}(r,t,c)$ is constructed for each class $c \in C$ given a certain IoU $t$:
\begin{equation}
    \tilde{p}(r,t,c) = \max_{r' \geq r} p(r',t,c),
\end{equation}
so that the curve will decrease monotonically instead of in a zigzag pattern.

Following COCO \cite{linMicrosoftCOCOCommon2015}, $\tilde{p}(r,t,c)$ is interpolated and evaluated at 101 points where recall \(r \in [0:.01:1]\) for AP calculation. We specify the IoU thresholds of interest by adopting two commonly used AP metrics: AP$_{50}$ and AP$_{50-95}$. The former measures precision at a single IoU of 50\%, while the latter averages precision across 10 IoU thresholds ranging from 50\% to 95\% in increments of 5\% (\(t \in [.50:.05:.95]\)). To summarize, AP$_{50-95}$ can be expressed as follows:
\begin{equation}
    \text{AP}_{50-95} = \frac{1}{10} \sum_{t \in [.50:.05:.95]} \left( \frac{1}{|C|} \sum_{c \in C} \left( 0.01 \sum_{r \in [0:.01:1]} \tilde{p}(r,t,c) \right) \right)
\end{equation}
Note that AP$_{50-95}$ is no greater than AP$_{50}$ since the precision decreases for a given recall with higher thresholds. For a well-trained model on diverse and complex datasets, such as LP, both metrics exhibit similar trends. These two consistent patterns will also be observed in the following tables. Consequently, the following sections will primarily focus on the AP$_{50-95}$. Since AP$_{50-95}$ is also commonly abbreviated as AP, we will use these two terms interchangeably.

\subsection{Implementation details}
\label{sec:implementation}

\subsubsection{Implementation of DreamBooth with SDXL}
We instantiate DreamBooth with SDXL as the subject-driven image generation framework for the data diversification phase. The SDXL model is preferable, as it significantly outperforms previous versions of Stable Diffusion models such as SD-1.5 \cite{rombachHighResolutionImageSynthesis2022}. We incorporate prior preservation loss to prevent overfitting and adjust its weight ($\lambda$ in \autoref{eq:loss}) to 1.0. Contrary to SD-1.5, training the text encoder of SDXL has limited effects; we still opt to fine-tune it to enhance the overall performance marginally. We extract 69 instances for the training of DreamBooth with SDXL in total. The minimum number of images per instance is set to 3 (as suggested by \cite{ruizDreamBoothFineTuning2023}), while the majority of instances have 10 images each (see \autoref{fig:instance}). We implement dynamic training steps by adjusting the maximum training steps based on the number of images collected for the corresponding instance. 100-160 steps per image are proven to be suitable, combining with the AdamW optimizer \cite{loshchilovDecoupledWeightDecay2017} with a learning rate $1 \times 10^{-4}$ for U-Net and $5 \times 10^{-6}$ for the text encoder. The SNR-gamma is set to 5 by default. Other important hyperparameters are explained in \autoref{app:dreambooth}.

We fine-tune two models from the same official pre-trained checkpoint for each instance: one with slightly more steps to ensure resemblance and one with fewer for diversification. For training, a general class-specific prompt, "a photo of a \{class\_name\}," is employed to generate class images for prior preservation loss. As for instance-specific prompts, we use the instance name (product number if available, otherwise a descriptive word) to form the prompt as "a photo of a <\{instance\_name\}> \{class\_name\}." We also explore different approaches to constructing identifiers and class names. However, our experiments reveal that subtle modifications have negligible effects on performance as long as the identifier is sufficiently distinctive and the class name is adequately precise. To meet the requirements of data diversification, we craft inference text prompts encompassing a broad spectrum of scenarios, as listed in \autoref{app:prompt_dreambooth}. The training processes are executed on either a single NVIDIA A100 80GB GPU or two NVIDIA 3090 GPUs, with each configuration capable of completing the training of one instance-specific SDXL model in less than an hour. The inference process requires substantially less GPU memory and is compatible with any memory optimization techniques for SDXL.

\subsubsection{Implementation of Grounding DINO}
The off-the-shelf Grounding DINO base model is employed for bounding box annotation. Through extensive comparative experiments, we determined that a box threshold of 0.27 and a text threshold of 0.25 generally return high-quality bounding boxes, resulting in only 18 images without annotations, which are consequently excluded from the pipeline. The original collected images and approved generated images are isometrically resized to 800/1333 pixels for the shorter/longer side before going through the model. As for text prompts, we follow instructions given in \autoref{sec:gdino} to construct all three prompt types as separate inputs to the Grounding DINO model and apply filtering and NMS to the combined results to obtain the candidate bounding boxes to be reviewed. The annotation process is mainly carried out (including speed test as shown in \autoref{tab:gdino_tiny_base}) by an NVIDIA A100 80GB GPU with FP16 for our case.

\subsubsection{Implementation of LMM}
We leverage GPT-4o by calling OpenAI's API for pseudo-label review. The bounding boxes and class labels are directly drawn on top of the image, and the synthesized image is compressed to 512 pixels on the longer side while keeping the aspect ratio before being specified as the vision prompt for the API. Detailed task descriptions, judging criteria, hints, suggestions, and reminders are all listed in the text prompt. The prompt construction rule is outlined in \autoref{sec:lmm}, and the exact content of text prompts is detailed in \autoref{app:prompt_lmm}. We employ the GPT-4o-based pseudo-label review only to images containing multiple annotations or annotations with confidence scores lower than 0.5 (after NMS in \autoref{sec:gdino}), which accounts for approximately 5K images since we empirically find that Grounding DINO generally provides accurate bounding boxes and labels when the confidence score exceeds 0.5. As for InternVL-1.5, we use it for a simpler task of photorealism review to filter out unrealistic images generated by each SDXL model trained by the DreamBooth framework. We run the InternVL-1.5 model locally on an NVIDIA A100 80GB GPU with the visual prompt as a single generated image and the text prompt given in \autoref{tab:prompt_internvl}.

\subsubsection{Implementation of YOLO}
As the final step of the DART pipeline, we fine-tune COCO pre-trained YOLOv8 and YOLOv10 models as the output object detector to be deployed. We follow common practice \cite{YOLOv5Ultralytics2020,liYOLOv6V3FullScale2023,liYOLOv6SingleStageObject2022,wangYOLOv7TrainableBagoffreebies2022,jocherUltralyticsYOLOv82023,wangYOLOv9LearningWhat2024,wangYOLOv10RealTimeEndtoEnd2024} to resize all images to 640$\times$640. Then, AdamW is implemented with a weight decay set to 5e-4 as the optimizer. Next, we conduct extensive experiments focusing on learning rate and its scheduling for each model and each experiment setup. The optimal initial learning rates range from 7e-5 to 5e-4, with a general trend indicating that larger models and bigger datasets benefit from lower learning rates. Mosaic augmentation \cite{bochkovskiyYOLOv4OptimalSpeed2020, YOLOv5Ultralytics2020} is adopted to enhance data diversity. Details on other critical hyperparameters are provided in \autoref{app:yolo}. Speed tests (including NMS for YOLOv8, referenced from the official codebase or paper) are conducted on an NVIDIA T4 GPU using TensorRT and FP16. The test results are reported in \autoref{tab:yolo}.

\subsection{Ablation study on the overall DART pipeline}
\label{sec:ablation}
In this section, we perform an ablation study on the overall DART pipeline. We systematically analyze the contributions of each major component by evaluating the test performances of trained real-time object detection models under various configurations. Specifically, we incrementally integrate the first three stages of DART, i.e., Grounding DINO for annotation, GPT-4o for review, and DreamBooth with SDXL for data diversification, as shown in \autoref{tab:ablation}. To evaluate the impact of each newly incorporated component, we leverage the overall performance of DART's final stage as the metric. Various YOLO models with different versions and scales, including YOLOv8n, YOLOv8s, YOLOv10n, and YOLOv10s, are fine-tuned for a comprehensive assessment. We observe a consistent trend in performance increase across all four YOLO models as new components of DART are introduced. Therefore, the subsequent analyses will primarily use YOLOv8n as the representative model. For the complete test results on the other models, please refer to \autoref{tab:ablation}.

Our baseline configuration (\#0) utilizes a YOLOv8n model without any additional modules from DART, yielding the lowest performance with an AP of just 0.064. In this setup, the model is trained on one of the most abundant datasets, LVIS \cite{guptaLVISDatasetLarge2019}. Despite LVIS's extensive coverage of over 1.2K categories across 164K images, it falls short of capturing all the classes present in our self-collected LP dataset. Consequently, we assign each LP's category to its closest kin in LVIS during inference, guided by similarity scores calculated from class names and descriptions by SBERT \cite{reimersSentenceBERTSentenceEmbeddings2019}. The results, however, are still disastrous, with a merely non-zero AP of 0.064, suggesting the necessity of the DART pipeline. Integrating the bounding box annotation stage of DART improves the AP to a usable level of 0.771. Further incorporation of GPT-4o to rule out inferior pseudo-labels shows an incremental performance increase, resulting in a proper AP of 0.802 and thereby indicating the additional value of LMMs for the DART pipeline. The full power of DART is achieved by the final integration of data diversification. Adding data generated by SDXLs trained for each instance in a DreamBooth framework (and approved by InternVL-1.5) to the original training set boosts the final best AP to 0.832 for the YOLOv8n model, underlining the superior efficacy of the comprehensive DART pipeline.

\begin{table}[!h]
\centering
\caption{Ablations on key stages of the DART pipeline.}
\label{tab:ablation}
\begin{tabular}{ccccccc}
\toprule
\textbf{\#} & \textbf{Model}  & \textbf{Grounding DINO} & \textbf{GPT-4o}& \textbf{DreamBooth} & \textbf{AP$^{test}_{50}$} & \textbf{AP$^{test}_{50-95}$} \\ 
\midrule
0 & \multirow{4}{*}{YOLOv8n}  &                &                &              & 0.091 & 0.064    \\
1 &                          & \checkmark     &                &               & 0.864 & 0.771    \\
2 &                          & \checkmark     & \checkmark     &               & 0.899 & 0.802    \\
\rowcolor[gray]{0.92}
3 &                          & \checkmark     & \checkmark     & \checkmark    & 0.915  & 0.832    \\ 
\midrule
4 & \multirow{3}{*}{YOLOv8s}  & \checkmark     &                &              & 0.883 & 0.789    \\
5 &                           & \checkmark     & \checkmark     &              & 0.903 & 0.815    \\
\rowcolor[gray]{0.92}
6 &                          & \checkmark     & \checkmark     & \checkmark    & 0.924 & 0.850    \\ 
\midrule
7 &\multirow{3}{*}{YOLOv10n} & \checkmark     &                &               & 0.854 & 0.738    \\
8 &                          & \checkmark     & \checkmark     &               & 0.877 & 0.777    \\
\rowcolor[gray]{0.92}
9 &                          & \checkmark     & \checkmark     & \checkmark    & 0.879  & 0.802    \\ 
\midrule
10 &\multirow{3}{*}{YOLOv10s} & \checkmark     &                &              & 0.867  & 0.781        \\
11 &                         & \checkmark     & \checkmark     &               & 0.889  & 0.809        \\
\rowcolor[gray]{0.92}
12 &                          & \checkmark     & \checkmark     & \checkmark   & 0.917  & 0.833        \\
\bottomrule
\end{tabular}

\end{table}

\subsection{Quantitative Analysis}
\label{sec:analysis}

In this section, we conduct internal analyses for the four major stages of the proposed DART pipeline by investigating the best solutions for each key module via comprehensive experiments and ablation studies within each stage.

\subsubsection{Analysis of DreamBooth with SDXL}
\label{sec:dreambooth_exp}
We begin by examining the impact of incorporating extra training data generated by SDXL models trained for each instance using the DreamBooth framework. Across all configurations, we maintain a consistent test set of 3K images and augment the training set with generated data (with both images and pseudo-labels approved by LMM) in varying multiples (from 0 to 4) of the original 12K images.

\autoref{tab:ratio} presents the results of training YOLOv8 models under varying ratios of generated to original training data, highlighting the influence of data shift and diversification. When trained exclusively on generated data (1:0 ratio), the YOLOv8n model achieves a low AP of 0.031. This poor performance stems from a significant distributional mismatch between the original and generated datasets by design. Specifically, our prompt design for generation emphasizes maximal novelty to ensure semantic diversity (see \autoref{app:prompt_gdino}), which deliberately deviates from the distribution of the original images. Moreover, although DreamBooth with SDXL can produce images with fine-grained resemblance to the original objects, it does not consistently preserve identity-level details. Together, these factors induce an extreme data shift, severely impairing generalization to the original test distribution. By contrast, training solely on original data (0:1 ratio) eliminates this shift and yields an AP of 0.815.

\begin{table}[!h]
\centering
\caption{Impact of training data ratios (generated:original) on YOLOv8 model performance.}
\label{tab:ratio}
\begin{tabular}{ccccc}
\toprule
\textbf{Training data ratio} & \multicolumn{2}{c}{\textbf{YOLOv8n}} & \multicolumn{2}{c}{\textbf{YOLOv8s}} \\
\textbf{generated:original} & \textbf{AP$^{test}_{50}$} & \textbf{AP$^{test}_{50-95}$}  & \textbf{AP$^{test}_{50}$} & \textbf{AP$^{test}_{50-95}$} \\ 
\midrule
1:0     & 0.047 & 0.031 & 0.047 & 0.031 \\
0:1     & 0.899 & 0.802 & 0.903 & 0.815 \\
0.5:1   & 0.902 & 0.815 & 0.911 & 0.834 \\
1:1     & 0.908 & 0.822 & 0.916 & 0.836 \\
2:1    & 0.909 & 0.825 & 0.914 & 0.843 \\
\rowcolor[gray]{0.92}
3:1     & 0.915 & 0.832 & 0.924 & 0.850 \\
4:1     & 0.914 & 0.826 & 0.918 & 0.839 \\
\bottomrule
\end{tabular}

\end{table}

\begin{figure}[!h]
    \centering
    \includegraphics[width=0.75\columnwidth]{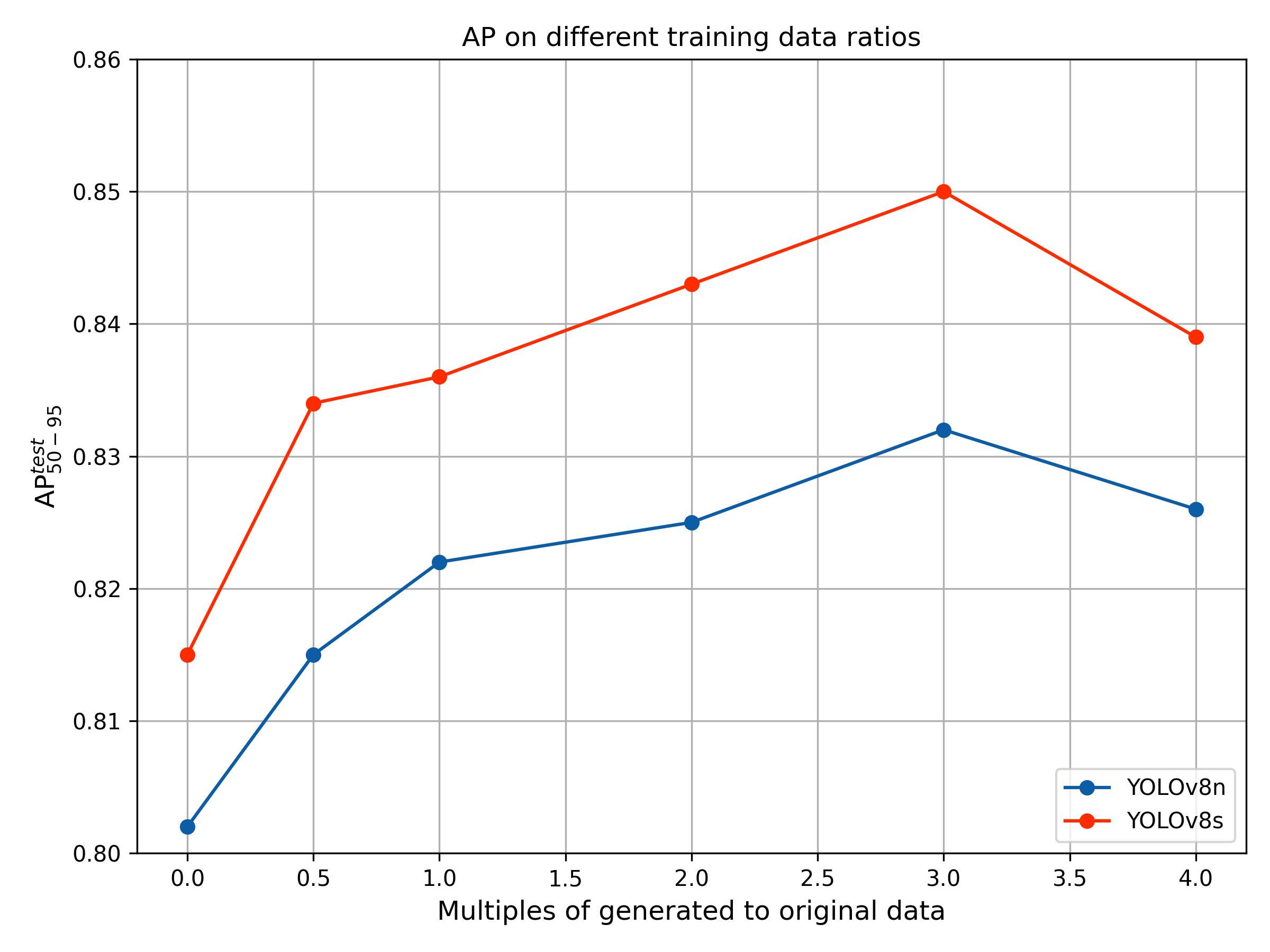}
    \caption{Test performance of YOLOv8 models across different training data ratios (generated:original).}
    \label{fig:ratio}
\end{figure}

Interestingly, as the proportion of generated data increases alongside original data, model performance improves rather than degrades, revealing the effect of data diversification. At a 3:1 ratio (75 \% of the data are generated), the YOLOv8n model achieves its highest AP of 0.832 (see \autoref{fig:ratio}). This performance gain arises from the enriched diversity in training samples, which promotes broader feature learning and improved generalization. The overall trend thus reflects a trade-off: while excessive data shift alone is detrimental, a moderate inclusion of diverse synthetic data can counterbalance this by expanding the model’s exposure to varied contexts. However, beyond a certain threshold, the harmful effects of data shift begin to outweigh the benefits of a marginal increase in diversification, leading to a slight decline in AP.

In summary, original data is indispensable for a decent test performance, while generated data offers significant benefits, especially when the ratio is such that the impact of data shift and diversification mechanisms balance. For the LP dataset, the optimal scenario involves using three times the generated data as the original data. The almost identical trends observed for both YOLOv8n and YOLOv8s in \autoref{fig:ratio} further corroborate this conclusion.

\subsubsection{Analysis of Grounding DINO}
\label{sec:gdino_exp}

Two Grounding DINO variants, which utilize Swin tiny and base \cite{liuSwinTransformerHierarchical2021} as the backbone, respectively, are tested for the bounding box annotation task. It can be seen from \autoref{tab:gdino_tiny_base} that using the more powerful base variant increases test performance for the YOLOv8n model, suggesting better consistency in bounding box annotations and label assignment. However, both of the Grounding DINO variants run too slow to fit the real-time requirements, even on one of the most powerful GPUs, NVIDIA A100. Consequently, we only adopt the Grounding DINO base model for all data annotation tasks for better bounding box quality, despite a slightly slower speed, which is negligible for the offline labeling process.

\begin{table}[!h]
\centering
\caption{Comparisons between Grounidng DINO tiny and base.}
\label{tab:gdino_tiny_base}
\begin{threeparttable}
\begin{tabular}{ccccc}
\toprule
\textbf{Grounding DINO} & \textbf{Input size} & \textbf{Latency}\tnote{$\dagger$} & \textbf{AP$^{test}_{50}$}\tnote{$\ddagger$} & \textbf{AP$^{test}_{50-95}$}\tnote{$\ddagger$} \\
\midrule
Tiny & 800$\times$1333 & 112ms & 0.858 & 0.745\\
\rowcolor[gray]{0.92}
Base & 800$\times$1333 & 161ms & 0.864 & 0.771\\
\bottomrule
\end{tabular}
\begin{tablenotes}
    \footnotesize
    \item[$\dagger$] Latency is tested on a NVIDIA A100 80GB with FP16.
    \item[$\ddagger$] APs are provided by a YOLOv8n model trained on all original images annotated by Grounding DINO tiny and base, respectively.
\end{tablenotes}
\end{threeparttable}

\end{table}

\begin{table}[!h]
\centering
\caption{Quantities and average confidence scores of Grounding DINO annotations under various types of text prompts and operations. In particular, "all" means concatenating all class names as one text prompt for each image. The three prompt types that are practically applied in DART are "original," "synonym," and "co-occurring". After combining the output from these three prompt types, "filtering" is applied to sift out most annotations with a confidence score < 0.5. Finally, class-agnostic non-maximum suppression "NMS" is employed to get rid of overlapping annotations. More details regarding prompt types and operations can be found in \autoref{sec:gdino}.}
\label{tab:gdino_ann}
\begin{tabular}{lcc}
\toprule
\multicolumn{1}{c}{\textbf{Type}}     & \textbf{\# annotations} & \textbf{Average score} \\
\midrule
\multicolumn{1}{c}{all}        & 76913          & 0.33          \\
\midrule
\multicolumn{1}{c}{original/co-occurring} & 33271          & 0.49          \\
+ synonym/co-occurring  & 76664          & 0.47          \\
+ filtering ($\geq$0.5) & 30580          & 0.66          \\
\rowcolor[gray]{0.92}
+ NMS      & 17309          & 0.65          \\
\bottomrule
\end{tabular}
\end{table}

Prompt engineering plays a vital role in the performance of Grounding DINO. In \autoref{sec:gdino}, we outline the construction process for three types of prompts (original, synonym, and co-occurring) and introduce approaches for merging their outputs, including filtering and NMS. In this section, we will discuss the results from these prompts and merging operations reported thoroughly in \autoref{tab:gdino_ann}. As reported in \autoref{sec:gdino}, simple concatenation of all class names as a single text prompt for one image (type "all") leads to inferior performance reflected quantitatively by the low average confidence score of 0.33 in the table and qualitatively by the first row of \autoref{fig:gdino_err}. Therefore, we start with only a predefined class name as the original prompt and the co-occurring trick for multiple categories. Next, synonym replacement is applied for all text prompts, which results in more than double annotations with only a marginal change in the average confidence score. Synonym prompts are indispensable because our experiments reveal that a substantial number of remaining bounding boxes, which survive the high-probability filtering and class-agnostic NMS, are originally derived from synonyms. We ultimately achieve non-overlapping bounding boxes with an overall score of 0.65, where the average score for each category is depicted by the blue bars in \autoref{fig:bar} (note that the y-axis starts at 0.5). This ensures that, at least within the Grounding DINO framework, only high-quality bounding boxes are selected. These bounding boxes will be further examined by GPT-4o.

\begin{figure}[!h]
    \centering
    \includegraphics[width=0.9\textwidth]{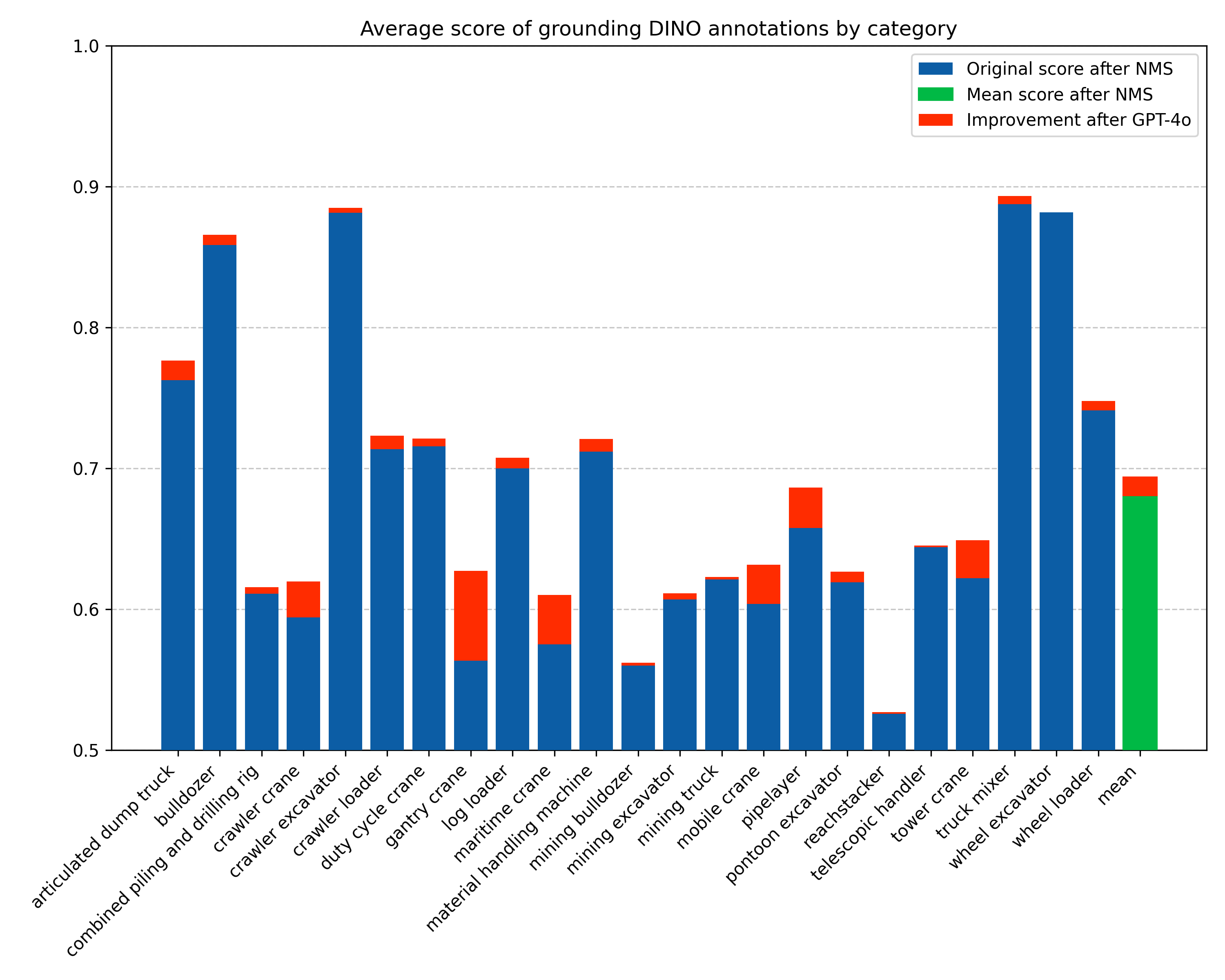}
    \caption{Class-wise average confidence score of Grounding DINO annotations after NMS filtering (blue) and GPT-4o review (red).}
    \label{fig:gdino_score}
\end{figure}

\subsubsection{Analysis of LMM}
\label{sec:lmm_exp}
DART utilizes GPT-4o for bounding box review. The detailed construction of visual and text prompts is thoroughly discussed in \autoref{sec:lmm} and further detailed in \autoref{app:prompt_lmm}. According to the responses from the pseudo-label review by GPT-4o, we eliminate slightly less than 2K images with only disapproved bounding boxes. This results in improvements in the confidence score of bounding boxes among all categories as displayed by the red bars in \autoref{fig:bar}, indicating the consistent judgment between GPT-4o and Grounding DINO regarding high-quality bounding box annotations.

\begin{figure}[!h]
    \centering
    \begin{subfigure}[b]{0.45\columnwidth}
        \centering
        \includegraphics[width=\columnwidth]{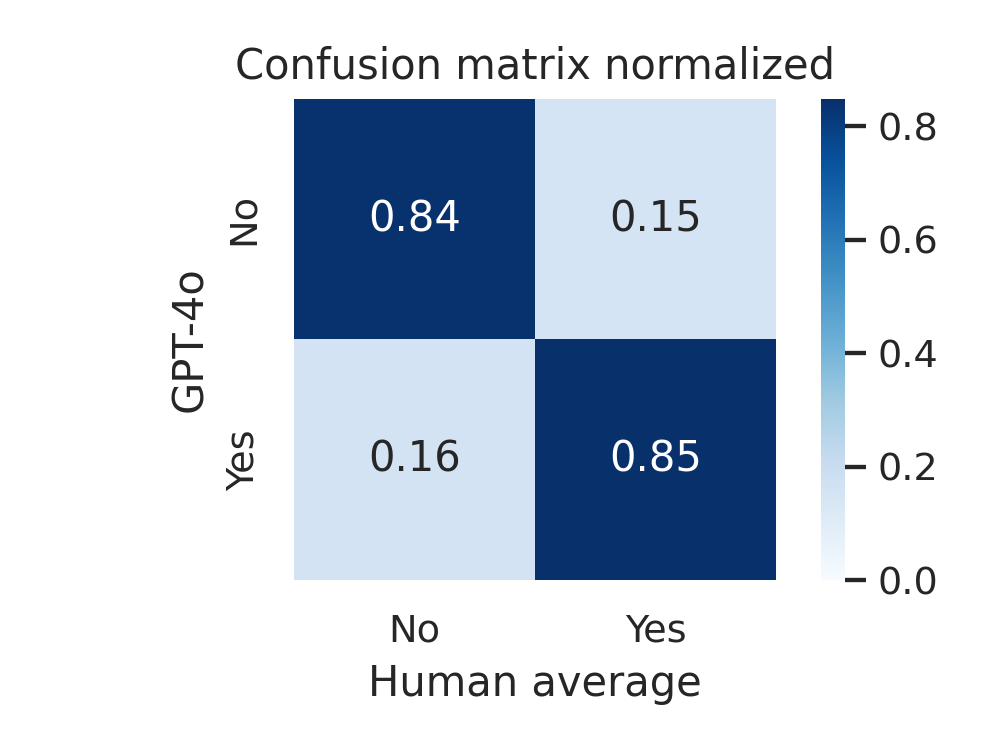}
        \caption{GPT-4o vs human average}
    \end{subfigure}
    \hfill
    \begin{subfigure}[b]{0.45\columnwidth}
        \centering
        \includegraphics[width=\columnwidth]{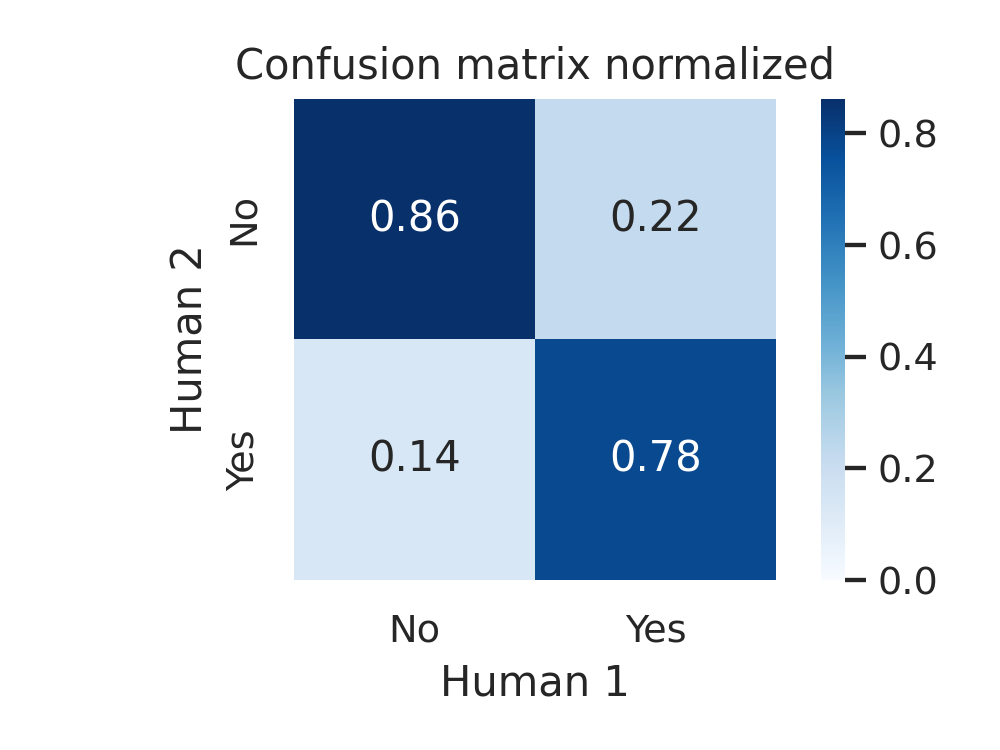}
        \caption{Human vs human}
    \end{subfigure}
    \caption{Comparisons of bounding box review consistency. The left confusion matrix displays the agreement between GPT-4o and the human average, whereas the right matrix compares the annotations of two experienced human labelers. GPT-4o demonstrates human-level performance in bounding box review given the same text and visual information.}
    \label{fig:cm_gpt_human}
\end{figure}

It turns out that GPT-4o's decision is even more in line with experienced human labelers. We conduct a small-scale human evaluation experiment by providing experienced labelers with the same instructions as GPT-4o and recording their responses to 50 randomly selected images with pseudo-labels drawn on top. In \autoref{fig:cm_gpt_human}, the left confusion matrix illustrates the results of GPT-4o vs average human performance, while the right exhibits the agreement level between two labelers for reference. Notably, GPT-4o and human labelers give consistent responses for approximately 85\% of the cases, a rate comparable to the agreement between two human labelers.

Despite GPT's exceptional visual comprehension abilities, its API costs are high. Therefore, we leverage the open-source model InternVL-1.5 for the straightforward task of verifying the photorealism of generated images. More details on this task are also given in \autoref{app:prompt_lmm}.

\begin{table}[!h]
\centering
\caption{Ablations on LMM-based reviews.}
\label{tab:lmm}
\begin{tabular}{cccc}
\toprule
\textbf{GPT-4o} & \textbf{InternVL 1.5} & \textbf{AP$^{test}_{50}$} & \textbf{AP$^{test}_{50-95}$} \\
\midrule
 & & 0.871 & 0.782 \\
\checkmark & & 0.909 & 0.818 \\
\rowcolor[gray]{0.92}
\checkmark & \checkmark & 0.915 & 0.832 \\
\bottomrule
\end{tabular}

\end{table}

We design a scenario to evaluate the effectiveness of both LMMs by comparing the test performance of a YOLOv8n model trained with and without the images they disapproved. First, we construct a dataset using the optimal generated-to-original data ratio of 3:1, as identified in \autoref{sec:gdino_exp}. As shown in the first row of \autoref{tab:lmm}, training a YOLOv8n model directly on this unfiltered dataset results in an AP of 0.782. In contrast, training solely on the original data achieves a higher AP of 0.802, indicating that naively adding generated images—without any quality filtering—can degrade performance. This underscores the necessity of effective filtering: when the same dataset is curated using GPT-4o to remove low-quality generated samples, the AP rises to 0.818. This not only demonstrates the advantage of GPT-4o-based filtering but also confirms the benefit of data diversification when guided by reliable quality control. Finally, applying InternVL-1.5 to filter for photorealism among SDXL-generated images leads to the best performance, achieving an AP of 0.832.

\subsubsection{Analysis of YOLO}
\label{sec:yolo_exp}

\begin{figure}[!h]
    \centering
    \begin{subfigure}[b]{.95\textwidth}
        \centering
        \includegraphics[height=0.4\textheight]{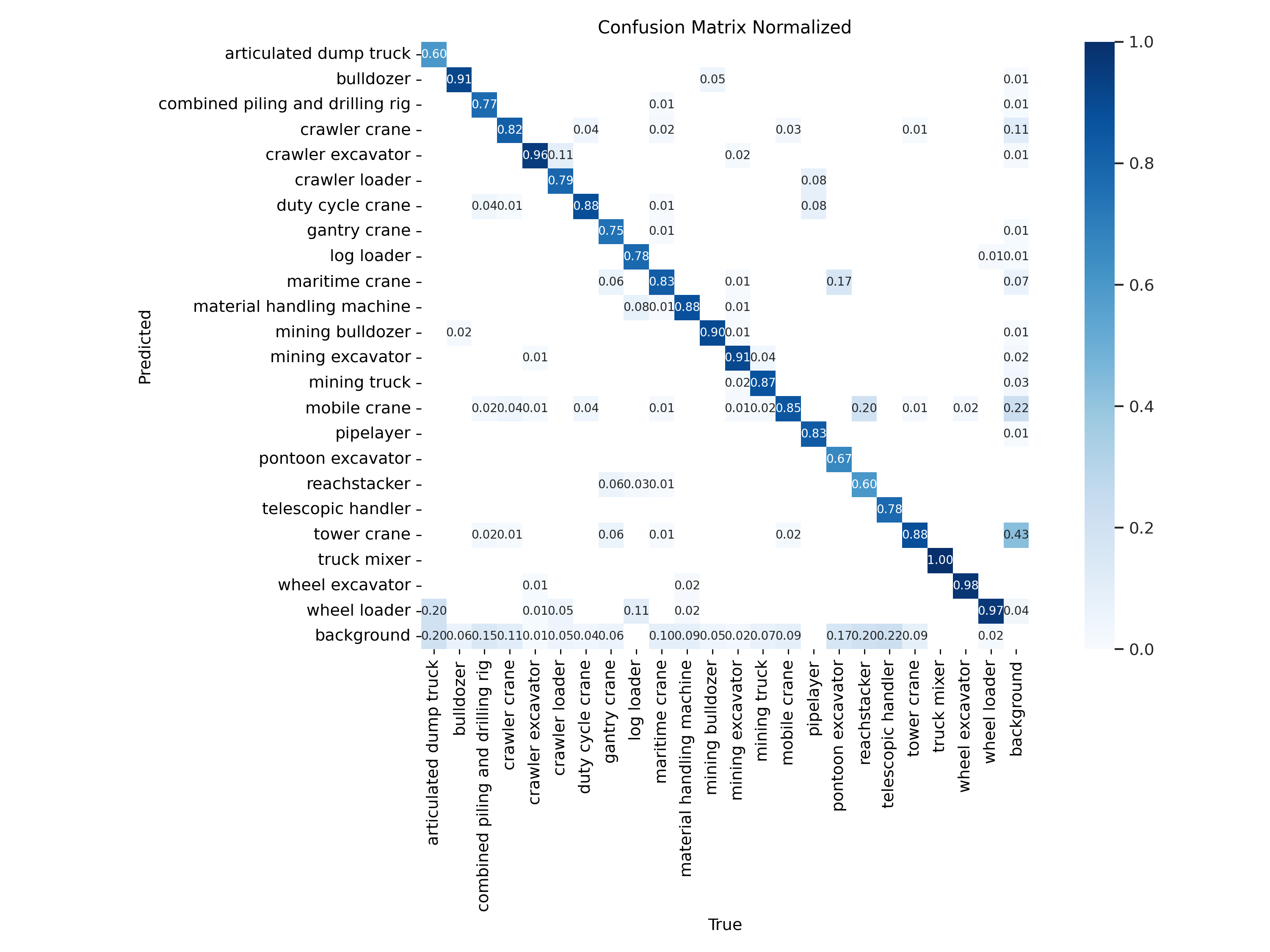}
        \caption{only with DART's annotation and training}
    \end{subfigure}
    \vspace{10pt}
    \begin{subfigure}[b]{.95\textwidth}
        \centering
        \includegraphics[height=0.4\textheight]{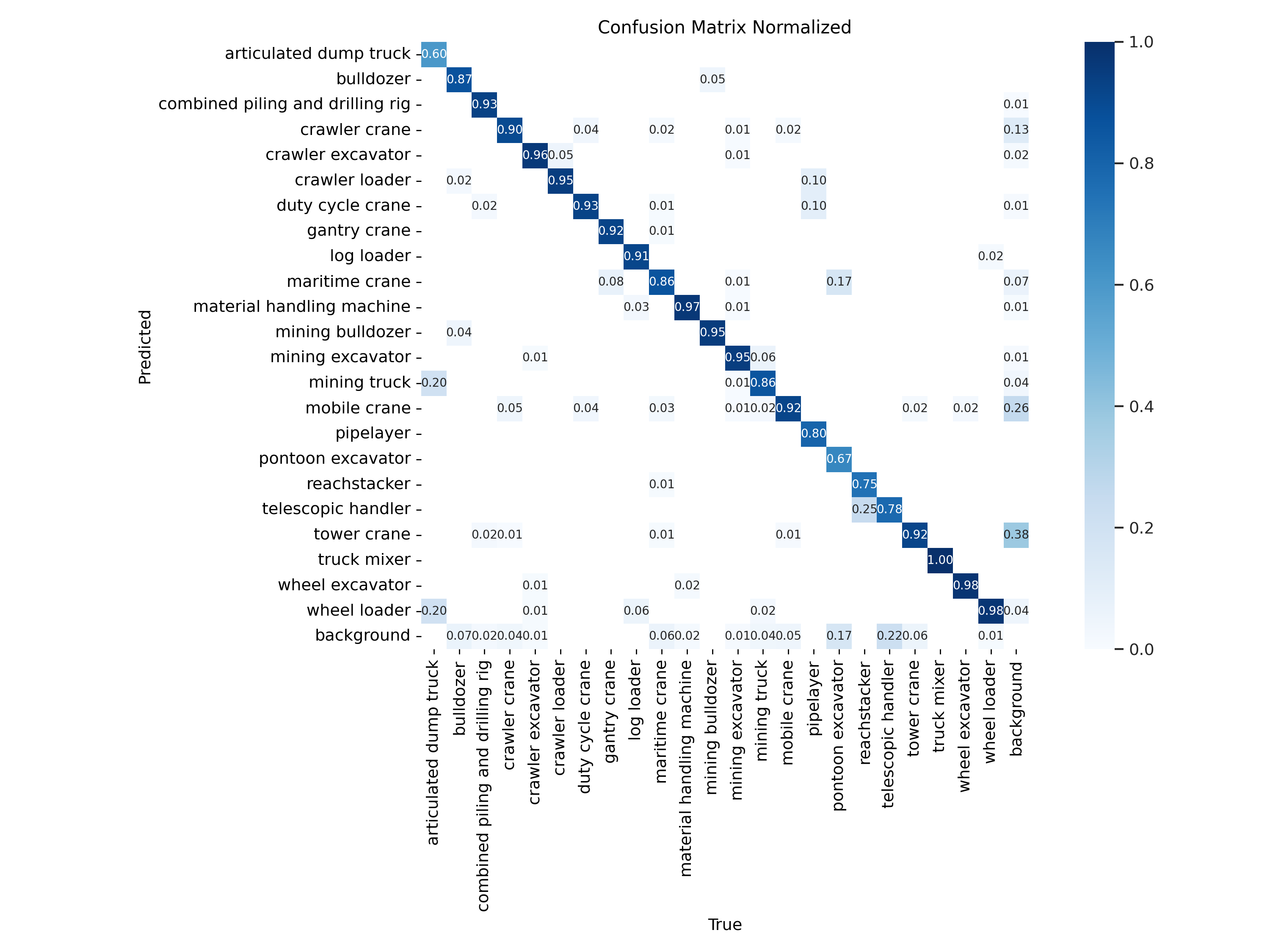}
        \caption{with full DART}
    \end{subfigure}
    \caption{Confusion matrices derived from the class-wise test performance of the same YOLOv8n model enhanced by incomplete (a) and full (b) DART pipeline.}
    \label{fig:cm_yolo}
\end{figure}

YOLO models are chosen as the final real-time object detector for DART. \autoref{fig:cm_yolo} depicts the class-wise test performance of two identical YOLOv8n models applied with partial (a) and full (b) DART. The first confusion matrix is derived from the YOLOv8n model, on which only the annotation and training stage of DART is employed. Note that we choose these two DART components from the outset since the pre-trained YOLO models fail catastrophically in object detection on the LP dataset as explained in \autoref{sec:ablation}. The realization that providing zero-shot performance comparison is meaningless already emphasizes the essential need for DART.

Without the full ability of DART, we observe several misclassifications (non-diagonal elements) and subpar true positive rates (diagonal elements), particularly in categories such as "combined pilling and drilling rig," "crawler loader," "log loaders," and so on, indicating room for improvement. However, certain categories, such as crawler excavators and truck mixers, already exhibit satisfactory performance. We hypothesize that this is due to their relative prevalence compared to other categories and their inclusion in several public datasets (e.g., LVIS). This enables Grounding DINO to provide more accurate and consistent bounding boxes, as indicated by the high confidence scores of these classes in \autoref{fig:gdino_score}. Consequently, these categories do not require further data diversification. Instead, we focus on diversifying data for categories with temporarily lower performance, allocating more quota of generated images to these classes to improve their results.

The second confusion matrix in \autoref{fig:cm_yolo} represents the results from the full DART pipeline. Obviously, the selective data diversification process described above pays off. Misclassifications are reduced across almost all categories. True positive detections for the diversified categories considerably elevate without negatively impacting the performance of the previously well-performing categories. This indicates the effectiveness of selective data diversification, which is therefore employed for all experiments involving the addition of generated diversified data

\begin{table}[!h]
\centering
\caption{Comparisons of YOLO models.}
\label{tab:yolo}
\begin{threeparttable}
\begin{tabular}{ccccccc}
\toprule
\textbf{Model} & \textbf{Input size} & \textbf{\#Param}. & \textbf{FLOPs}  & \textbf{Latency}\tnote{$\dagger$} & \textbf{AP$^{test}_{50}$} & \textbf{AP$^{test}_{50-95}$} \\
\midrule
\rowcolor[gray]{0.92}
YOLOv8n & 640$\times$640 & 3.2M & 8.7G  & 2.47ms & 0.915 & 0.832 \\
YOLOv8s& 640$\times$640 & 11.2M & 28.6G  & 2.74ms & 0.924 & 0.850 \\
\rowcolor[gray]{0.92}
YOLOv10n & 640$\times$640 & 2.3M & 6.7G  & 1.84ms & 0.879 & 0.802 \\
YOLOv10s & 640$\times$640 & 7.2M & 21.6G  & 2.49ms & 0.917 & 0.833 \\
\bottomrule
\end{tabular}
\begin{tablenotes}
    \footnotesize
    \item[$\dagger$] Latency is tested on a NVIDIA Tesla T4 with TensorRT FP16, referenced from the official codebase or paper.
\end{tablenotes}
\end{threeparttable}
\end{table}

Finally, the statistics for all real-time object detectors in this study are listed in \autoref{tab:yolo}. Each YOLO variant is enhanced using the optimal configurations derived from the DART pipeline. This includes photorealistic synthetic images generated by fine-tuned SDXL models via DreamBooth, subsequently filtered by InternVL-1.5 and mixed with original data at the optimal ratio of 3:1. Bounding box annotations are provided by the Grounding DINO base model, while its output pseudo-labels are validated through GPT-4o. Final hyperparameters are selected via extensive fine-tuning experiments to ensure peak performance. The AP values are always derived from the same test set. 

\begin{figure}[!h]
    \centering
    \includegraphics[width=1\columnwidth]{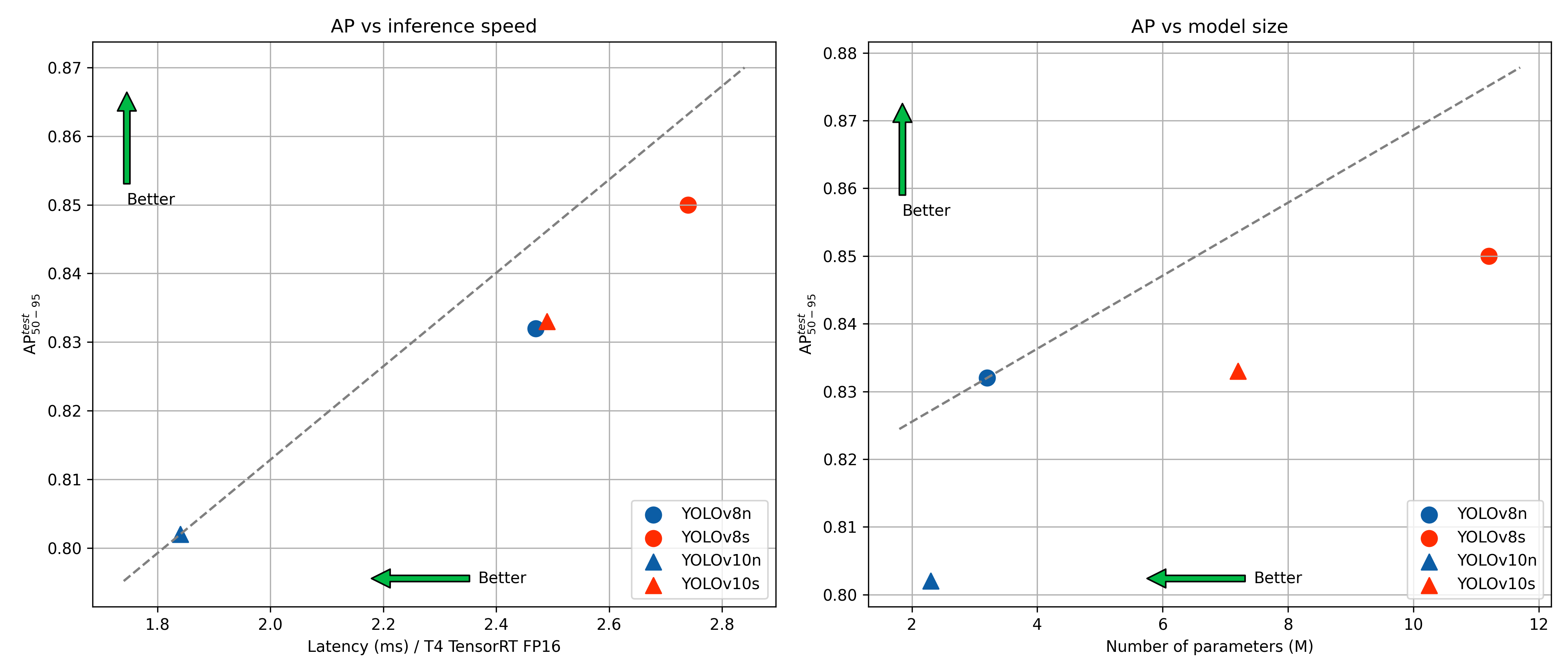}
    \caption{Test performance comparisons of YOLOv8 and YOLOv10 models in terms of accuracy-latency (left) and accuracy-size (right) tradeoff. YOLOv10n and YOLOv8n reach the Pareto frontier in these two scenarios, AP vs. inference speed and AP vs. model size, respectively. The dotted lines in both figures are derived from nearly collinear points of the three non-optimal models and are translated to the position of the optimal model to highlight its superiority.}
    \label{fig:yolo_performance}
\end{figure}

As seen from the third column of \autoref{tab:yolo}, the YOLOv10 models are designed to possess fewer parameters than the YOLOv8 model of the same category (indicated by the suffix n or s). Due to efficiency-driven model design, the YOLOv10n model achieves an AP of 0.802 in just 1.84ms (on a T4 GPU with TensorRT FP16). This is emphasized in the left subfigure of \autoref{fig:yolo_performance}, where other models fall slightly behind YOLOv10n in the AP-speed tradeoff.  Increasing the number of parameters by approximately 40\% and inference time by 30\%, the YOLOv8n model significantly boosts the AP to 0.832. The right diagram of \autoref{fig:yolo_performance} clearly shows that YOLOv8n achieves a Pareto optimal balance between AP and model size, outperforming all other models in this regard. If the sole focus is on AP, YOLOv8s becomes the optimal choice, albeit at the cost of the highest parameter count and the slowest inference time. We reasonably speculate that larger and more complex models, such as the m/l/x variants of the YOLO series or state-of-the-art transformer-based object detectors \cite{zongDETRsCollaborativeHybrid2023,wangInternImageExploringLargeScale2023} would further elevate AP. However, this contradicts DART's goal of real-time performance on edge devices, so they are not included in this study.

In summary, the fastest model in our experiments is YOLOv10n, the most accurate one is YOLOv8s, but the overall best model for real-time object detection purposes is YOLOv8n, as it achieves the Pareto optimal balance between AP and model size, with a practical AP performance of 0.832. We propose the trained YOLOv8n as the final component of the DART pipeline. In contrast to the disastrous performance without DARG and the subpar results of an incomplete DART, the comprehensive DART pipeline clearly demonstrates superior efficacy in enhancing the capabilities of real-time object detection models. The fact that DART accomplishes these advancements without manual annotation and extra data collection further highlights its significance in effectiveness and efficiency.

\subsection{Qualitative Analysis}
\label{sec:vis}

In this section, we conduct qualitative analyses using visualizations to examine the model’s performance during the data diversification, bounding box annotation, and LMM-based review stages. We focus on three specific scenarios in this section, while numerous other visualizations can be found in \autoref{app:vis}.

\begin{figure}[!h]
    \centering
    \setlength{\tabcolsep}{1pt}
    {\scriptsize
    \begin{tabular}{c@{\hskip 5pt} c@{\hskip 5pt} c c c c}
    
        \begin{tabular}{c c}
            \includegraphics[width=0.092\linewidth,height=0.092\linewidth]{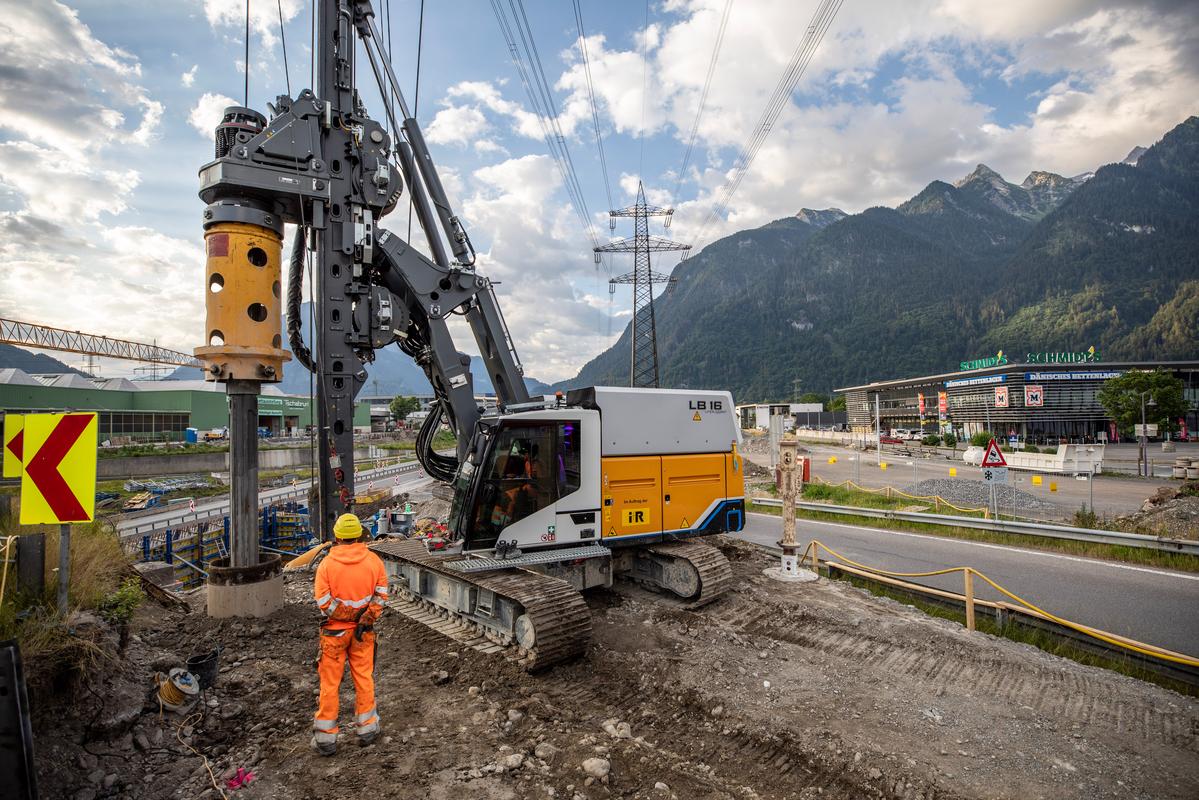} & 
            \includegraphics[width=0.092\linewidth,height=0.092\linewidth]{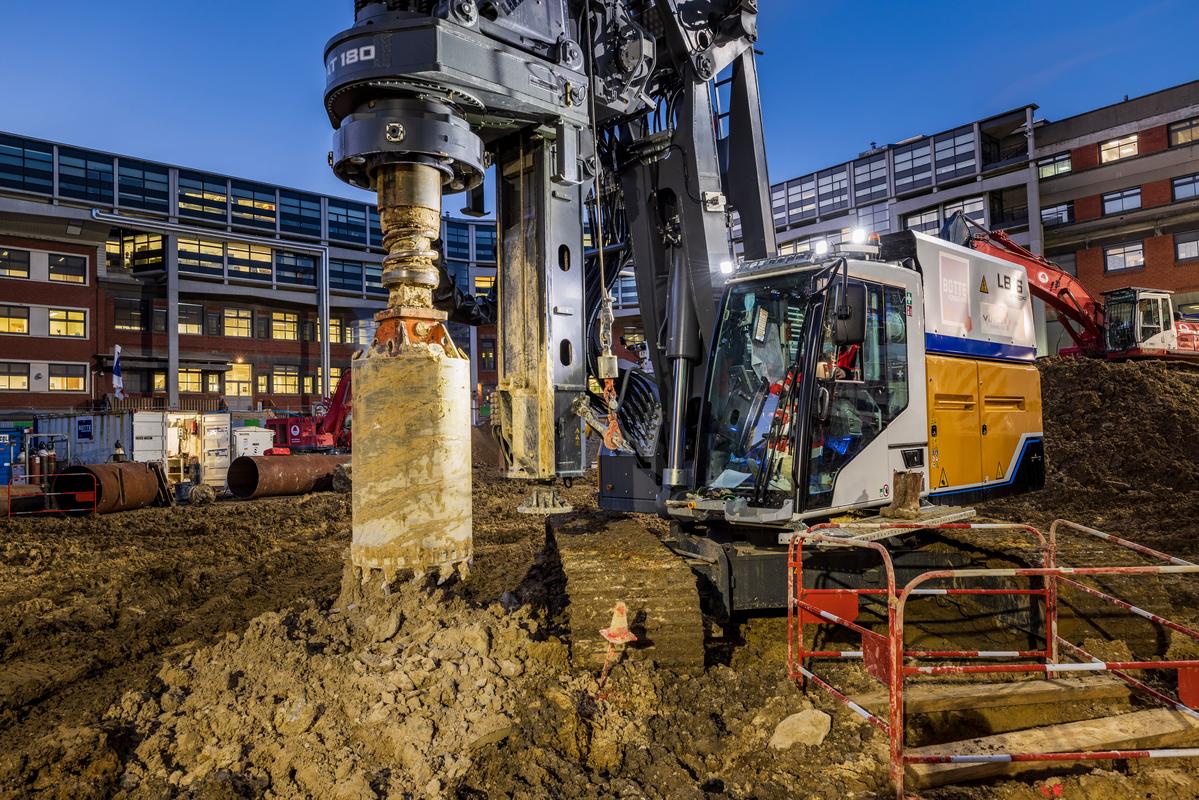} \\
            \includegraphics[width=0.092\linewidth,height=0.092\linewidth]{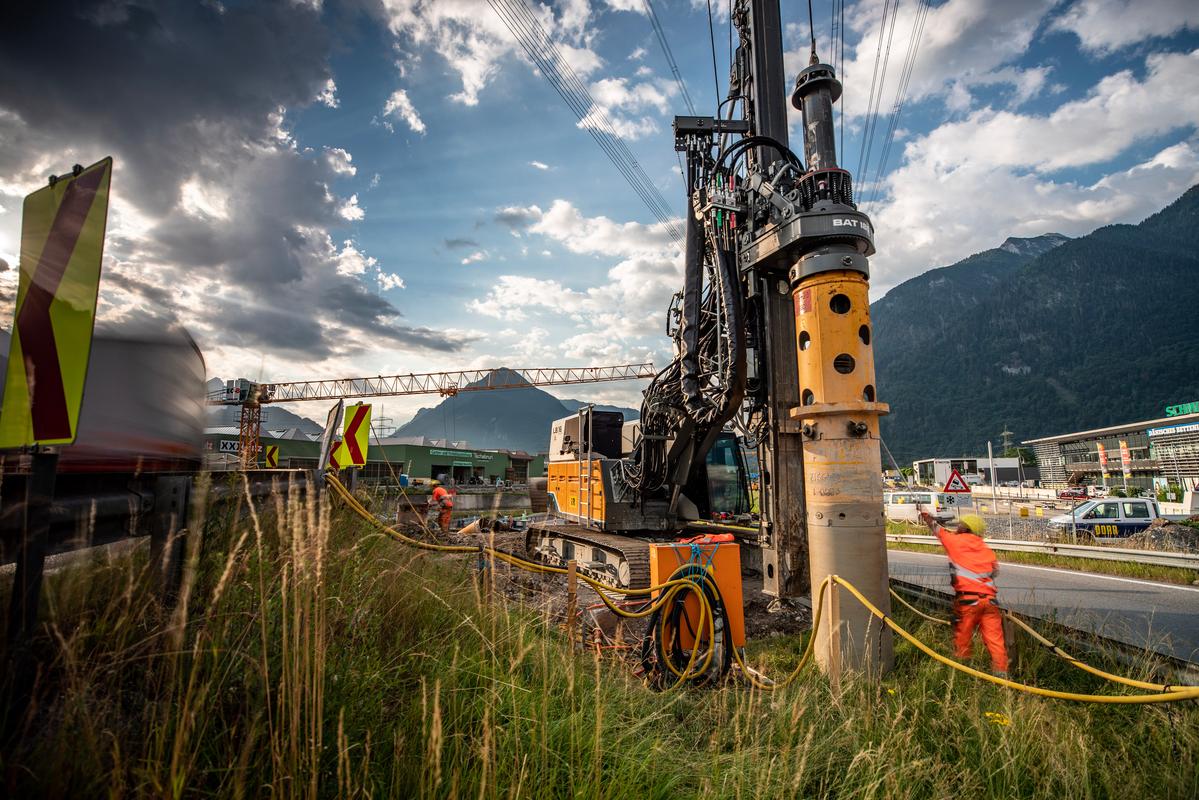} & 
            \includegraphics[width=0.092\linewidth,height=0.092\linewidth]{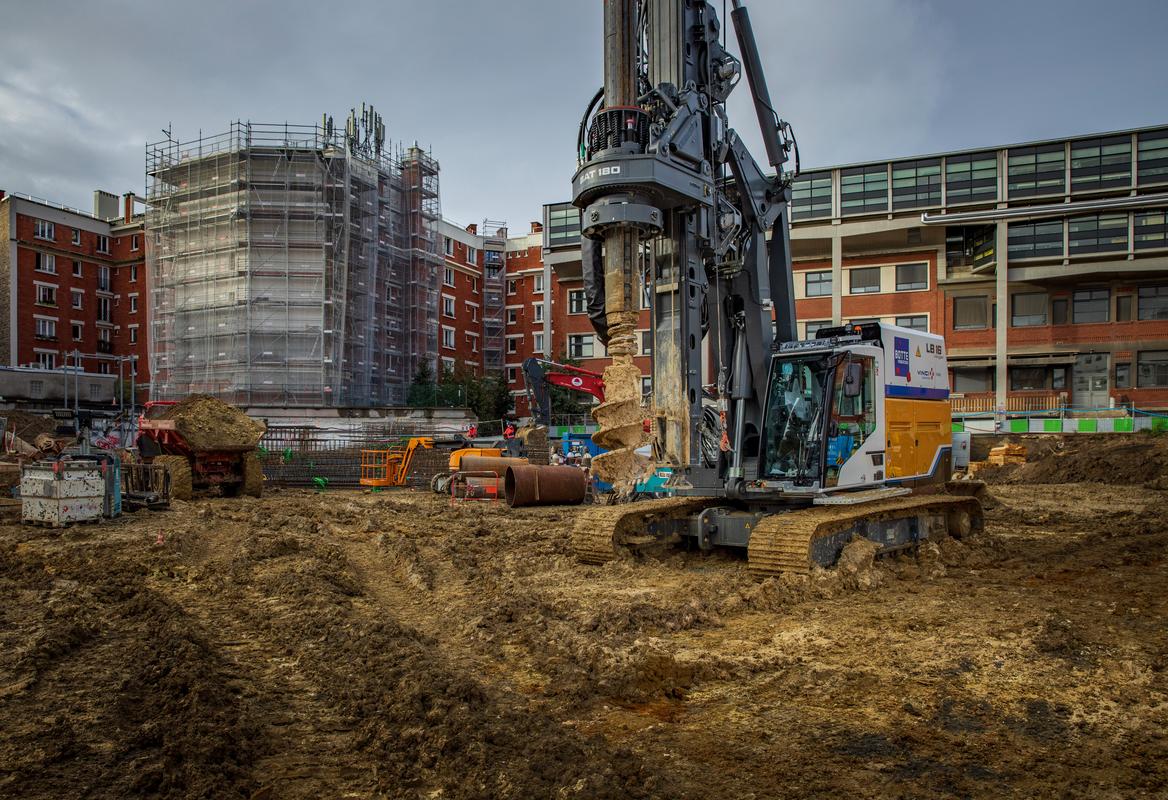}
        \end{tabular}
        &
        $\rightarrow$
        &
        \begin{tabular}{c}
        \includegraphics[width=0.184\linewidth]{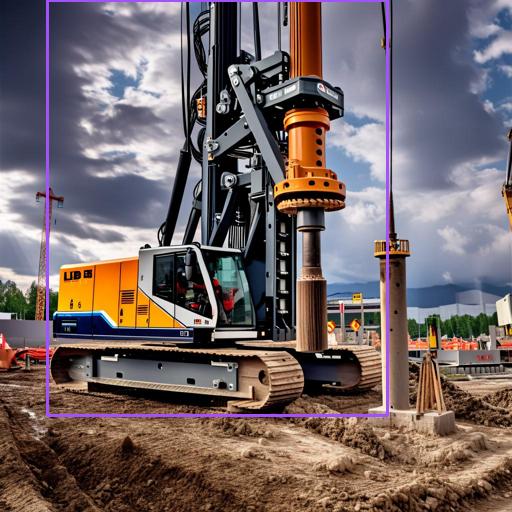}
        \end{tabular} &
        \begin{tabular}{c}
        \includegraphics[width=0.184\linewidth]{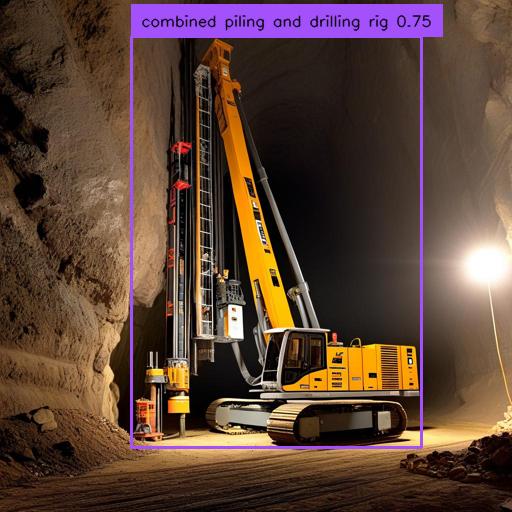} 
        \end{tabular} &
        \begin{tabular}{c}
        \includegraphics[width=0.184\linewidth]{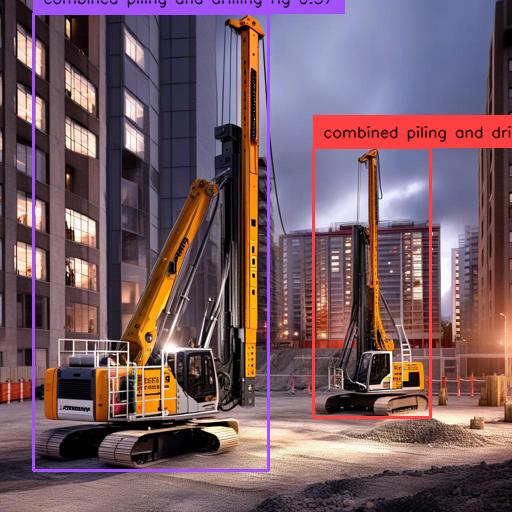} 
        \end{tabular} &
        \begin{tabular}{c}
        \includegraphics[width=0.184\linewidth]{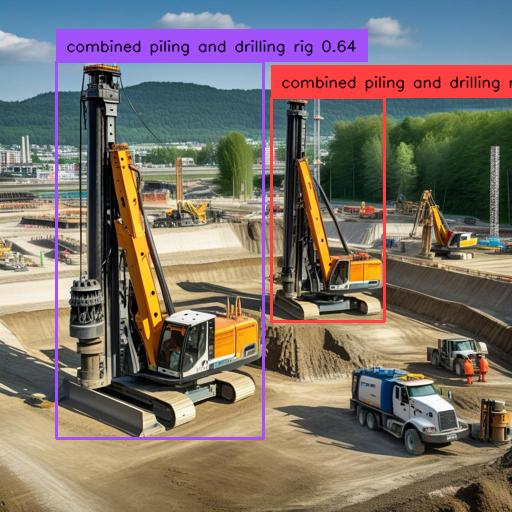}
        \end{tabular} \\
        
        {\footnotesize Instance data} & & {\begin{tabular}{c@{}c@{}c@{}c@{}} 3. Cloudy\\construction site \end{tabular}} & {\begin{tabular}{c@{}c@{}c@{}c@{}} 34. Underground\\construction site \end{tabular}} & {\begin{tabular}{c@{}c@{}c@{}c@{}} 10. Right\\side \end{tabular}} & {\begin{tabular}{c@{}c@{}c@{}c@{}} 41. Multiple\\machines sunny \end{tabular}} \\ \\

    \end{tabular}}
    \caption{Visualization of data diversification and bounding box annotation for the category “combined pilling and drilling rig." Instance data on the left are utilized for training an SDXL under the DreamBooth framework. The trained SDXL generates diversified data during inference using designed prompts. The IDs and descriptions of the corresponding prompts used to generate the image are shown as the corresponding subtitles. Please refer to \autoref{tab:prompts_dreambooth} for the exact content of each text prompt. Bounding box coordinates and labels are given by Grounding DINO and are drawn directly on top of the generated images on the right side. Also see \autoref{fig:approved_annotated_generated_images_app} and \autoref{fig:sdxl_vs_sd15} for more illustrations of other classes.}
    \label{fig:approved_annotated_generated_images}
\end{figure}

\autoref{fig:approved_annotated_generated_images} showcases a randomly selected category, “combined pilling and drilling rig,” to demonstrate the visualization of the data diversification and bounding box annotation stages. Additional illustrations of other categories can be found in \autoref{fig:approved_annotated_generated_images_app} and \autoref{fig:sdxl_vs_sd15}. As described in \autoref{sec:methods}, we first select some instance data and use them to train SDXL under the DreamBooth framework. The trained SDXL generates diversified data during inference with the help of designed prompts (with their IDs and descriptions shown as subtitles for each generated image on the right side; refer to \autoref{tab:prompts_dreambooth} for the corresponding text). After approval by InternVL-1.5 (omitted in the figure but showcased in \autoref{fig:image_grid_gen_d}), pseudo-labels (directly drawn on top of each image) are obtained through Grounding Dino. As illustrated in \autoref{fig:approved_annotated_generated_images}, the SDXL model trained with DreamBooth accurately generated realistic objects. Guided by diversified text prompts, the resulting generated images effectively depicted the scenes described. The designed prompts used for data generation also allow for the customization of background, pose, and quantity of target objects, greatly enriching the original dataset to meet personalized requirements. The bounding box annotations are also highly precise. In summary, DART significantly increases dataset diversity and generates accurate pseudo-labels for the newly created data. These enhancements substantially improve the performance of downstream YOLO models compared to training on the original dataset alone, as shown repeatedly in \autoref{sec:ablation} and \autoref{sec:analysis}.

\begin{figure}[!h]
    \centering
    \begin{tabular}{ccc}
        \begin{subfigure}[b]{0.32\textwidth}
            \includegraphics[width=\textwidth]{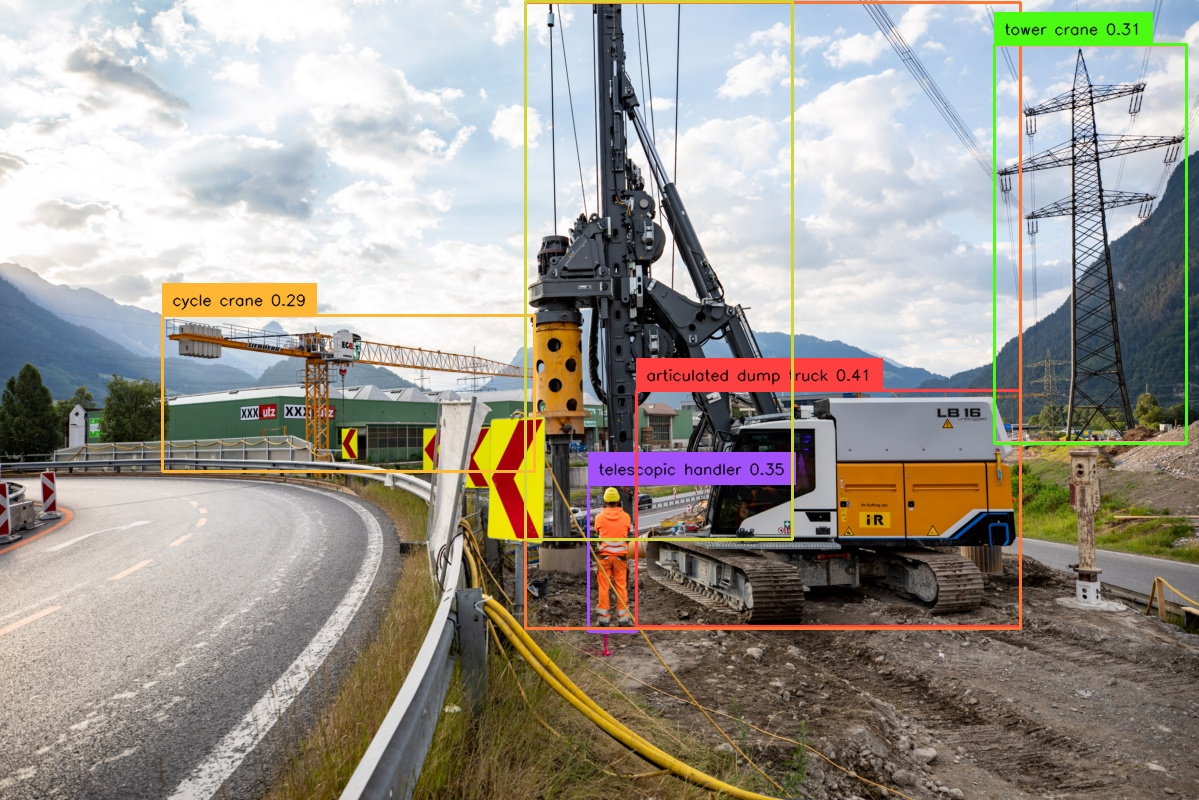}
            \caption{All prompt 1}
        \end{subfigure} &
        \begin{subfigure}[b]{0.32\textwidth}
            \includegraphics[width=\textwidth]{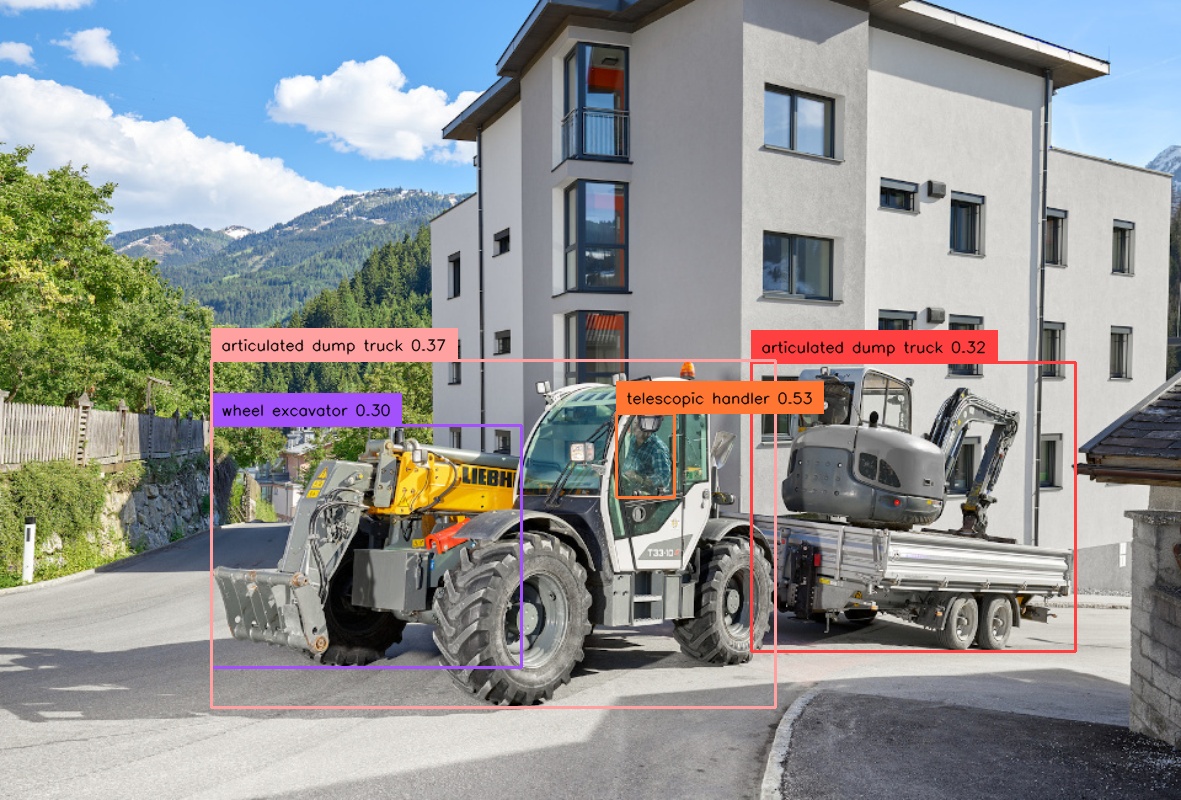}
            \caption{All prompt 2}
        \end{subfigure} &
        \begin{subfigure}[b]{0.32\textwidth}
            \includegraphics[width=\textwidth]{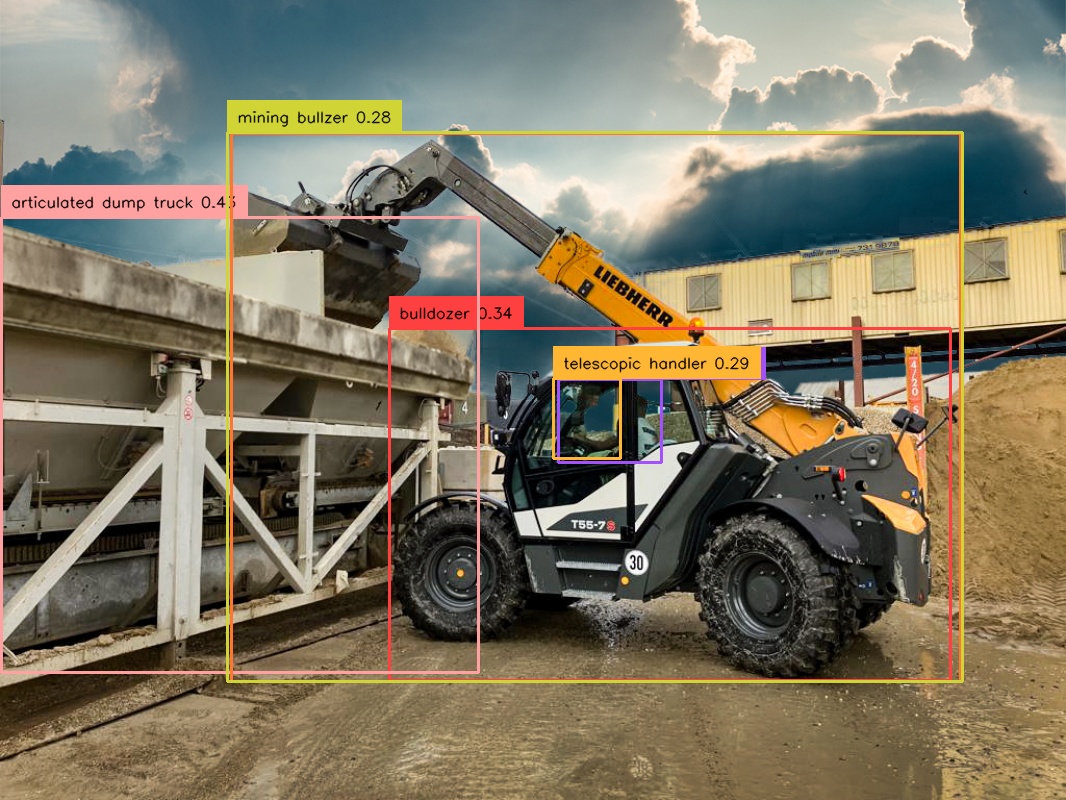}
            \caption{All prompt 3}
        \end{subfigure} \\

        \begin{subfigure}[b]{0.32\textwidth}
            \includegraphics[width=\textwidth]{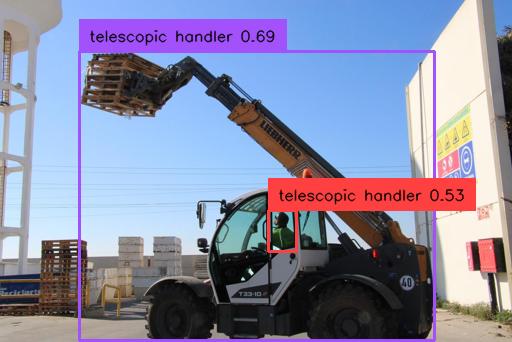}
            \caption{Ambiguity of text}
        \end{subfigure} &
        \begin{subfigure}[b]{0.32\textwidth}
            \includegraphics[width=\textwidth]{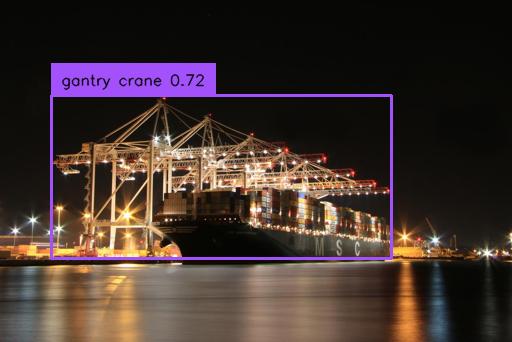}
            \caption{Ambiguity of labeling}
        \end{subfigure} &
        \begin{subfigure}[b]{0.32\textwidth}
            \includegraphics[width=\textwidth]{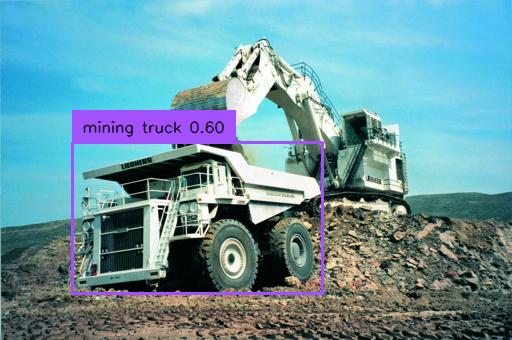}
            \caption{Omission}
        \end{subfigure} \\
        
        \begin{subfigure}[b]{0.32\textwidth}
            \includegraphics[width=\textwidth]{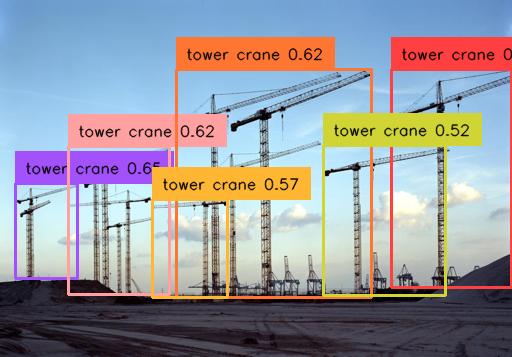}
            \caption{Small objects}
        \end{subfigure} &
        \begin{subfigure}[b]{0.32\textwidth}
            \includegraphics[width=\textwidth]{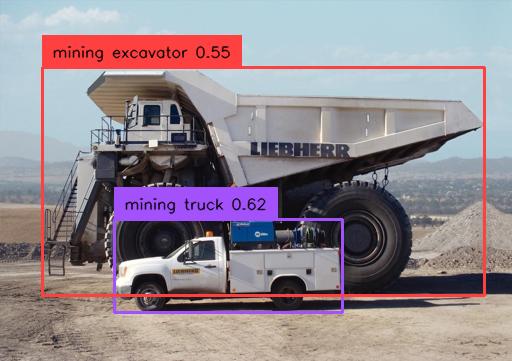}
            \caption{Misclassification*}
        \end{subfigure} &
        \begin{subfigure}[b]{0.32\textwidth}
            \includegraphics[width=\textwidth]{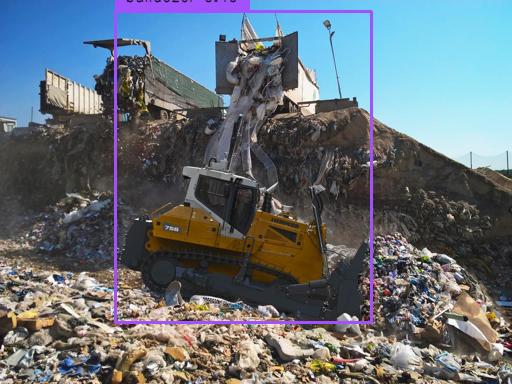}
            \caption{Misalignment}
        \end{subfigure} \\
    \end{tabular}
    \caption{Common data annotation errors by OVD (Grounding DINO). The first row is annotation results using all class names as a single text prompt. The second and third rows demonstrate common issues for OVD. The * denotes the only case when the issue is not discovered by GPT-4o-based pseudo-label review.}
    \label{fig:gdino_err}
\end{figure}

\begin{figure}[!htbp]
    \centering
    \includegraphics[width=0.95\columnwidth]{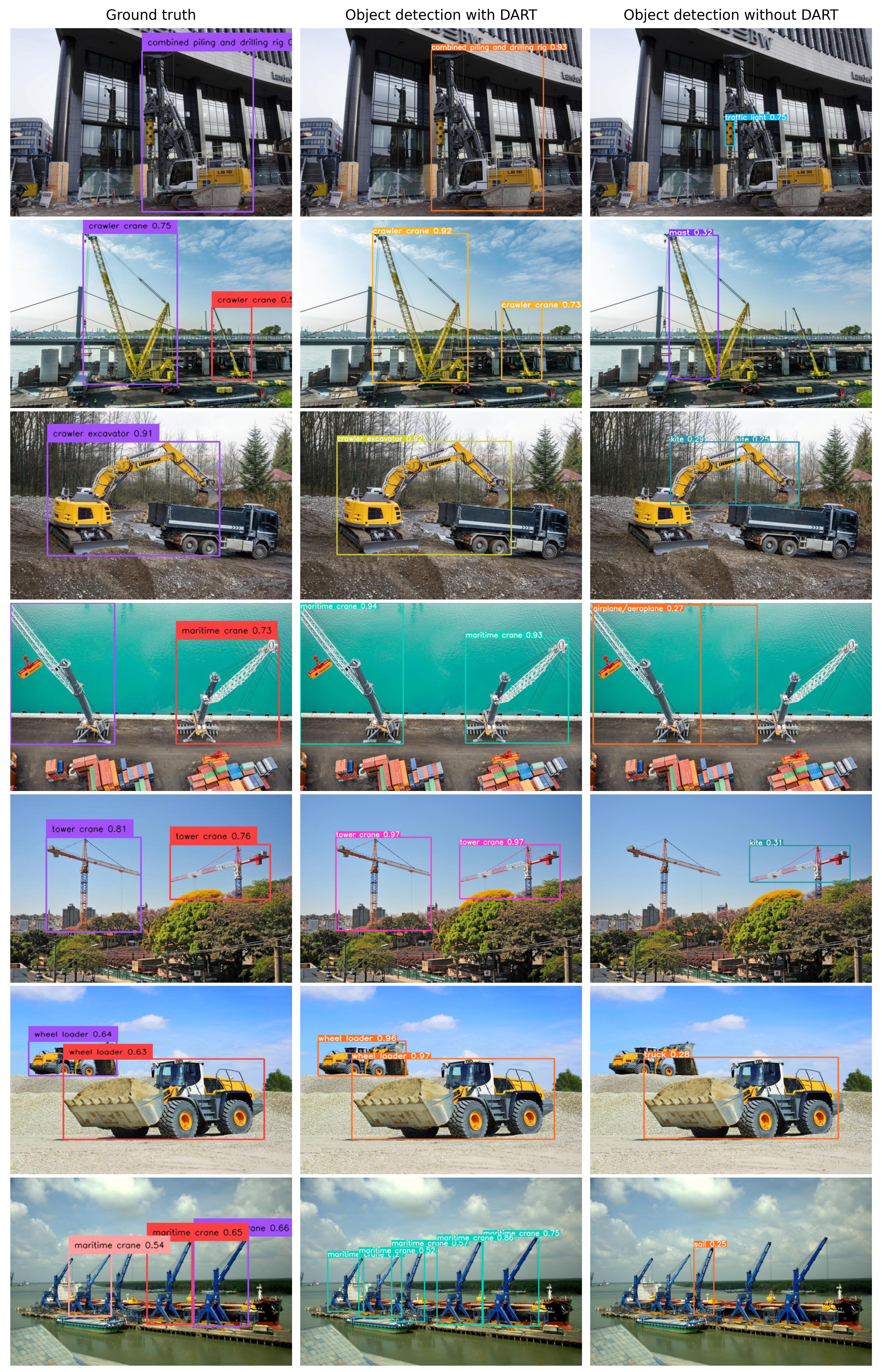}
    \caption{Comparisons of object detection results with and without DART on test set images.}
    \label{fig:predictions_1}
\end{figure}

Derived from the responses from GPT-4o-based pseudo-label review, Grounding DINO delivers feasible results in even 60\% of hard cases with confidence scores below 0.5. Nonetheless, current state-of-the-art open-vocabulary detection (OVD) models still exhibit certain issues. As shown in \autoref{fig:gdino_err}, Grounding DINO’s performance significantly declines when the prompt is simply too long. This limitation prevents us from using fully concatenated class names as a single text prompt (which is the case for the first row in \autoref{fig:gdino_err} and marked as "all prompt"). Instead, we must adopt a workaround involving original class names, synonyms, and co-occurring classes and their combinations as detailed in \autoref{sec:gdino}. Additionally, text ambiguity can pose problems. For instance, “handler” (case d in \autoref{fig:gdino_err}) can mean either an object or a person, and “gantry crane” (e) can refer to either the entire structure or individual cranes. Furthermore, OVD is certainly not immune to common challenges faced by all object detection models, including omitted target objects (f), difficulties with small objects (g), incorrect classification (h), and bounding box coordinates (i). Fortunately, LMM can assist in identifying these errors. Our GPT-4o-based pseudo-label review detects 5 out of 6 instances of errors in the last two rows of \autoref{fig:gdino_err}, demonstrating both the necessity and efficacy of the proposed LMM-based review.

Finally, we visually compare object detection results with and without DART in \autoref{fig:predictions_1}. The first column in the figure visualizes ground truth (labeled by Grounding DINO and approved by GPT-4o), while the other two showcase prediction results of a YOLOv8n model trained with and without DART. Each row picks an image from the test set, which is never seen by the object detector in either scenario. Clearly, without DART's customized diversification and annotation, the model performance remains poor even after fine-tuning on LVIS. On the contrary, introducing DART leads to a substantial performance boost, yielding almost identical results to the ground truth. This aligns with the dramatic AP increase from 0.064 to 0.832 discussed in \autoref{sec:yolo_exp}, demonstrating the necessity and effectiveness of DART.

In the last row of \autoref{fig:predictions_1}, we demonstrate an even more intriguing benefit of DART: the final YOLO model of DART can capture missed objects during the bounding box annotation process. We argue that the YOLO model benefits from the diversity of poses and backgrounds for the target object category provided by DART, allowing it to better learn the target object’s concept. Conversely, Grounding DINO is limited by the extensive range of its pre-trained public dataset, making it sometimes less effective in specialized detection tasks. This suggests a new approach for the annotation stage: fine-tuning OVD with manually labeled target objects before using it for bounding box generation. We leave this idea for future exploration. On the other hand, the fact that training a tiny real-time object detection model under the DART framework without any manual labeling and extra data collection leads to rectifying errors made by the state-of-the-art OVD model highlights the effectiveness of the DART pipeline. Despite the imperfections of existing OVD models, DART's robust framework compensates for their shortcomings.

\section{Limitaion}
\label{sec:limitation}
While our DART pipeline demonstrates significant performance, boosting the AP of an already fine-tuned YOLOv8n model from 0.064 to 0.832 without any manual labeling, it is still constrained by weaknesses in the current instantiations of each module. Grounding DINO's issues are illustrated and discussed in \autoref{sec:vis}. The Dreambooth with SDXL for data diversification generates objects that closely mimic originals in various poses and scenes but still suffer from slight dissimilarity and hallucination, which result in initially unrealistic generations and, eventually, the necessity of original data during training. For images disapproved by pseudo-label review, human intervention is still preferred to indiscriminate deletion, which leads to an inevitable loss of information. Compared to proprietary LMMs such as GPT-4o, open-source models like InternVL perform suboptimally and can only handle simple scenarios. A few categories show minimal improvement after diversification, primarily due to inadequate labeling of test set data and deficiencies noted above.

\section{Conclusion}
\label{sec:conclusion}
In this paper, we introduce \textbf{DART}, an automated end-to-end object detection pipeline to eliminate the need for human labeling and extensive data collection while maintaining high model performance. The Liebherr Product (LP) dataset with 15K images across 23 categories of construction machines is collected to validate the effectiveness of our approach. We have demonstrated the efficacy of the DART pipeline through extensive experimentation and analysis. Our results consistently show that incremental incorporation of each component of DART always contributes to significant improvements in AP. Detailed quantitative ablation studies, coupled with qualitative analyses, reveal the optimal configuration of each stage. Notably, our researches indicate that a 3:1 ratio of generated (by fine-tuned SDXL models via DreamBooth) to original data yields the best results for data diversification on the LP dataset. Open-vocabulary bounding box annotation (Grounding DINO) has proven effective in generating credible bounding boxes without human intervention. GPT-4o has demonstrated human-level semantic understanding capabilities in the task of pseudo-label review. Furthermore, our analysis reveals that YOLOv8n offers the best tradeoff between accuracy and model size, making it the optimal choice for real-time object detection on edge devices. These findings demonstrate DART’s superior efficacy in enhancing the capabilities of real-time object detection models. In addition to its effectiveness, the modularity of DART ensures its adaptability to future advancements and customization for any specific target object or environment. Remarkably, DART accomplishes these results without requiring any manual annotation and extra data collection, further highlighting its significance. The efficiency and automation of DART not only streamline object detection applications but also reveal its potential to revolutionize the field.

\appendix
\section{More model details}
\label{app:imp}
In this section, we discuss additional implementation details of the core modules of DART.

\subsection{DreamBooth}
\label{app:dreambooth}

The hyperparameters used for the Dreambooth with SDXL model are detailed in \autoref{tab:hp_dreambooth}. We use the official SDXL checkpoint from Stability AI but replace the VAE with madebyollin's version (for a fair comparison of FP16 and BF16, where we later find no apparent differences between the two) to initialize the SDXL model. The initialized SDXL model is then trained using LoRA. We choose the AdamW optimizer with a learning rate $1 \times 10^{-4}$ for U-Net and $5 \times 10^{-6}$ for the text encoder. The learning rate schedule is constant, with no warmup steps. We implement dynamic training steps by adjusting the maximum training steps based on the number of images collected for the corresponding instance. Specifically, two models are trained for each instance with maximum training steps set to either 120 or 140 times the actual number of images collected for the corresponding instance during preprocessing. For weighting, the prior preservation loss's weight is set to 1 and the SNR-gamma to 5 for all experiments.

\begin{table}[!h]
\centering
\caption{Hyperparameters of Dreambooth with SDXL.}
\label{tab:hp_dreambooth}
\begin{tabular}{cc}
\toprule
\textbf{Hyperparameters }   &   \textbf{Values} \\
\midrule
pre-trained SDXL model          &   stabilityai/stable-diffusion-xl-base-1.0    \\
pre-trained VAE model           &   madebyollin/sdxl-vae-fp16-fix               \\
optimizer                       &   AdamW   \\
$\beta_1$                       &   0.9 \\
$\beta_2$                       &   0.999   \\
weight decay                    &   $1 \times 10^{-4}$  \\
mixed precision                 &   BF16    \\
resolution                      &   1024 $\times$ 1024  \\
batch size                      &   1                                           \\  
learning rate for U-Net         &   $1 \times 10^{-4}$                          \\
learning rate for text encoder  &   $5 \times 10^{-6}$                          \\  
learning rate schedule          &   constant                                    \\
learning rate warmup steps      &   0                                         \\
training steps                  &   $\max(800, \{120,140\}\times \text{\#images})$     \\
gradient accumulation steps     &   1   \\
prior preservation loss weight  &   1                                           \\
SNR-gamma                       &   5                                           \\
\bottomrule
\end{tabular}
\end{table}

\begin{figure}[!h]
    \centering
    \includegraphics[width=0.75\columnwidth]{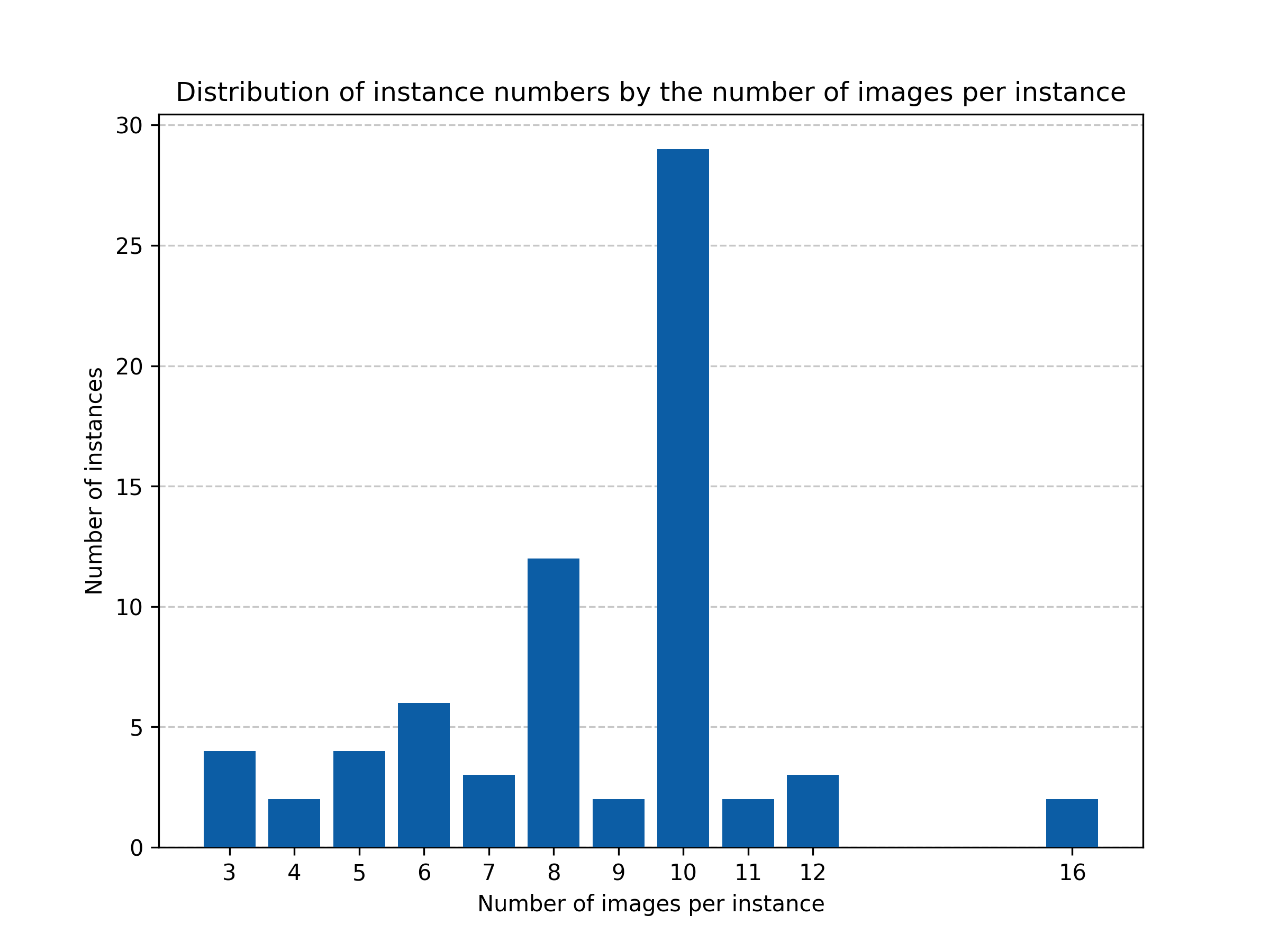}
    \caption{Distribution of instance numbers by the number of images per instance.}
    \label{fig:instance}
\end{figure}

\autoref{fig:instance} illustrates the distribution of the number of instances by the number of images per instance. The minimum number of images per instance is set to 3 (as suggested by \cite{ruizDreamBoothFineTuning2023}), while the maximum is 16. We do not find a significant correlation between the number of images per instance and the quality of the trained SDXL via DreamBooth. Most instances contain approximately 10 images each, which we empirically find to be a suitable choice.

\subsection{LMM}
\label{app:lmm}
Besides GPT-4o, we have also tried GPT-4v with the "gpt-4-vision-preview" API and observed that GPT-4v is particularly strict when handling bounding boxes, which leads to the "relaxation" and "reminder" section of the user prompt listed in \autoref{tab:prompt_gpt}. Additionally, we find that if the prompt is too lengthy, all LMMs, except for GPT-4o, experience a decline in performance. Therefore, we deliberately shorten the prompt for InternVL-1.5, as shown in \autoref{tab:prompt_internvl}. For GPT-4v, we discover a workaround for this issue: using multi-turn dialogues, where we first ask the questions (which corresponds to "task," "question," and "output" section in \autoref{tab:prompt_gpt}) and then give suggestions ("relaxation" and "reminder" section in \autoref{tab:prompt_gpt})) in the second turn. This approach helps avoid overly long prompts and improves model performance. In contrast, GPT-4o doesn't suffer from long prompts and dominates its predecessor in the bounding box review task. Therefore, we exclusively use it for the pseudo-label review segment of our DART pipeline.

\subsection{YOLO}
\label{app:yolo}

\autoref{tab:hp_yolo} presents hyperparameters used for YOLO models. We follow common practice \cite{YOLOv5Ultralytics2020,liYOLOv6V3FullScale2023,liYOLOv6SingleStageObject2022,wangYOLOv7TrainableBagoffreebies2022,jocherUltralyticsYOLOv82023,wangYOLOv9LearningWhat2024,wangYOLOv10RealTimeEndtoEnd2024} to resize all images to 640$\times$640. In almost all experiments, training is conducted over 60 epochs since model performance plateaus after 20-30 epochs. AdamW optimizer is always employed with default beta settings and a weight decay set to 5$\times$10$^{-4}$. For each experiment setup and model type, we meticulously fine-tune learning-rate-related parameters and only display the optimal hyperparameter sets derived from the experiment setup of the optimal generated-to-original data ratio (3:1) in \autoref{tab:hp_yolo}. It turns out that the optimal learning rates range from 7e-5 to 5e-4, with a general trend indicating that larger models and bigger datasets benefit from lower learning rates. The final learning rate factor is always 50\% at best. The learning rate schedule is either cosine or linear, depending on the model. Warm-up epochs are set between 1 and 4, with a warm-up momentum of 0.8 and a warm-up bias learning rate of 0.1. The box loss weight is set to 7.5, the class loss weight to 0.5, and the DFL loss weight to 1.5. HSV augmentations are specified by saturation with 0.7, value with 0.4, and hue with 0.015. Mosaic augmentation is always applied (with probability 100\%) except for the last 10 epochs when it's turned off for better alignment. Additional parameters include translation at 0.1 fraction of total width and height, horizontal flip rate at 0.5, and scale at 0.5 (resizing between 50\% and 150\%).

\begin{table}[!h]
\caption{Hyperparameters of YOLOv8 and YOLOv10 models.}
\label{tab:hp_yolo}
\small
\centering
\begin{tabular}{cc}
    \toprule
    \textbf{Hyperparameter} & \textbf{YOLOv8n/v8s/v10n/v10s} \\
    \midrule
    epochs & 60 \\
    optimizer                       &   AdamW   \\
    $\beta_1$                       &   0.9 \\
    $\beta_2$                       &   0.999   \\
    weight decay & 5$\times$10$^{-4}$ \\
    mixed precision                 &   FP16    \\
    resolution                      &   640 $\times$ 640  \\    
    initial learning rate & 27/7.3/50/7.0$\times$10$^{-5}$ \\
    final learning rate factor & 0.5 \\
    batch size                      &   64/32/32/32    \\  
    learning rate schedule & cosine/cosine/cosine/linear \\
    warm-up epochs & 4/1/3/4 \\
    warm-up momentum & 0.8 \\
    warm-up bias learning rate & 0.1 \\
    box loss weight & 7.5 \\
    class loss weight & 0.5 \\
    DFL loss weight & 1.5  \\
    HSV saturation  & 0.7 \\
    HSV value  & 0.4 \\
    HSV hue  & 0.015 \\ 
    translation  & 0.1 \\
    flip probability    & 0.5 \\
    scale  & 0.5 \\
    mosaic probability  & 1.0 \\
    close mosaic epochs  &   10 \\
    \bottomrule
\end{tabular}
\vspace{-10pt}
\end{table}

\section{Prompt engineering}
\label{app:prompt}

Prompt engineering is critical for the performance of three out of four major components of our proposed DART pipeline. In this section, we provide details about how the prompts are constructed for each component.

\subsection{DreamBooth}
\label{app:prompt_dreambooth}

\autoref{tab:prompts_dreambooth} displays all the prompts used for inference in DreamBooth with SDXL to diversify our dataset. For each prompt, we provide an overall description, the exact prompt text, and its respective ID. Prompts 1-48 are general prompts applied across all classes. Prompts 49-57 are specific to large land-based machinery, such as mining equipment, while prompts 58-67 are tailored for water-related machines, such as maritime cranes.

\begin{longtable}[!h]{|p{0.03\textwidth}|p{0.22\textwidth}|p{0.72\textwidth}|} 
\caption{A list of all prompts used during data generation via trained SDXL by DreamBooth. The "\{placeholder\}" refers to the specific instance name wrapped by "<>" followed by its class name, e.g. "<TA230> articulated dump truck" for the case in \autoref{fig:dreambooth}.}
\label{tab:prompts_dreambooth} \\
\hline
\textbf{\#} & \textbf{Description} & \textbf{Prompt text} \\
\hline
\endfirsthead
\multicolumn{3}{c}%
{{\tablename\ \thetable{} -- continued from previous page}} \\
\hline
\textbf{ID} & \textbf{Description} & \textbf{Prompt text} \\
\hline
\endhead
\hline
\endfoot
\hline
\endlastfoot
1 &  construction site & A photo of a \{placeholder\} on a construction site. The image is high quality and photorealistic. The \{placeholder\} may be partially visible, at a distance, or obscured. The background is complex, providing a realistic context. \\
\hline
2 & Sunny\newline construction site & A high-quality, photorealistic image of a \{placeholder\} under a bright, sunny sky at a bustling construction site. \\
\hline
3 & Cloudy\newline construction site & A detailed, photorealistic image of a \{placeholder\} operating at a construction site on a cloudy day. \\
\hline
4 & Rainy\newline  construction site & A high-resolution, photorealistic image of a \{placeholder\} working at a muddy construction site during heavy rain. \\
\hline
5 & City street & A photorealistic image of a \{placeholder\} on a busy city street, surrounded by tall buildings and traffic. \\
\hline
6 & Rural area & A high-quality, photorealistic image of a \{placeholder\} operating in a rural area, with fields and farmhouses in the background. \\
\hline
7 & Mining site & A detailed, photorealistic image of a \{placeholder\} working at a mining site, with rocky terrain and machinery in the background. \\
\hline
8 & Harbor & A high-resolution, photorealistic image of a \{placeholder\} operating at a harbor, with ships and cranes in the background. \\
\hline
9 & Left side & A photorealistic image of a \{placeholder\} positioned on the left side of the frame at a construction site, with workers in the background. \\
\hline
10 & Right side & A high-quality, photorealistic image of a \{placeholder\} on the right side of the image, operating in a busy urban environment. \\
\hline
11 & In the distance & A detailed, photorealistic image of a \{placeholder\} visible in the distance at a construction site, with a panoramic view of the area. \\
\hline
12 & Facing left & A high-resolution, photorealistic image of a \{placeholder\} facing left, working at a construction site with scaffolding in the background. \\
\hline
13 & Facing right & A photorealistic image of a \{placeholder\} facing right, operating in a rural setting with rolling hills in the background. \\
\hline
14 & Partially visible & A high-quality, photorealistic image of a \{placeholder\} partially visible behind a building at a construction site. \\
\hline
15 & Nighttime\newline  construction & A photorealistic image of a \{placeholder\} working at night under artificial lights at a construction site. \\
\hline
16 & Foggy morning & A detailed, photorealistic image of a \{placeholder\} operating on a foggy morning, with limited visibility at a rural construction site. \\
\hline
17 & Snowy day & A high-resolution, photorealistic image of a \{placeholder\} working at a construction site during snowfall, with snow-covered equipment and ground. \\
\hline
18 & Highway\newline  construction & A photorealistic image of a \{placeholder\} operating on a highway under construction, with traffic cones and barriers. \\
\hline
19 & Desert construction & A high-quality, photorealistic image of a \{placeholder\} working at a construction site in a desert area, with sand dunes in the background. \\
\hline
20 & Urban demolition & A high-resolution, photorealistic image of a \{placeholder\} demolishing an old building in an urban area, with debris flying. \\
\hline
21 & Windy day & A high-quality, photorealistic image of a \{placeholder\} working at a construction site on a windy day, with dust and debris blowing in the wind. \\
\hline
22 & Industrial area & A detailed, photorealistic image of a \{placeholder\} operating in an industrial area, surrounded by factories and heavy machinery. \\
\hline
23 & Riverbank\newline construction & A high-resolution, photorealistic image of a \{placeholder\} working on a riverbank, with water and vegetation in the background. \\
\hline
24 & Urban parking lot & A photorealistic image of a \{placeholder\} operating in an urban parking lot construction site, with buildings and cars nearby. \\
\hline
25 & Dam construction & A high-quality, photorealistic image of \{placeholder\} working on a dam construction project, with water and concrete structures in the background. \\
\hline
26 & Railway\newline  construction & A detailed, photorealistic image of a \{placeholder\} working on a railway construction site, with tracks and trains in the background. \\
\hline
27 & Solar farm\newline  construction & A high-resolution, photorealistic image of a \{placeholder\} operating at a solar farm construction site, with solar panels and wide open fields. \\
\hline
28 & Skyscraper\newline construction & A photorealistic image of a \{placeholder\} working at a skyscraper construction site, with towering steel structures and cranes. \\
\hline
29 & Wind farm\newline construction & A high-quality, photorealistic image of a \{placeholder\} operating at a wind farm construction site, with wind turbines in the background. \\
\hline
30 & Bridge\newline construction & A detailed, photorealistic image of a \{placeholder\} working on a bridge construction site over a river. \\
\hline
31 & Mountain\newline construction & A high-quality, photorealistic image of a \{placeholder\} operating on a steep mountain construction site with rocky terrain. \\
\hline
32 & Forest\newline construction & A photorealistic image of a \{placeholder\} working in a forest area, with tall trees and underbrush surrounding the site. \\
\hline
33 & Coastal\newline construction & A detailed, photorealistic image of a \{placeholder\} operating at a coastal construction site, with the ocean and waves in the background. \\
\hline
34 & Underground\newline construction & A high-resolution, photorealistic image of a \{placeholder\} working in an underground tunnel construction site, with dim lighting and rocky walls. \\
\hline
35 & Large scale\newline construction & A high-quality, photorealistic image of multiple \{placeholder\}s working together on a large-scale construction project, with cranes and scaffolding. \\
\hline
36 & Airport\newline construction & A photorealistic image of a \{placeholder\} operating at an airport construction site, with planes and runways in the background. \\
\hline
37 & Suburban\newline construction & A detailed, photorealistic image of a \{placeholder\} working in a suburban area, with houses and trees around the construction site. \\
\hline
38 & Island construction & A high-resolution, photorealistic image of a \{placeholder\} operating on an island construction site, surrounded by water and palm trees. \\
\hline
39 & Urban renewal & A high-quality, photorealistic image of a \{placeholder\} working on an urban renewal project, surrounded by modern buildings and greenery. \\
\hline
40 & Industrial site & A photorealistic image of a \{placeholder\} operating at an industrial construction site, with factories and smokestacks in the background. \\
\hline
41 & Multiple machines sunny & A high-quality, photorealistic image of several \{placeholder\}s working together under a bright, sunny sky at a bustling construction site. \\
\hline
42 & Multiple machines rainy & A high-resolution, photorealistic image of several \{placeholder\}s operating at a muddy construction site during heavy rain. \\
\hline
43 & Multiple machines city street & A photorealistic image of multiple \{placeholder\}s on a busy city street, surrounded by tall buildings and traffic. \\
\hline
44 & Multiple machines rural area & A high-quality, photorealistic image of several \{placeholder\}s operating in a rural area, with fields and farmhouses in the background. \\
\hline
45 & Multiple machines harbor & A high-resolution, photorealistic image of several \{placeholder\}s operating at a harbor, with ships and cranes in the background. \\
\hline
46 & Multiple machines night & A photorealistic image of several \{placeholder\}s working at night under artificial lights at a construction site. \\
\hline
47 & Multiple machines snowy day & A high-resolution, photorealistic image of several \{placeholder\}s working at a construction site during snowfall, with snow-covered equipment and ground. \\
\hline
48 & Multiple machines highway & A photorealistic image of several \{placeholder\}s operating on a highway under construction, with traffic cones and barriers. \\
\hline
49 & Open pit mine & A high-quality, photorealistic image of a \{placeholder\} working in an open-pit mine, with terraced rock formations and heavy machinery around. \\
\hline
50 & Underground mine & A detailed, photorealistic image of a \{placeholder\} operating in an underground mine, with tunnels and mining equipment. \\
\hline
51 & Mining town & A high-resolution, photorealistic image of a \{placeholder\} working near a mining town, with workers' housing and facilities in the background. \\
\hline
52 & Quarry operation & A photorealistic image of a \{placeholder\} working in a quarry, with large stone blocks and cutting equipment. \\
\hline
53 & Mountain mining & A high-quality, photorealistic image of a \{placeholder\} operating in a mountain mining site, with steep slopes and rocky terrain. \\
\hline
54 & Remote mine & A photorealistic image of a \{placeholder\} working in a remote mining location, with rugged terrain and minimal infrastructure. \\
\hline
55 & Ore processing & A high-quality, photorealistic image of a \{placeholder\} involved in ore processing, with conveyor belts and sorting equipment. \\
\hline
56 & Tailings dam & A detailed, photorealistic image of a \{placeholder\} working near a tailings dam, with water and waste materials. \\
\hline
57 & Mining reclamation & A high-resolution, photorealistic image of a \{placeholder\} involved in mining reclamation, with land restoration and vegetation. \\
\hline
58 & Ocean dockyard & A high-quality, photorealistic image of a \{placeholder\} operating in an ocean dockyard, with large ships and containers in the background. \\
\hline
59 & River dredging & A detailed, photorealistic image of a \{placeholder\} dredging a river, with muddy water and vegetation along the banks. \\
\hline
60 & Coastal\newline maintenance & A high-resolution, photorealistic image of a \{placeholder\} performing coastal maintenance, with waves crashing in the background. \\
\hline
61 & Offshore\newline construction & A photorealistic image of a \{placeholder\} working on an offshore construction project, with deep sea and platform structures. \\
\hline
62 & Flood control & A high-quality, photorealistic image of a \{placeholder\} engaged in flood control operations, with sandbags and rising water levels. \\
\hline
63 & Canal work & A detailed, photorealistic image of a \{placeholder\} working on a canal, with narrow waterways and boats passing by. \\
\hline
64 & Shipyard repair & A photorealistic image of a \{placeholder\} in a shipyard, repairing large vessels with workers around. \\
\hline
65 & Waterfront park & A high-quality, photorealistic image of a \{placeholder\} operating in a waterfront park, with recreational areas and people nearby. \\
\hline
66 & Marine research & A detailed, photorealistic image of a \{placeholder\} assisting in marine research activities, with scientists and research equipment around. \\
\hline
67 & Dockside assembly & A photorealistic image of a \{placeholder\} involved in dockside assembly of marine structures, with cranes and construction materials. \\
\hline
\end{longtable}

\subsection{Grounding DINO}
\label{app:prompt_gdino}

\autoref{tab:synonyms} presents the synonyms (and superclasses, combined as one column) for each category in the LP dataset. The synonyms are used to construct specific prompts during the annotation phase. The detailed procedure is described in \autoref{sec:gdino}.

\begin{longtable}[!h]{|p{0.4\textwidth}|p{0.5\textwidth}|}
\caption{A list of synonyms (including superclasses) for each class in LP dataset.}
\label{tab:synonyms} \\
\hline
\textbf{Original Class} & \textbf{Synonyms} \\
\hline
\endfirsthead
\multicolumn{2}{c}%
{{\tablename\ \thetable{} -- continued from previous page}} \\
\hline
\textbf{Original class name} & \textbf{Synonyms} \\
\hline
\endhead
\endfoot
\hline
\endlastfoot

articulated dump truck & N/A \\ \hline
bulldozer & dozer, crawler tractor \\ \hline
crawler crane & track crane, crane \\ \hline
crawler excavator & track excavator, excavator \\ \hline
crawler loader & track loader \\ \hline
combined piling and drilling rig & drilling rig, piling rig \\ \hline
duty cycle crane & dragline, dragline excavator \\ \hline
gantry crane & container crane, maritime crane \\ \hline
log loader & log handling machine \\ \hline
maritime crane & harbor crane, port crane, crane \\ \hline
material handling machine & material handler \\ \hline
mining bulldozer & N/A \\ \hline
mining excavator & N/A \\ \hline
mining truck &  N/A\\ \hline
mobile crane & crane \\ \hline
pipelayer & sideboom \\ \hline
pontoon excavator & floating excavator, amphibious excavator, excavator \\ \hline
reachstacker & container handler, material handling equipment \\ \hline
telescopic handler & lull, telehandler, reach forklift, zoom boom, material handling equipment \\ \hline
tower crane & crane \\ \hline
truck mixer & cement mixer truck, concrete mixer truck \\ \hline
wheel loader & front end loader, bucket loader \\ \hline
wheel excavator & mobile excavator, excavator \\ \hline

\end{longtable}

\subsection{LMM}
\label{app:prompt_lmm}
GPT-4o receives detailed prompts for bounding box review. The prompts are broken into parts in \autoref{tab:prompt_gpt}, with the purposes given on the left and the exact text contents on the right. First, the system prompt briefly summarizes the bounding box review task and highlights three key metrics: precision, recall, and fit. The user prompt consists of the five remaining sections. The "task" section outlines the task and objectives in more detail, with \{target\} referring to the pre-defined class during data collection and \{secondary\_target\} indicating commonly co-occurring classes also recorded during data collection. Note that explicit mention of target types is not mandatory since the boxes in the images already have class labels, and GPT-4o exhibits excellent OCR capability. But it's still beneficial for corner cases where the label in the image is occluded. Next, the "questions" section elaborates on the specific questions corresponding to the three main criteria. The output format is specified in the "output" segment to facilitate subsequent processing. The "output" section also follows the guidelines of COT to enhance the model's performance. In our early experiments, we found that GPT-4 models tend to be overly strict in their judgments regarding the bounding box. To mitigate this, we summarize common issues in the "relaxation" and "reminder" sections and attach them to the end of the user prompt. Additionally, our tests with GPT-4v show that including the "relaxation" and "reminder" sections as part of the second round of dialogue instead of a single long text is more effective. However, GPT-4o can handle all sections of the user prompt at once.

\begin{longtable}[h]{|p{0.15\textwidth}|p{0.8\textwidth}|}
\caption{Text prompt for GPT-4o-based bounding box review. The “system” prompt in the table is used as the system prompt for the API, while the user prompt is created by sequentially concatenating the remaining parts.}
\label{tab:prompt_gpt}\\
\hline
\textbf{Description} & \textbf{Prompt text} \\
\hline
\endfirsthead
\multicolumn{2}{c}%
{{\tablename\ \thetable{} -- continued from previous page}} \\
\hline
\textbf{Description} & \textbf{Prompt text} \\
\hline
\endhead
\hline
\endfoot
\hline
\endlastfoot

system & You are an AI bounding box annotation evaluator. Your task is to evaluate the correctness of bounding box annotations for given images and target objects. The bounding boxes are directly drawn as colored rectangles on top of the image. The class label is shown at the top-left corner of the corresponding bounding box. You will evaluate each bounding box annotation based on three criteria: precision, recall, and fit. \\
\hline
task & In this image, the target object for bounding box annotation is \{target\}. There may also be \{secondary\_target\}. Each of the existing \{target\}s and \{secondary\_target\}s should be annotated by a bounding box drawn as a colored rectangle. The goal is to accurately localize all \{target\}s and \{secondary\_target\}s using these bounding boxes. Your task is to evaluate whether all bounding boxes are correct. \\
\hline
questions & Correctness should be assessed in terms of precision, recall, and fit. Specifically, consider the following questions before making your judgment: \newline
1. Does each bounding box perfectly enclose one single target object? \newline
2. Are all target objects localized by a bounding box?\newline
3. Is each bounding box neither too loose nor too tight? \\
\hline
output & Please provide your evaluation in the following JSON format:\newline
```json\{\newline
"Precision": "Yes/No answer to question 1", \newline
"Recall": "Yes/No answer to question 2", \newline
"Fit": "Yes/No answer to question 3" \newline\}\newline
Please think step-by-step and be sure to provide the correct answers. Very briefly explain yourself before answering the question. \\
\hline
relaxation & Before finalizing your evaluation, please consider the following suggestions:\newline
1. For question 1 (Precision), if an object is occluded, the bounding box should be inferred based on a reasonable estimation of the object's size.\newline
2. For question 2 (Recall), it's fairly normal to have only one object in the dataset.\newline
3. For question 3 (Fit), don't be too harsh when bounding box edges just slightly cut off the object or just enclose a little bit of the outside area. \\
\hline
reminder & Always consider my suggestions before answering questions. But the suggestions are not strict rules. You can also use your own judgment. The most important thing is to answer the three questions correctly. \\
\hline
\end{longtable}

For InternVL-1.5, concise prompts are used to review the photorealism of a generated image, avoiding unnecessary complexity and the performance degradation issue for long text. There's only a single prompt used for each image, which is listed in \autoref{tab:prompt_internvl}. Like the bounding box review, we only keep images receiving positive feedback for all questions from InternVL-1.5.

\begin{table}[!h]
\caption{Text prompt for InternVL-1.5-based image photorealism review. The "\{target\}" is a placeholder for the actual class name.}
\label{tab:prompt_internvl}
\centering
\begin{tabular}{|p{0.15\textwidth}|p{0.8\textwidth}|}
\hline
\textbf{Description} & \textbf{Prompt text} \\
\hline
User & Is this image suitable as training data object detection for \{target\}? Answer YES or NO.\newline
Does the main object in the image look like an authentic \{target\}? Answer YES or NO.
\\
\hline
\end{tabular}

\end{table}

\section{More visulizations}
\label{app:vis}

In this section, we provide additional visualization results for qualitative analysis of the DART pipeline.

\begin{figure}[!htbp]
    \centering
    \includegraphics[width=0.95\columnwidth]{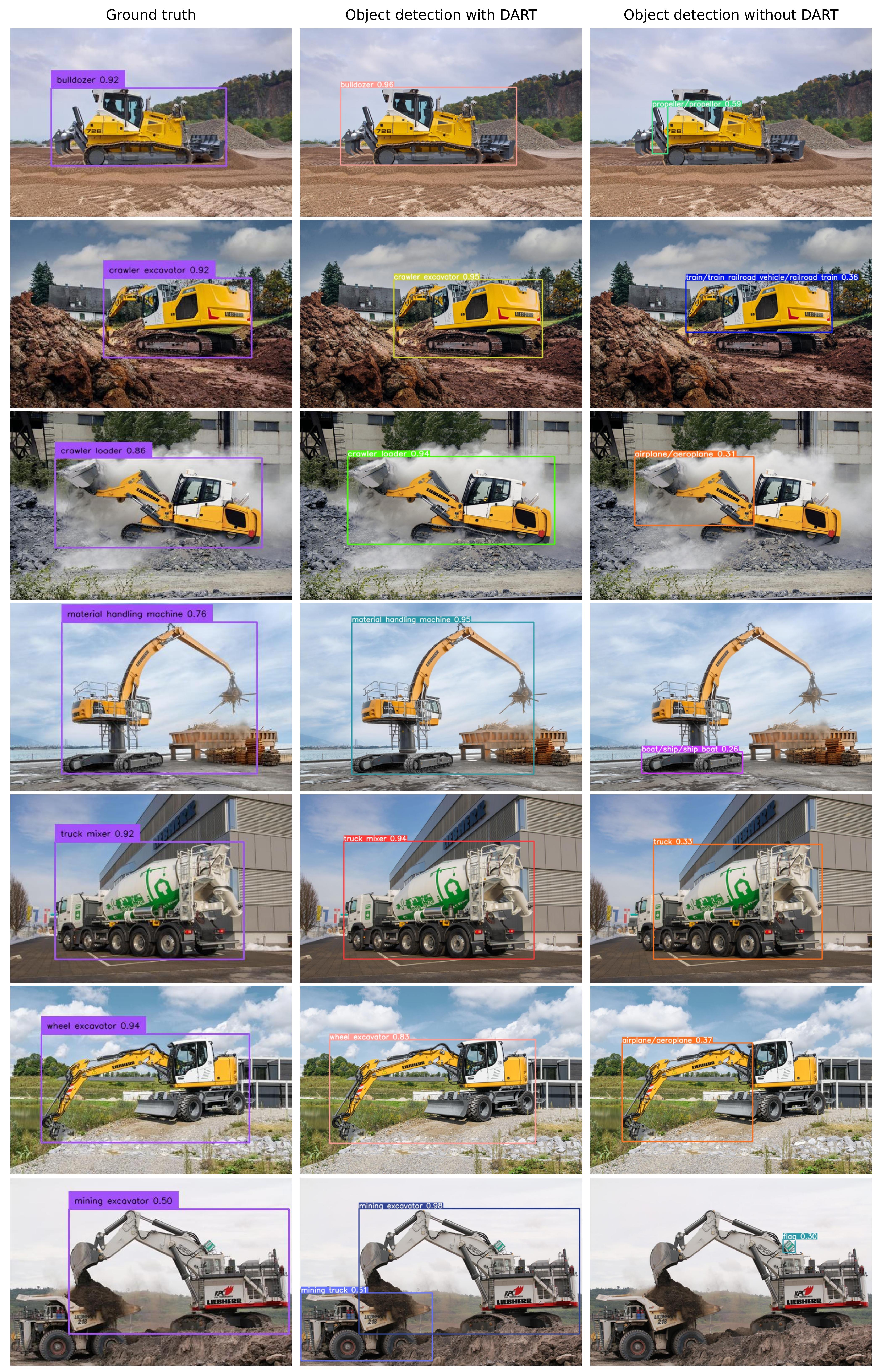}
    \caption{More comparisons of object detection results with and without DART on test set images.}
    \label{fig:predictions_2}
\end{figure}

We provide additional visual comparisons of object detection results with and without DART in \autoref{fig:predictions_2}. Same as in \autoref{fig:predictions_1}, the implementation of DART leads to a significant increase in model performance.

\begin{figure}[!hbt]
    \centering
    \setlength{\tabcolsep}{1pt}
    \renewcommand{\arraystretch}{1.0}
    {\scriptsize
    \begin{tabular}{c@{\hskip 5pt} c@{\hskip 5pt} c c c c}
    
        \begin{tabular}{c c}
            \includegraphics[width=0.092\linewidth,height=0.092\linewidth]{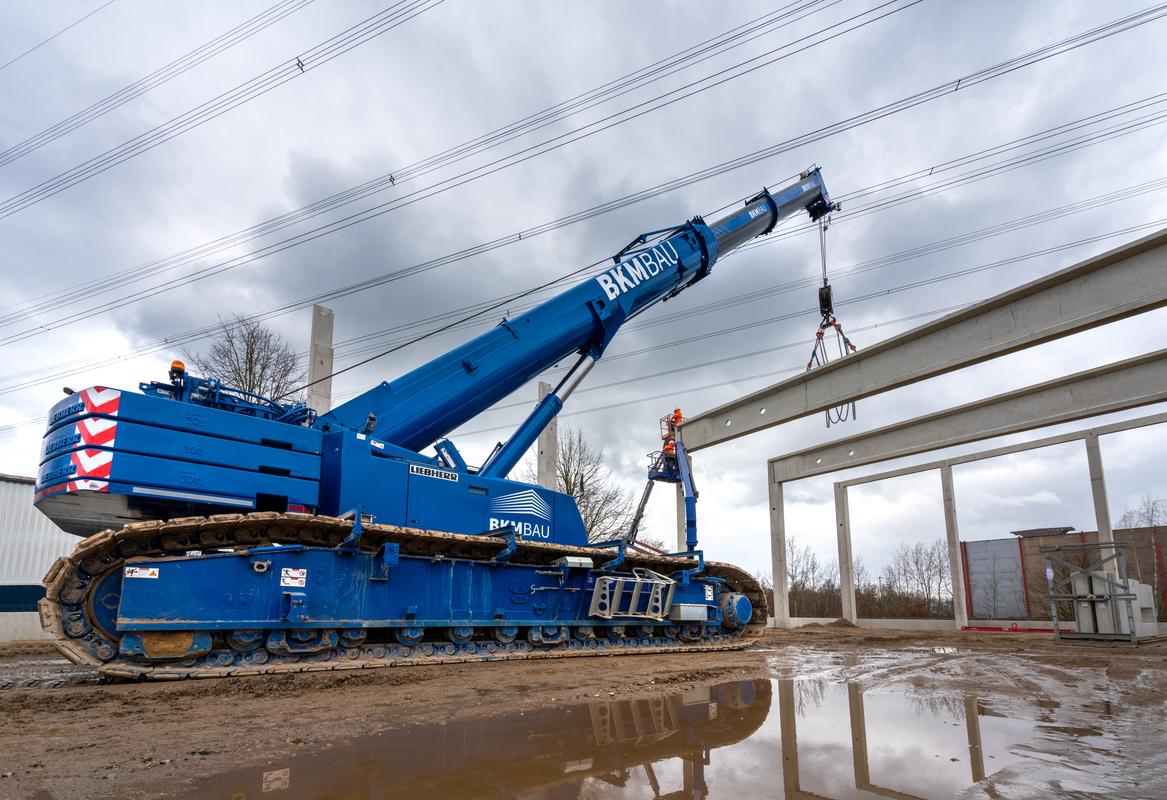} & 
            \includegraphics[width=0.092\linewidth,height=0.092\linewidth]{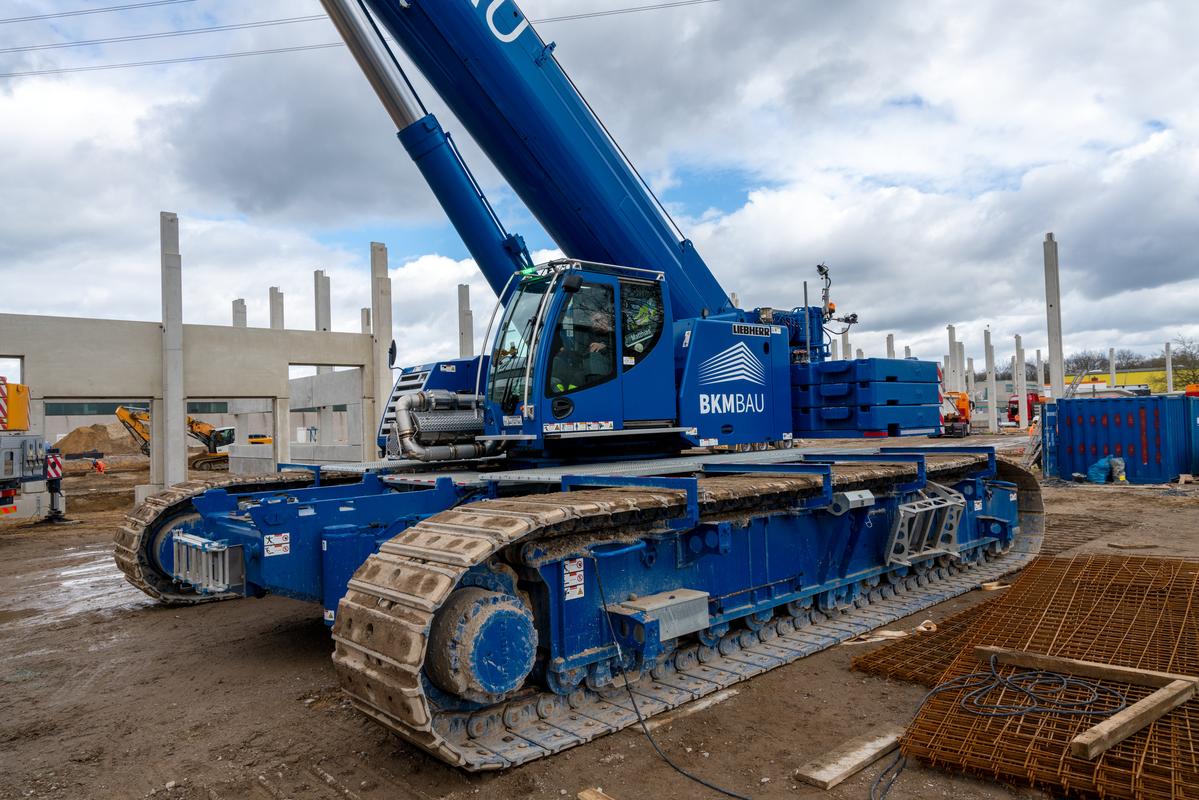} \\
            \includegraphics[width=0.092\linewidth,height=0.092\linewidth]{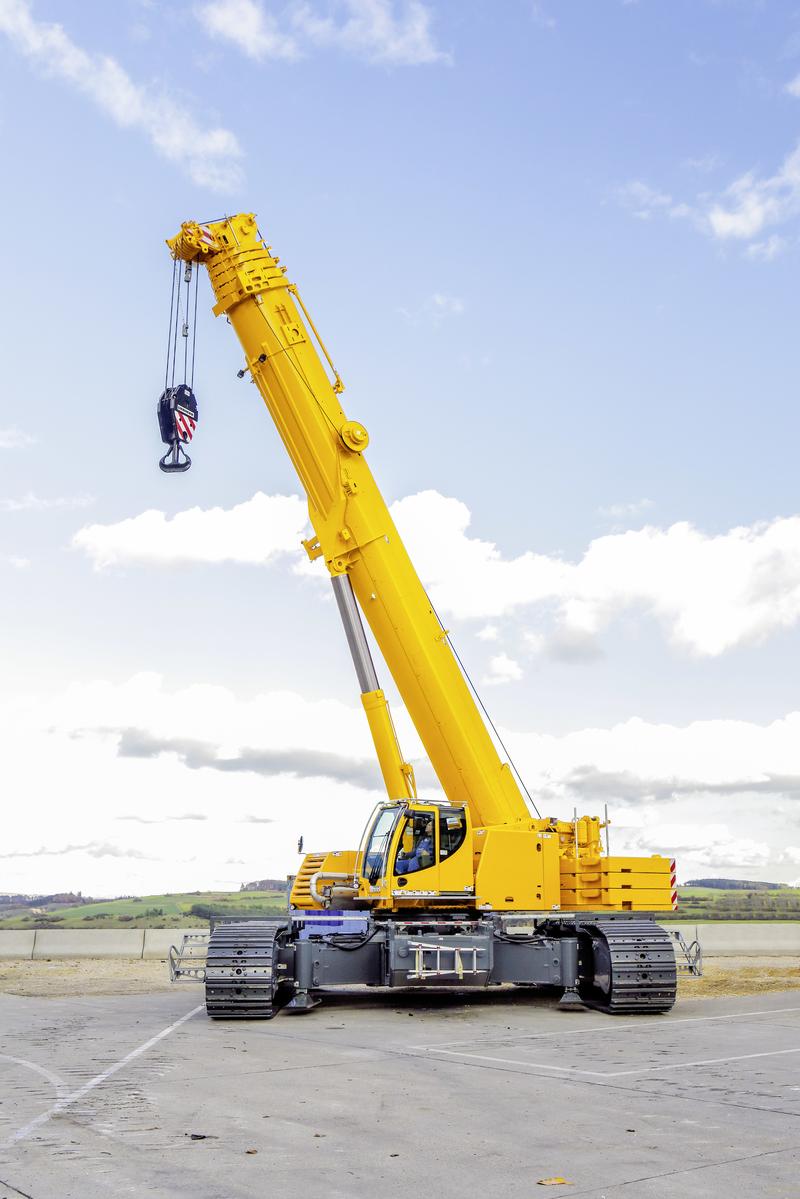} & 
            \includegraphics[width=0.092\linewidth,height=0.092\linewidth]{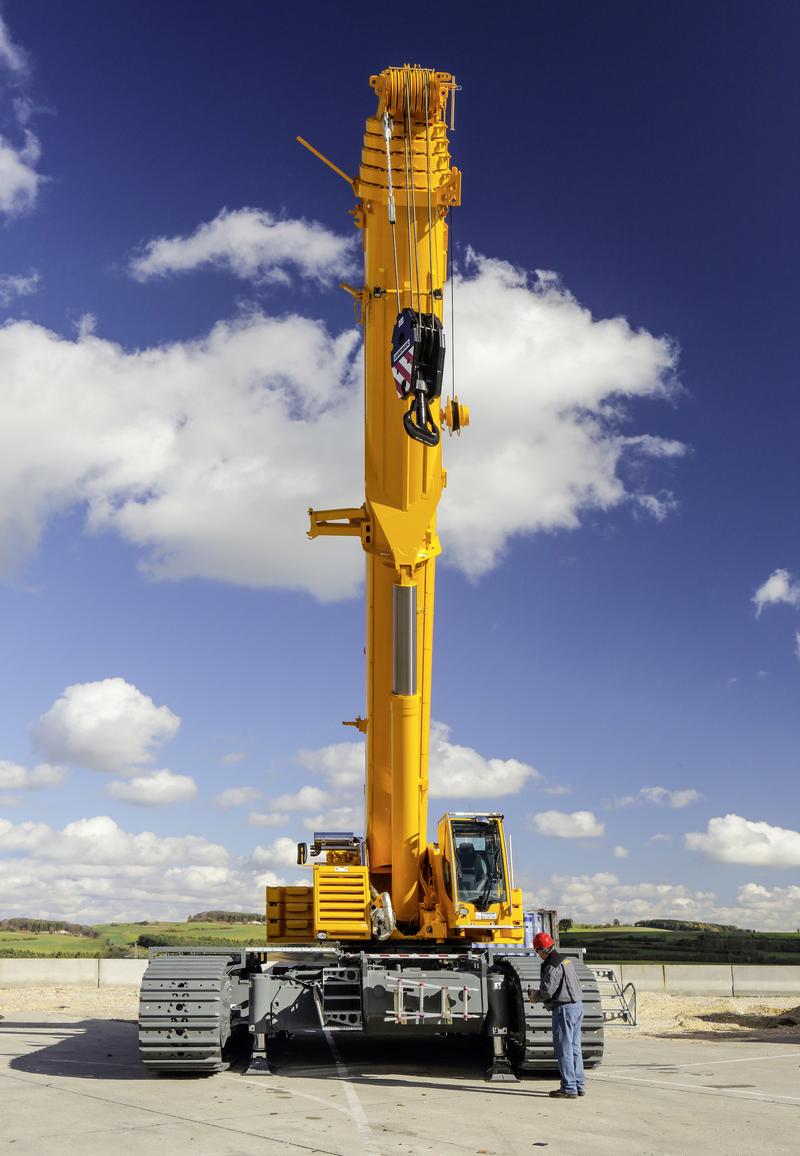}
        \end{tabular}
        &
        $\rightarrow$
        &
        \begin{tabular}{c}
        \includegraphics[width=0.184\linewidth]{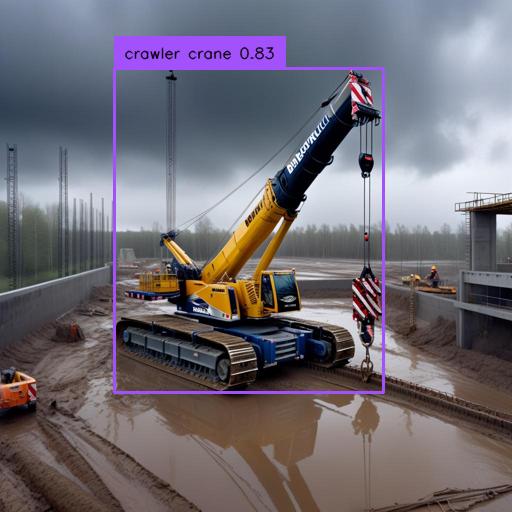}
        \end{tabular} &
        \begin{tabular}{c}
        \includegraphics[width=0.184\linewidth]{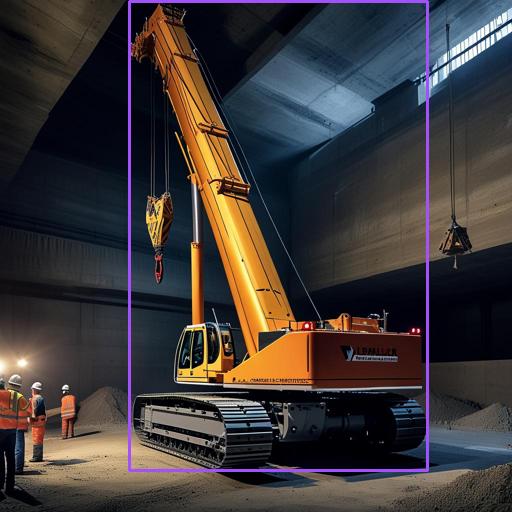} 
        \end{tabular} &
        \begin{tabular}{c}
        \includegraphics[width=0.184\linewidth]{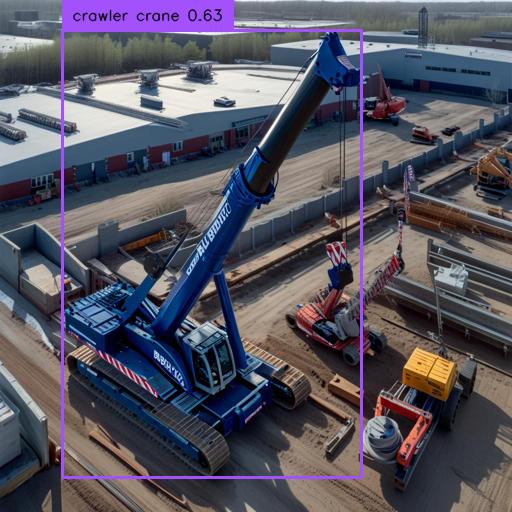} 
        \end{tabular} &
        \begin{tabular}{c}
        \includegraphics[width=0.184\linewidth]{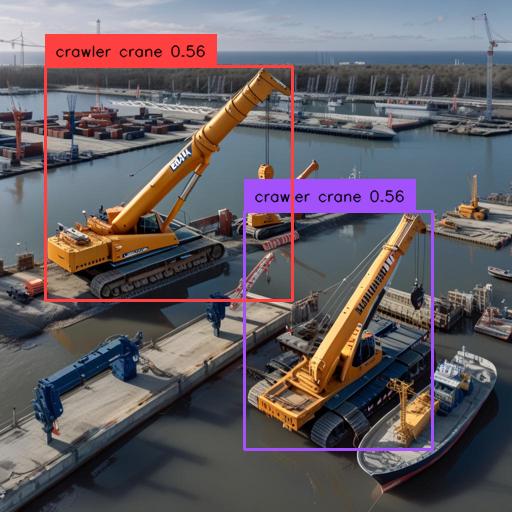}
        \end{tabular} \\
        
        {\footnotesize Instance data} & & {\begin{tabular}{c@{}c@{}c@{}c@{}} 4. Rainy\\construction site \end{tabular}} & {\begin{tabular}{c@{}c@{}c@{}c@{}} 34. Underground\\construction site \end{tabular}} & {\begin{tabular}{c@{}c@{}c@{}c@{}} 22. Industrial\\area \end{tabular}} & {\begin{tabular}{c@{}c@{}c@{}c@{}} 45. Multiple\\machines harbor \end{tabular}} \\ \\
        
        \begin{tabular}{c c}
            \includegraphics[width=0.092\linewidth,height=0.092\linewidth]{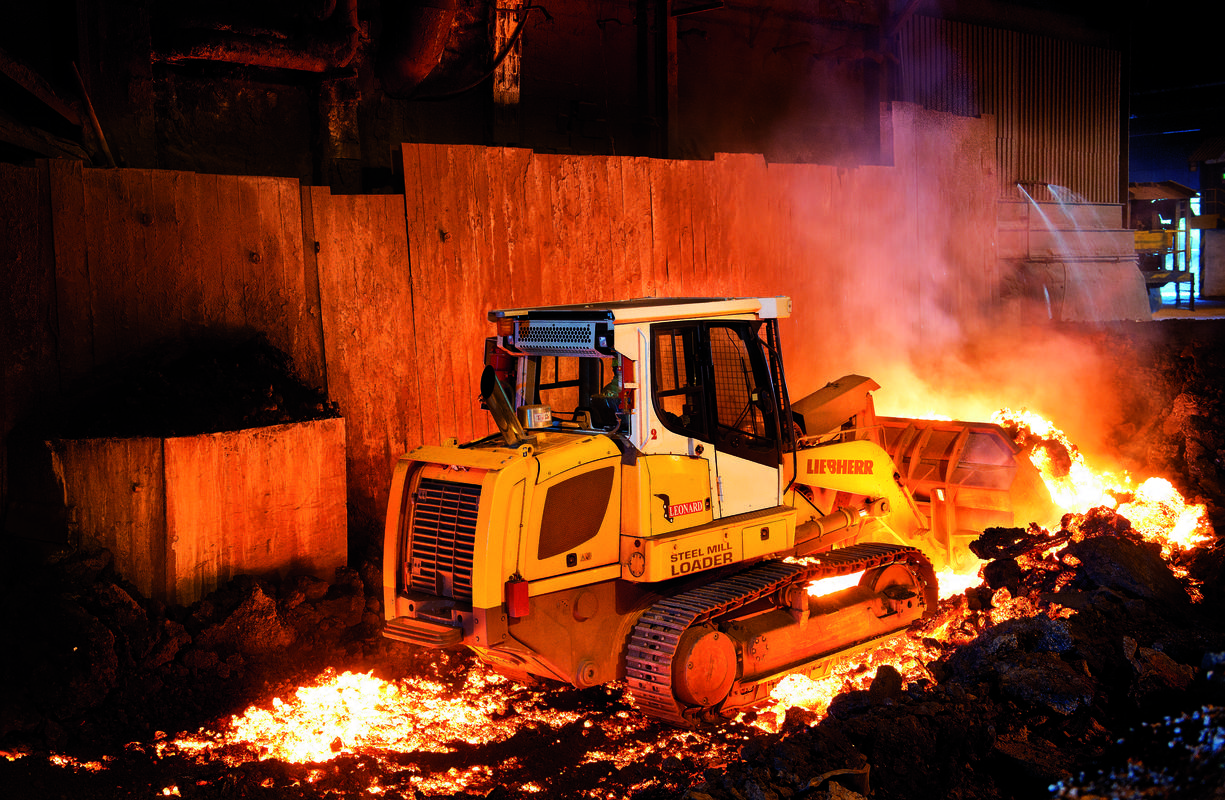} & 
            \includegraphics[width=0.092\linewidth,height=0.092\linewidth]{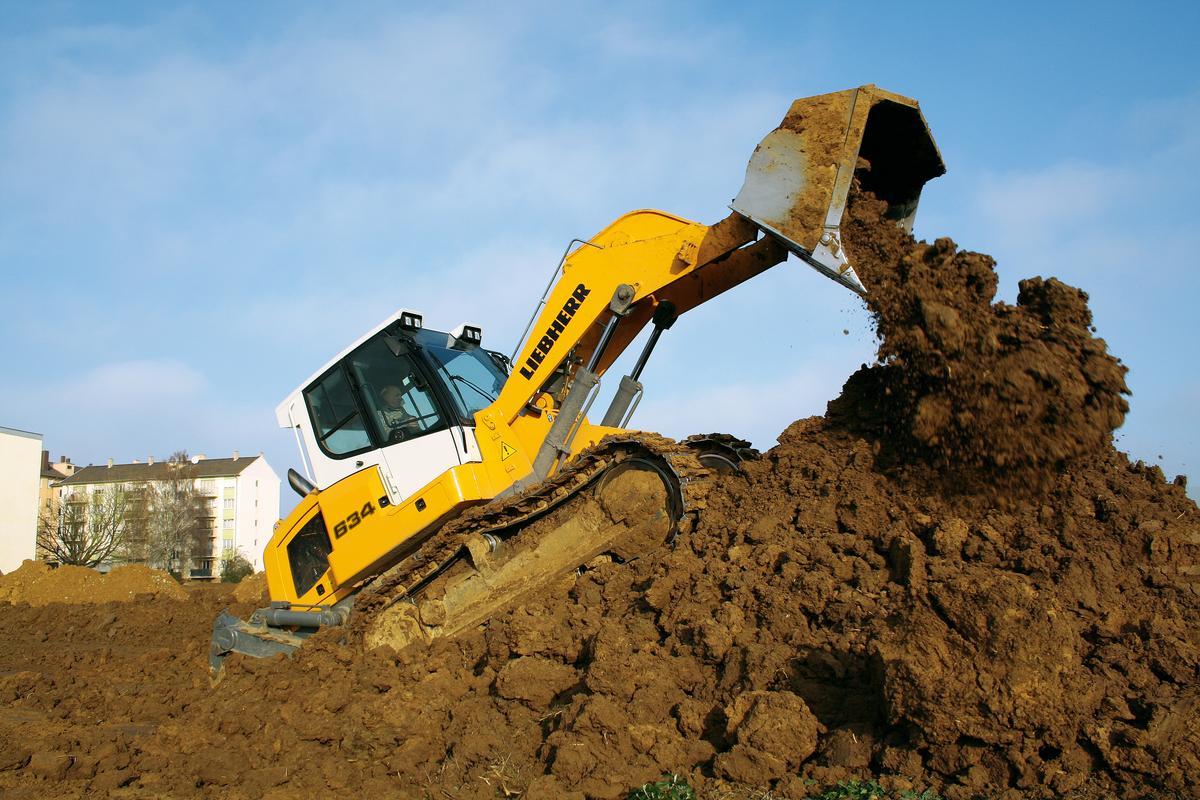} \\
            \includegraphics[width=0.092\linewidth,height=0.092\linewidth]{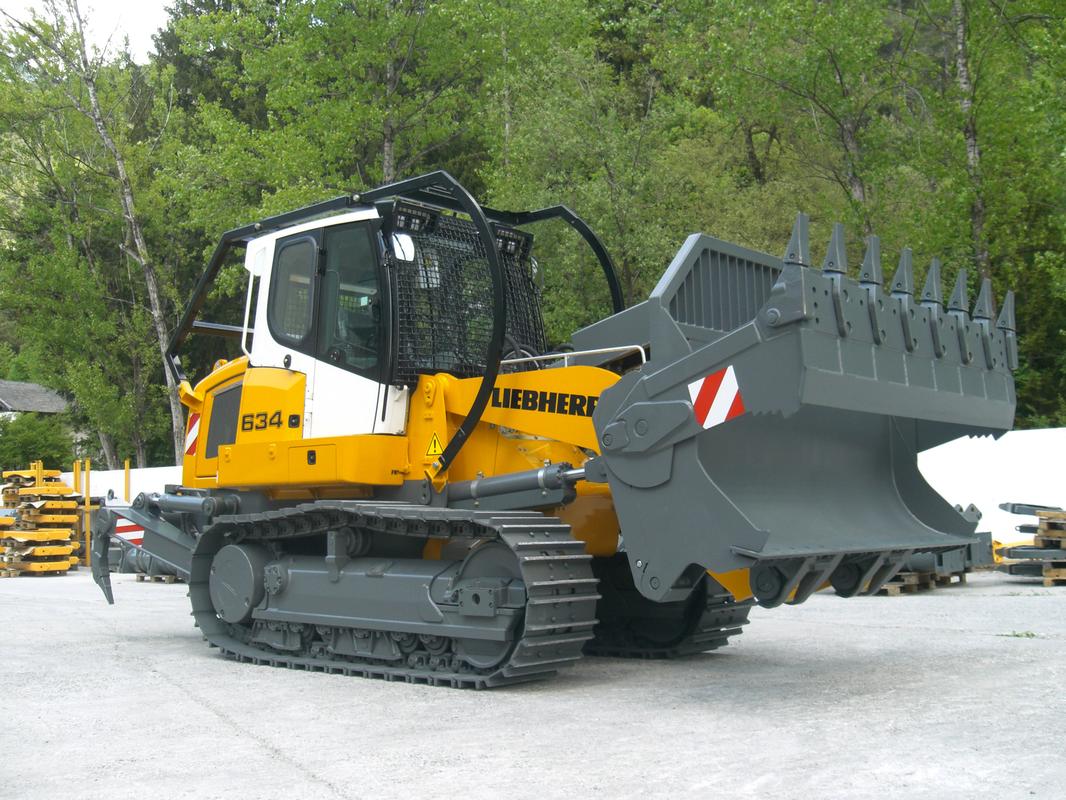} & 
            \includegraphics[width=0.092\linewidth,height=0.092\linewidth]{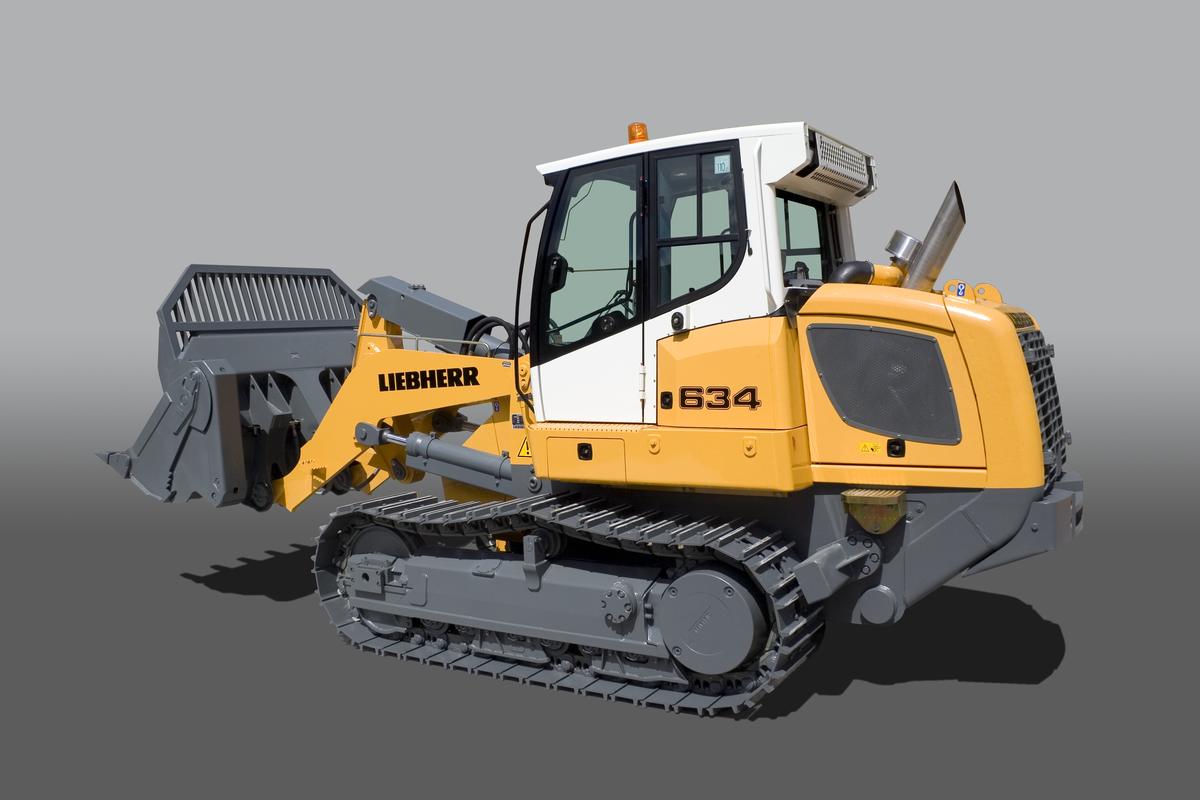}
        \end{tabular}
        &
        $\rightarrow$
        &
        \begin{tabular}{c}
        \includegraphics[width=0.184\linewidth]{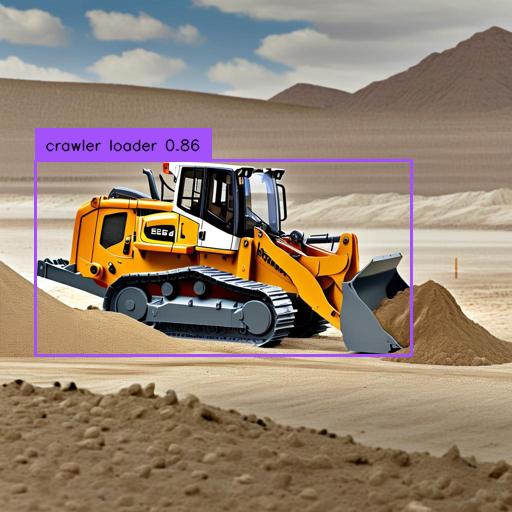}
        \end{tabular} &
        \begin{tabular}{c}
        \includegraphics[width=0.184\linewidth]{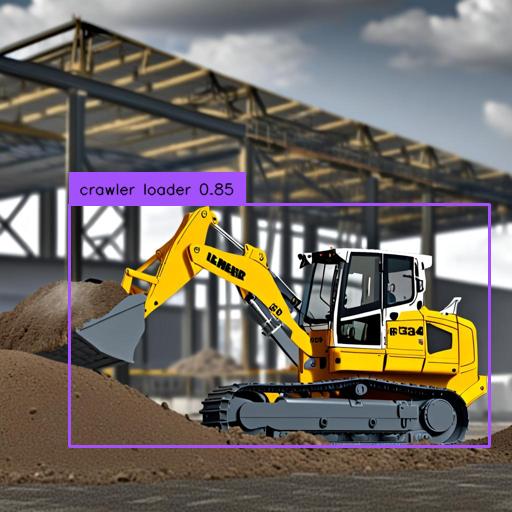} 
        \end{tabular} &
        \begin{tabular}{c}
        \includegraphics[width=0.184\linewidth]{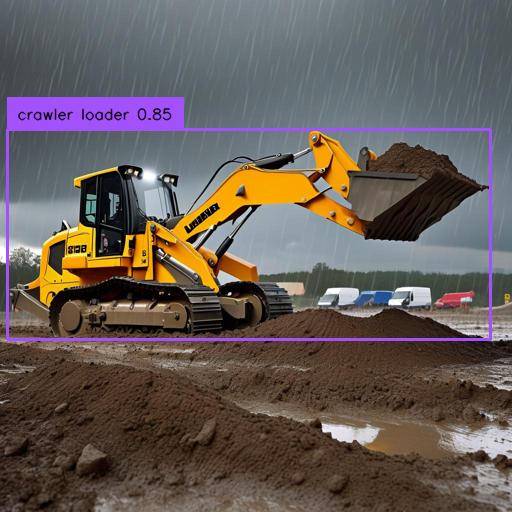} 
        \end{tabular} &
        \begin{tabular}{c}
        \includegraphics[width=0.184\linewidth]{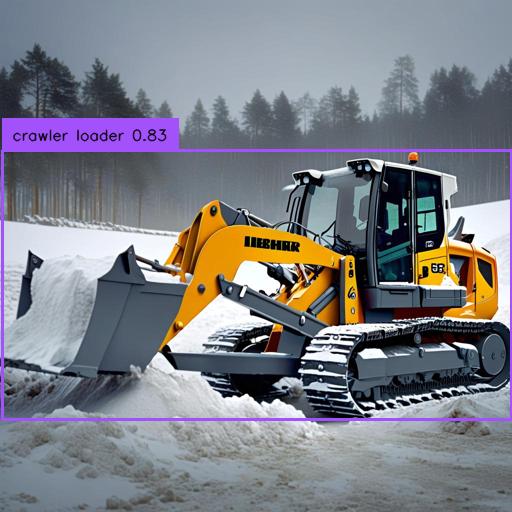}
        \end{tabular} \\
        
        {\footnotesize Instance data} & & {\begin{tabular}{c@{}c@{}c@{}c@{}} 19. Desert\\construction \end{tabular}} & {\begin{tabular}{c@{}c@{}c@{}c@{}} 40. Industrial\\site \end{tabular}} & {\begin{tabular}{c@{}c@{}c@{}c@{}} 4. Rainy\\construction site \end{tabular}} & {\begin{tabular}{c@{}c@{}c@{}c@{}} 17. Snowy\\day \end{tabular}} \\ \\
        
        \begin{tabular}{c c}
            \includegraphics[width=0.092\linewidth,height=0.092\linewidth]{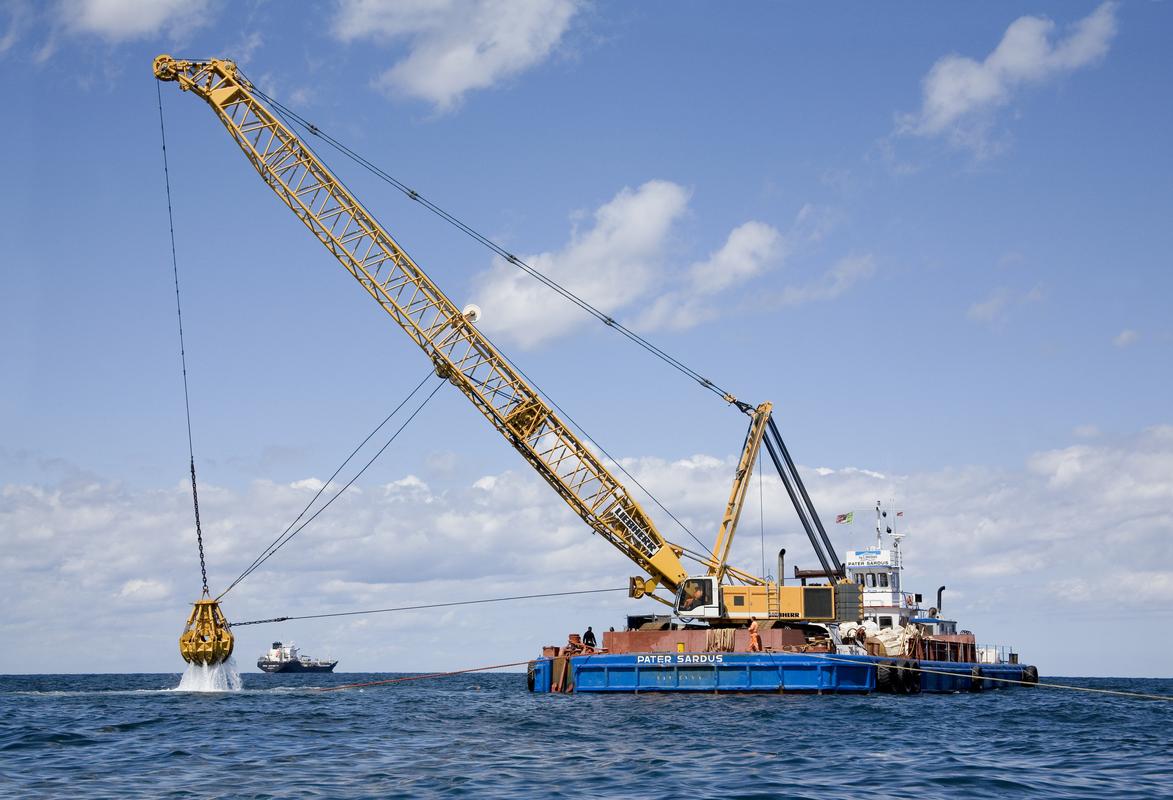} & 
            \includegraphics[width=0.092\linewidth,height=0.092\linewidth]{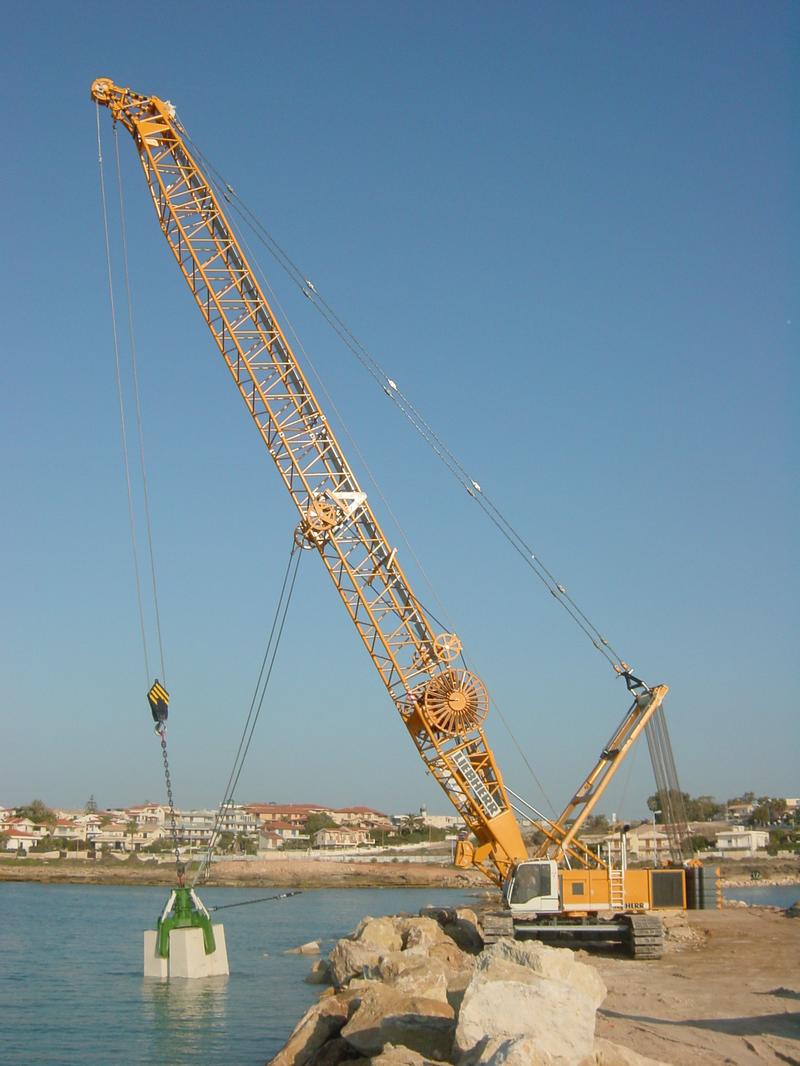} \\
            \includegraphics[width=0.092\linewidth,height=0.092\linewidth]{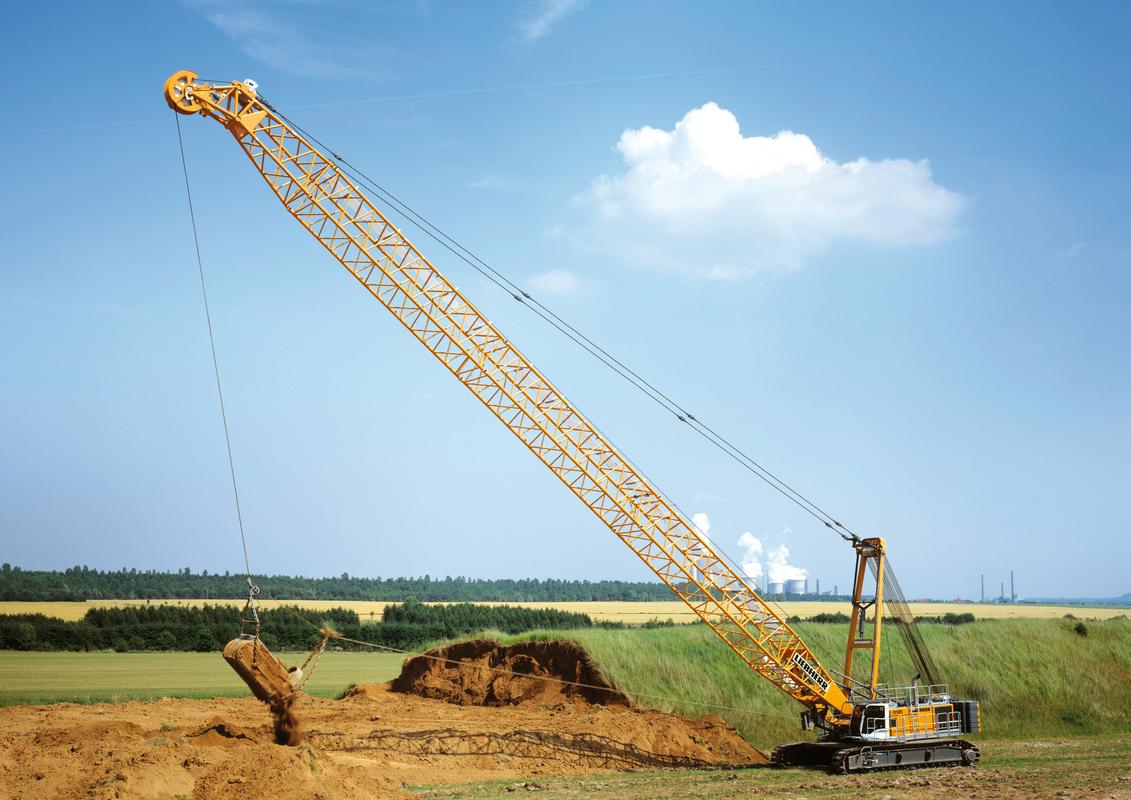} & 
            \includegraphics[width=0.092\linewidth,height=0.092\linewidth]{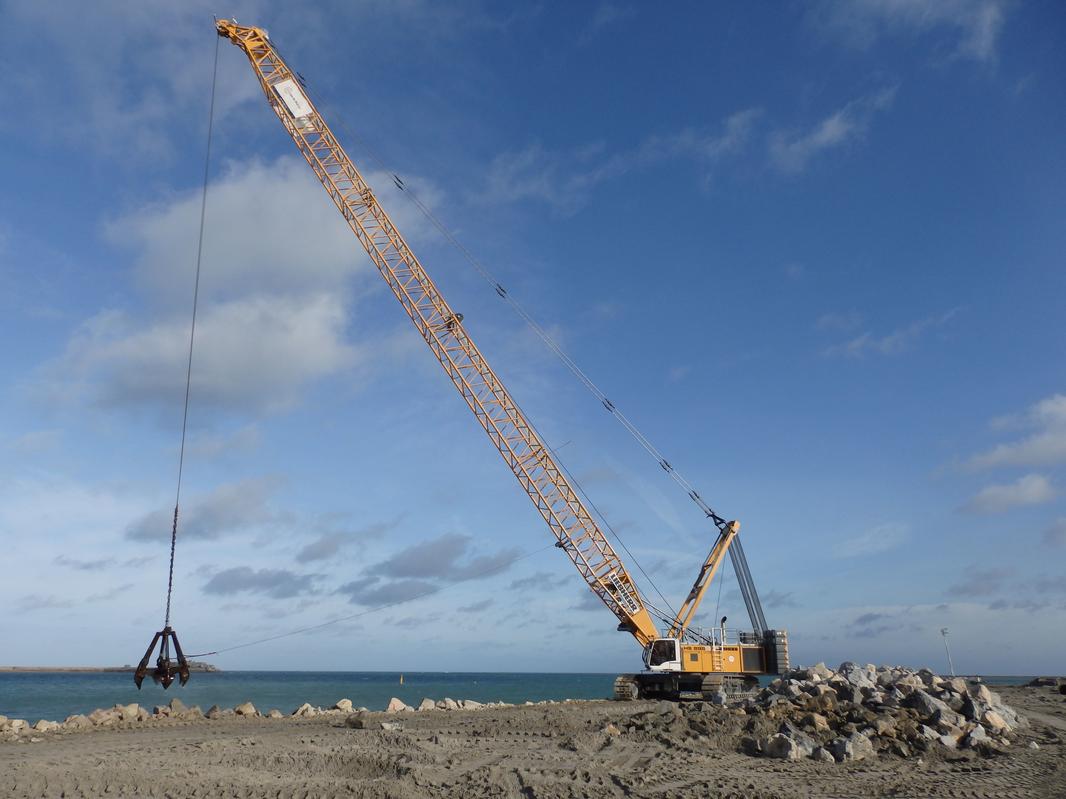}
        \end{tabular}
        &
        $\rightarrow$
        &
        \begin{tabular}{c}
        \includegraphics[width=0.184\linewidth]{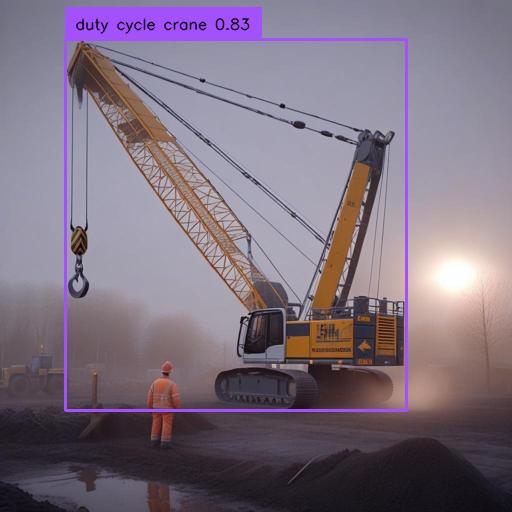}
        \end{tabular} &
        \begin{tabular}{c}
        \includegraphics[width=0.184\linewidth]{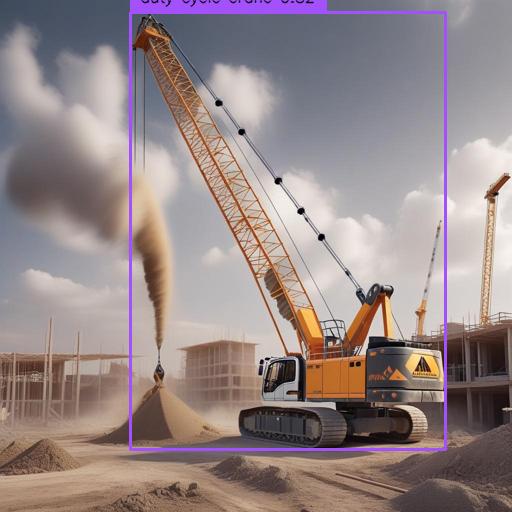} 
        \end{tabular} &
        \begin{tabular}{c}
        \includegraphics[width=0.184\linewidth]{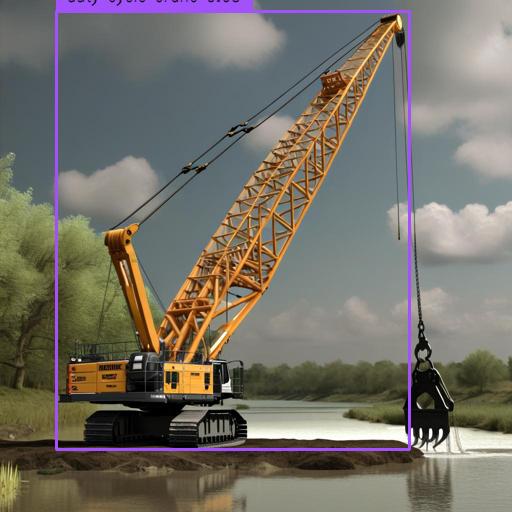} 
        \end{tabular} &
        \begin{tabular}{c}
        \includegraphics[width=0.184\linewidth]{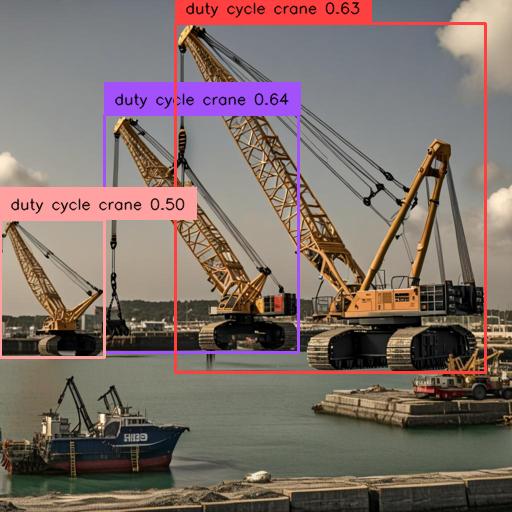}
        \end{tabular} \\
        
        {\footnotesize Instance data} & & {\begin{tabular}{c@{}c@{}c@{}c@{}} 16. Foggy\\morning \end{tabular}} & {\begin{tabular}{c@{}c@{}c@{}c@{}} 21. Windy\\day \end{tabular}} & {\begin{tabular}{c@{}c@{}c@{}c@{}} 23. Riverbank\\construction \end{tabular}} & {\begin{tabular}{c@{}c@{}c@{}c@{}} 45. Multiple\\machines harbor \end{tabular}} \\ \\

        \begin{tabular}{c c}
            \includegraphics[width=0.092\linewidth,height=0.092\linewidth]{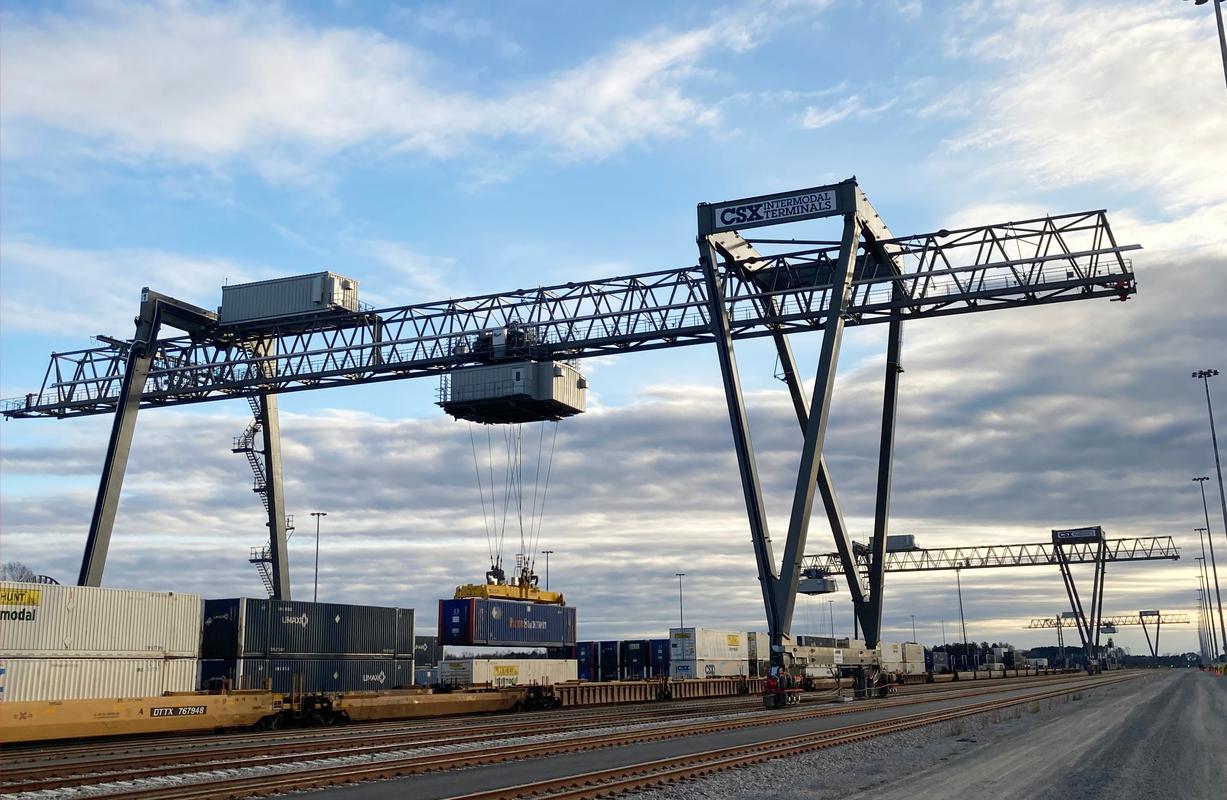} & 
            \includegraphics[width=0.092\linewidth,height=0.092\linewidth]{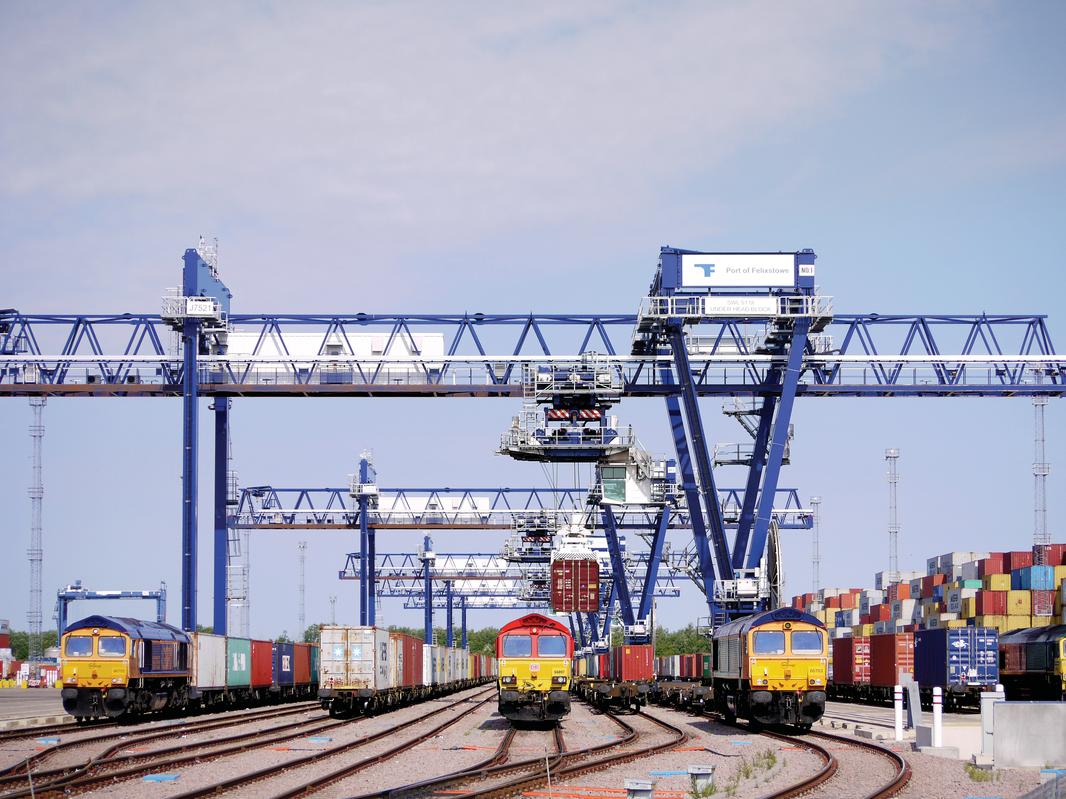} \\
            \includegraphics[width=0.092\linewidth,height=0.092\linewidth]{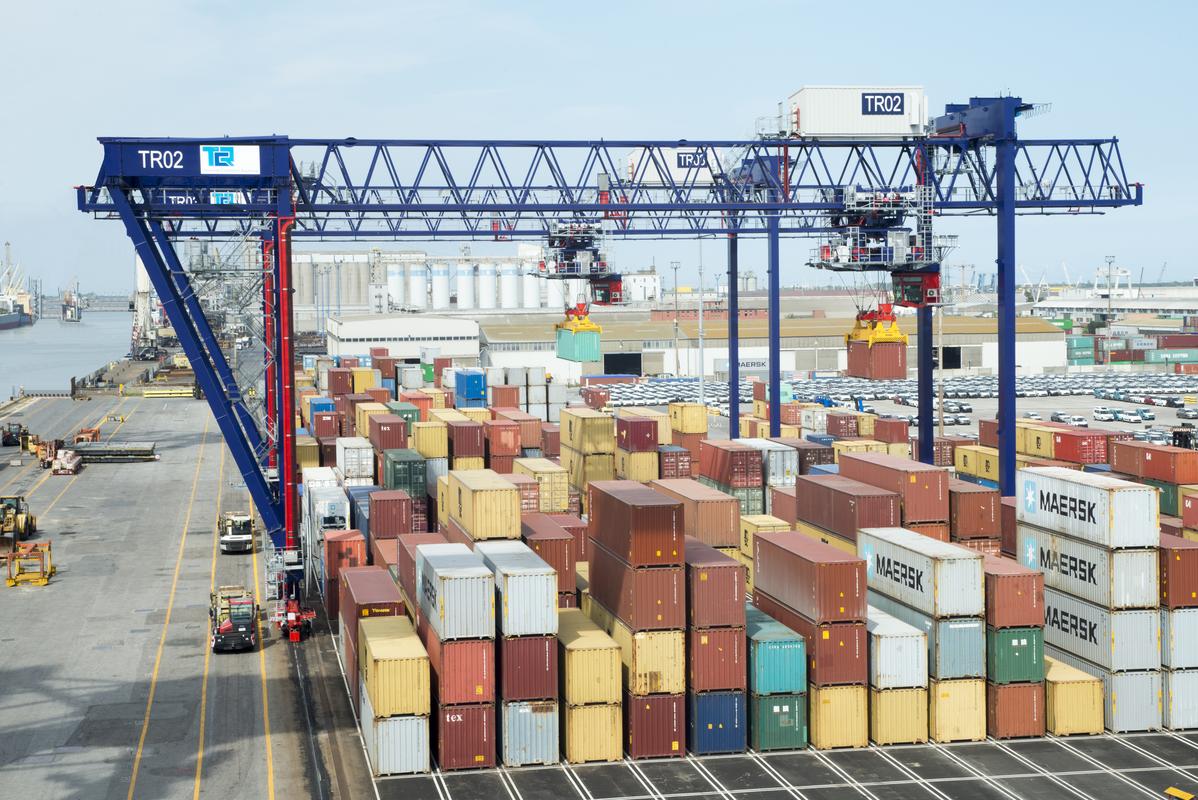} & 
            \includegraphics[width=0.092\linewidth,height=0.092\linewidth]{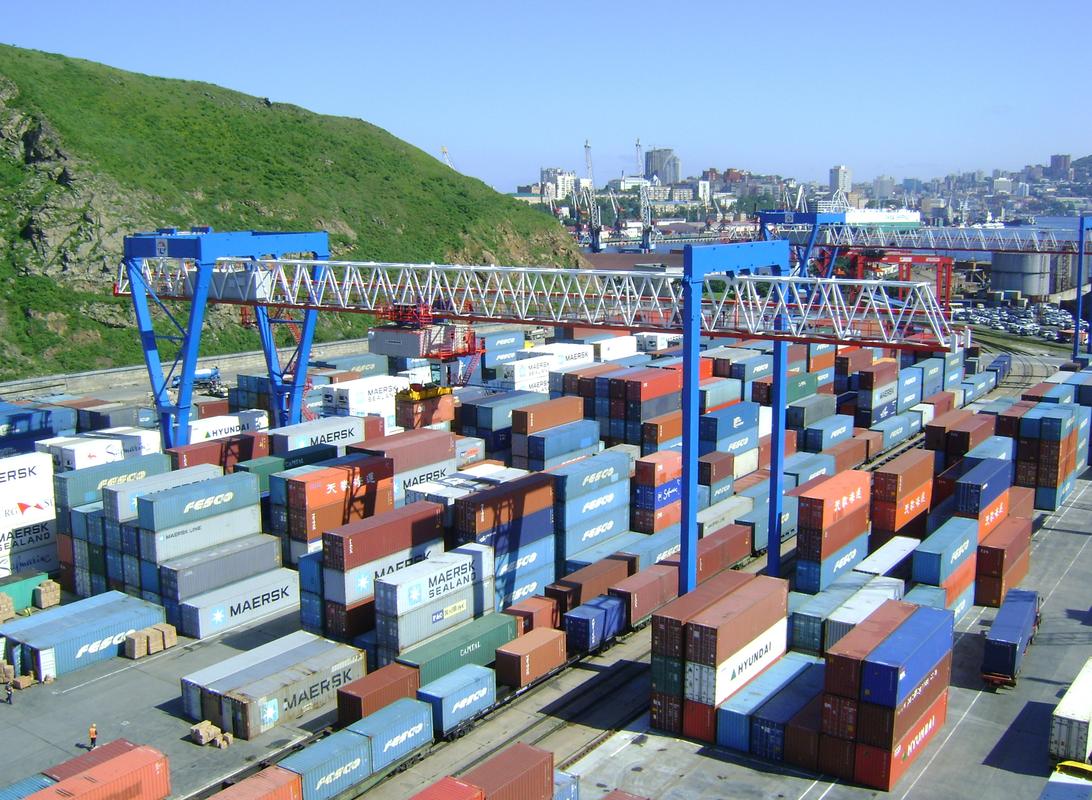}
        \end{tabular}
        
        &
        $\rightarrow$
        &
        \begin{tabular}{c}
        \includegraphics[width=0.184\linewidth]{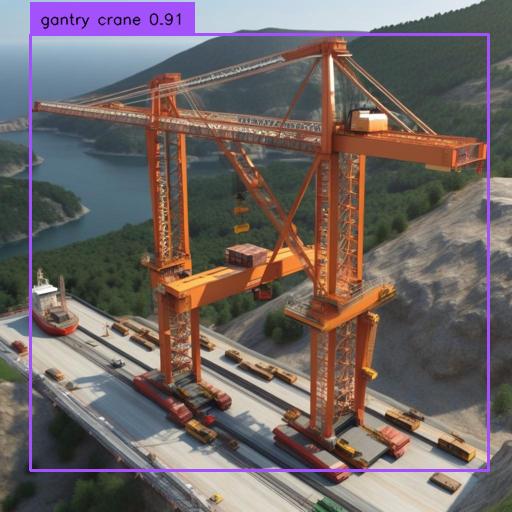}
        \end{tabular} &
        \begin{tabular}{c}
        \includegraphics[width=0.184\linewidth]{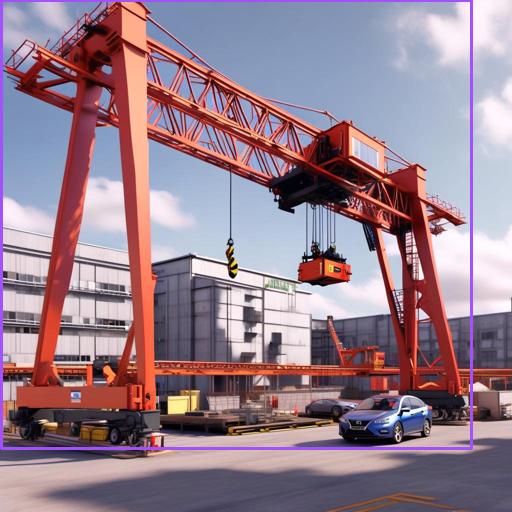} 
        \end{tabular} &
        \begin{tabular}{c}
        \includegraphics[width=0.184\linewidth]{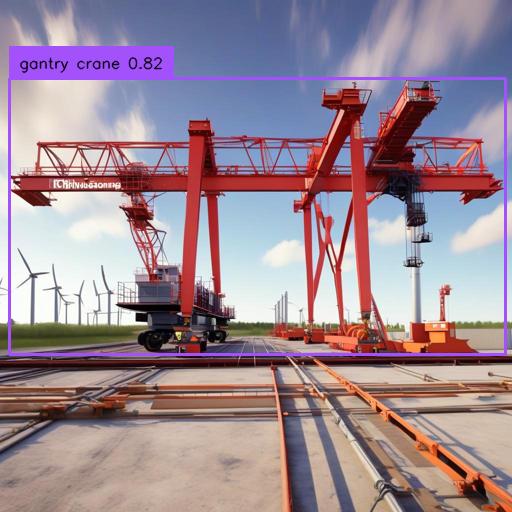} 
        \end{tabular} &
        \begin{tabular}{c}
        \includegraphics[width=0.184\linewidth]{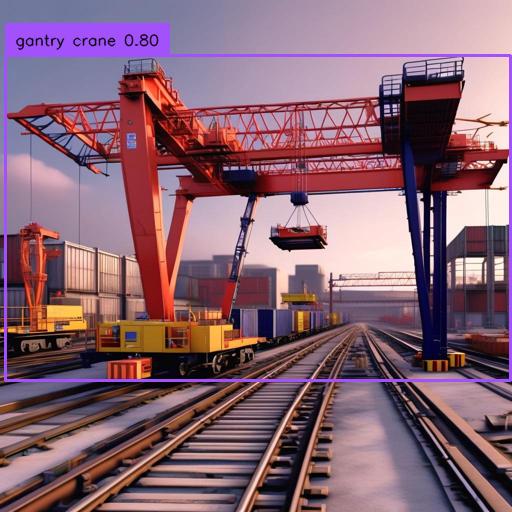}
        \end{tabular} \\
        
        {\footnotesize Instance data} & & {\begin{tabular}{c@{}c@{}c@{}c@{}} 31. Mountain\\construction \end{tabular}} & {\begin{tabular}{c@{}c@{}c@{}c@{}} 24. Urban\\parking lot \end{tabular}} & {\begin{tabular}{c@{}c@{}c@{}c@{}} 29. Wind farm\\construction \end{tabular}} & {\begin{tabular}{c@{}c@{}c@{}c@{}} 26. Railway\\construction \end{tabular}} \\ \\
        
        \begin{tabular}{c c}
            \includegraphics[width=0.092\linewidth,height=0.092\linewidth]{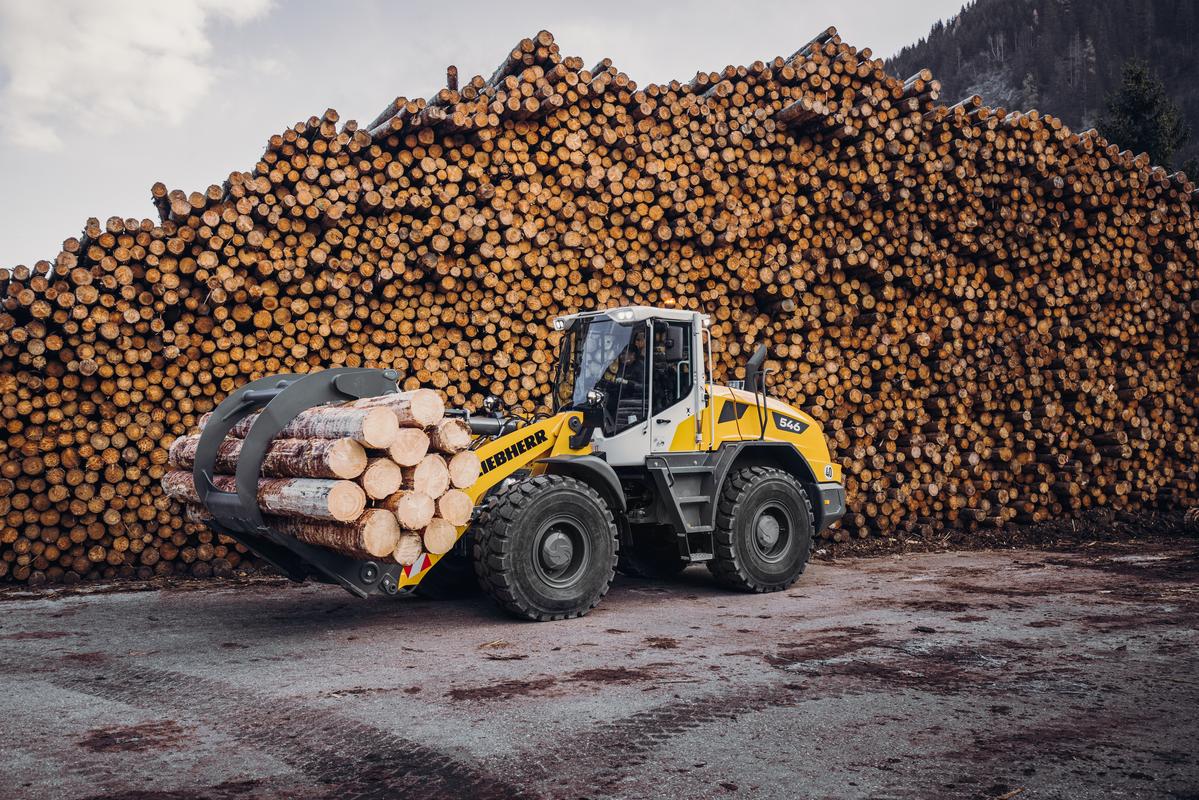} & 
            \includegraphics[width=0.092\linewidth,height=0.092\linewidth]{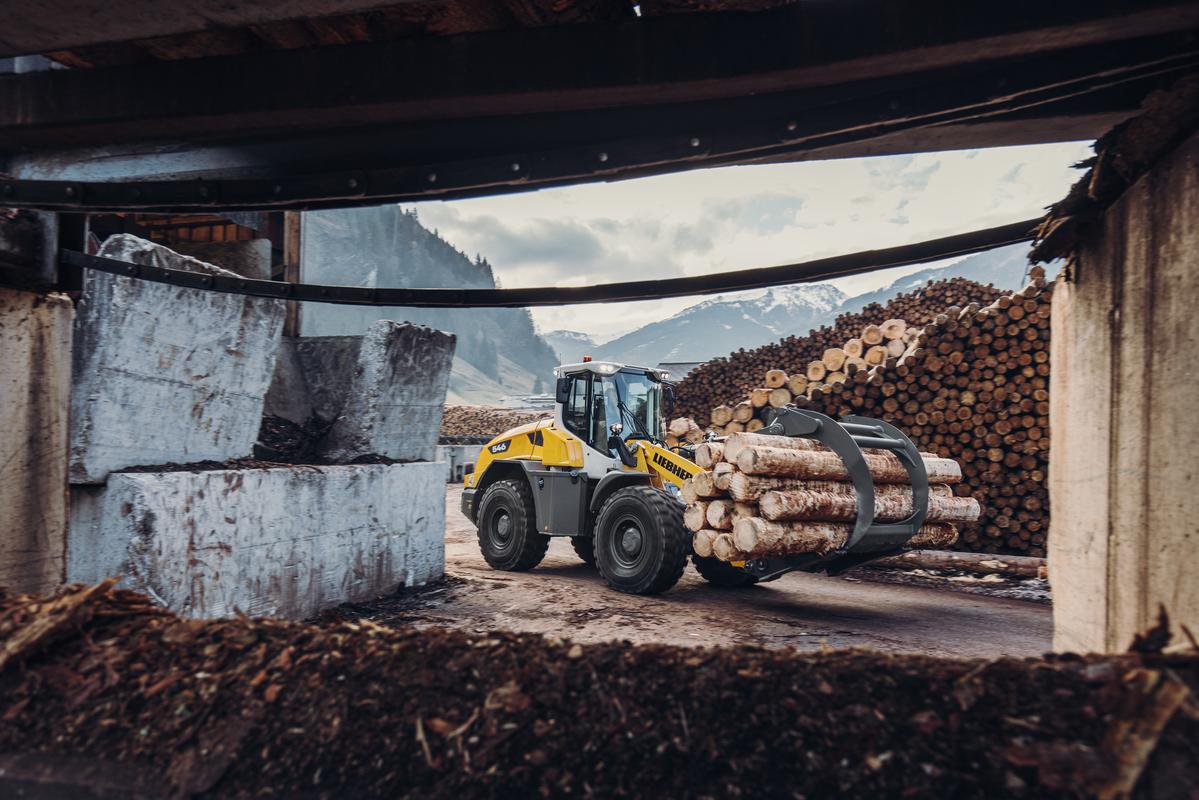} \\
            \includegraphics[width=0.092\linewidth,height=0.092\linewidth]{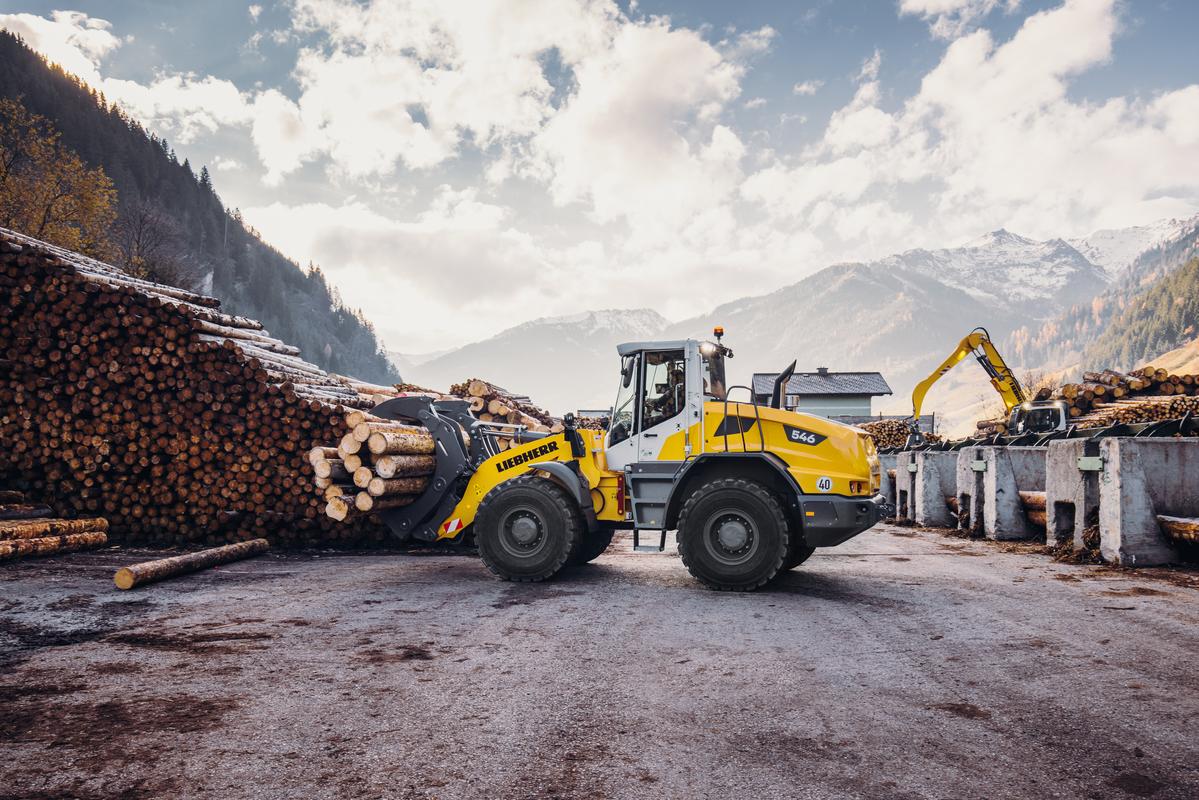} & 
            \includegraphics[width=0.092\linewidth,height=0.092\linewidth]{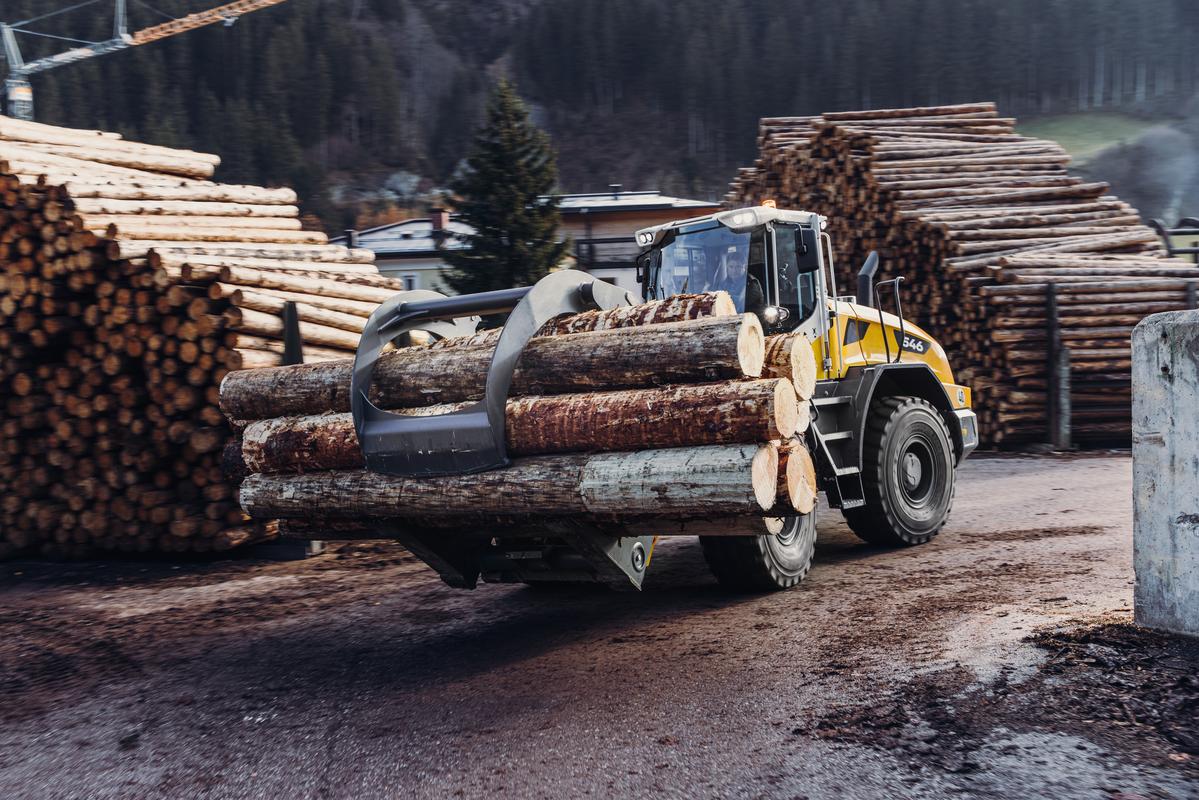}
        \end{tabular}
        
        &
        $\rightarrow$
        &
        \begin{tabular}{c}
        \includegraphics[width=0.184\linewidth]{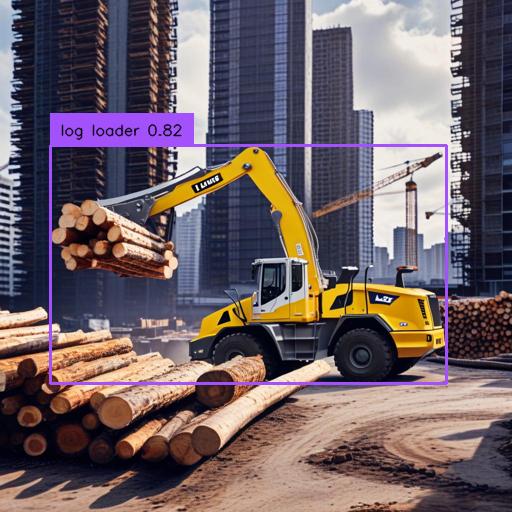}
        \end{tabular} &
        \begin{tabular}{c}
        \includegraphics[width=0.184\linewidth]{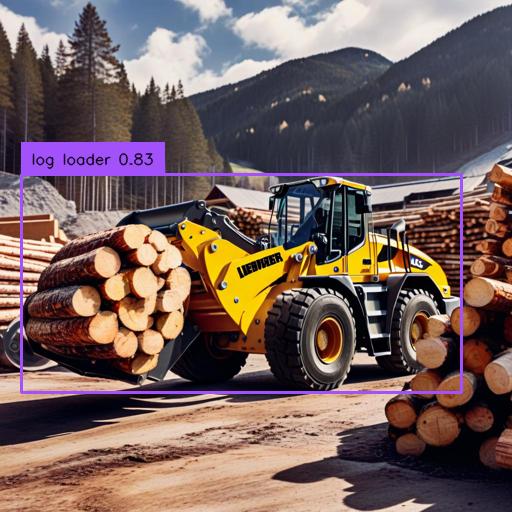} 
        \end{tabular} &
        \begin{tabular}{c}
        \includegraphics[width=0.184\linewidth]{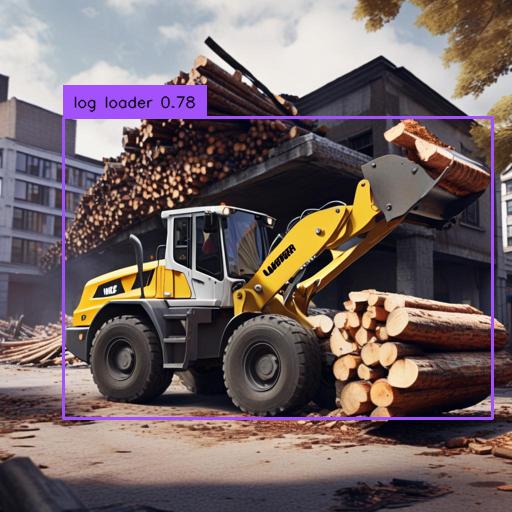} 
        \end{tabular} &
        \begin{tabular}{c}
        \includegraphics[width=0.184\linewidth]{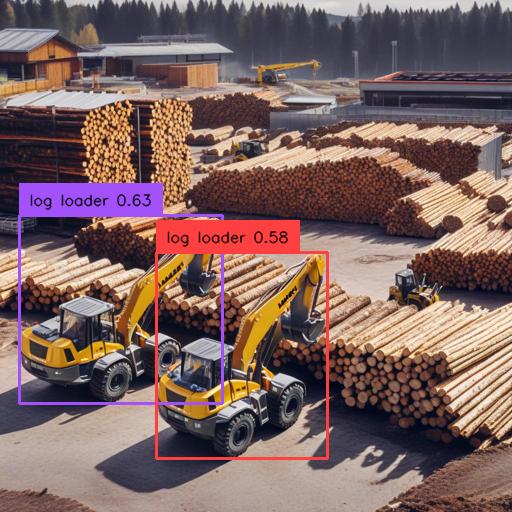}
        \end{tabular} \\
        
        {\footnotesize Instance data} & & {\begin{tabular}{c@{}c@{}c@{}c@{}} 28. Skyscraper\\construction \end{tabular}} & {\begin{tabular}{c@{}c@{}c@{}c@{}} 2. Sunny\\construction site \end{tabular}} & {\begin{tabular}{c@{}c@{}c@{}c@{}} 20.Urban\\demolition \end{tabular}} & {\begin{tabular}{c@{}c@{}c@{}c@{}} 35. Large\\scale construction \end{tabular}} \\ \\
        
    \end{tabular}}
    \caption{Visualization of data diversification and bounding box annotation for additional categories}
    \label{fig:approved_annotated_generated_images_app} 
\end{figure}

\autoref{fig:approved_annotated_generated_images_app} provides more examples for visualization of data diversification and bounding box annotation detailed in \autoref{sec:vis}.

\begin{figure}[!htbp]
    \centering
    \includegraphics[width=0.95\columnwidth]{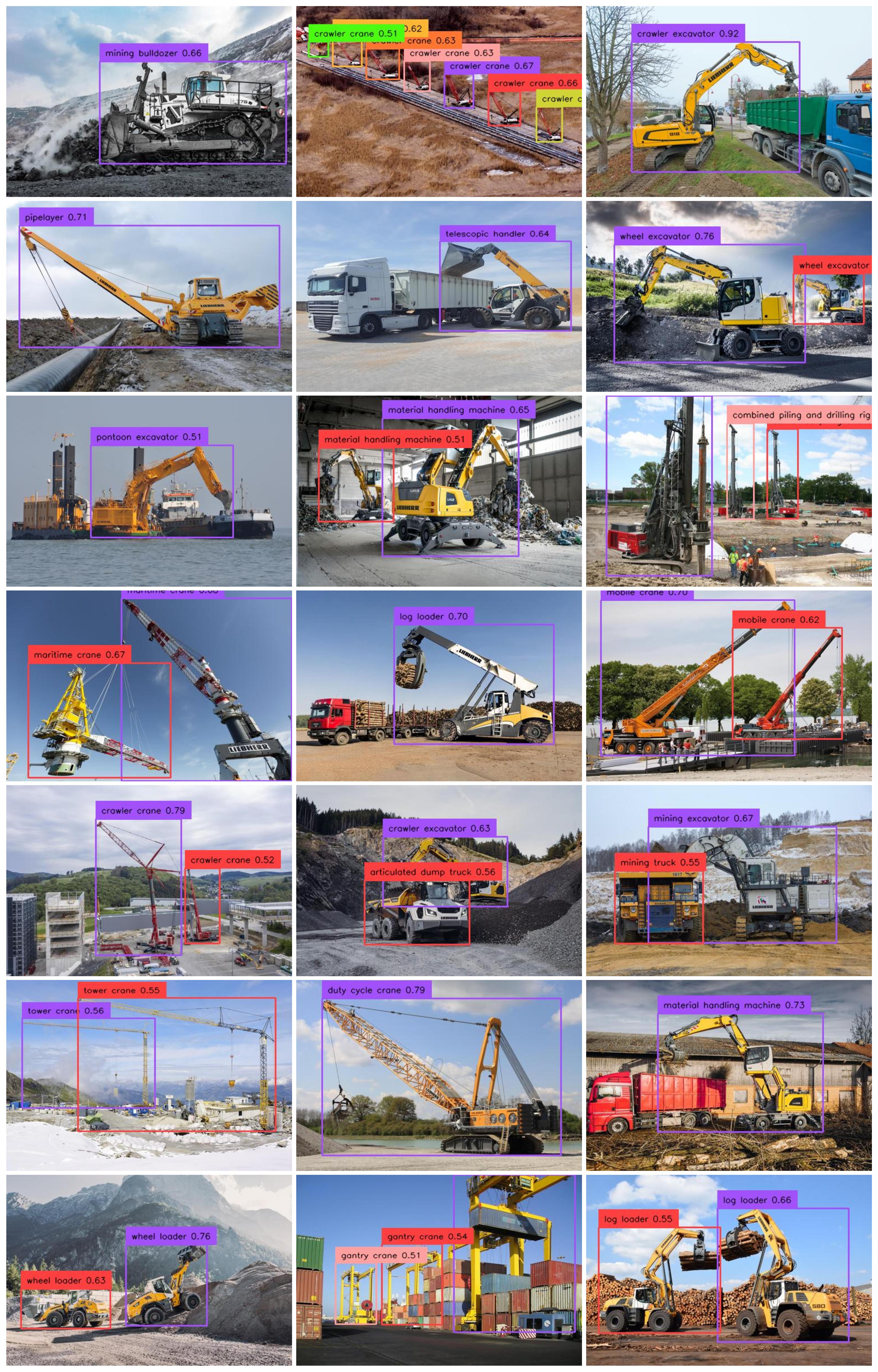}
    \caption{Original images annotated by Grounding DINO and approved by GPT-4o}
    \label{fig:image_grid_orig_aa}
\end{figure}

\begin{figure}[!htbp]
    \centering
    \includegraphics[width=0.95\columnwidth]{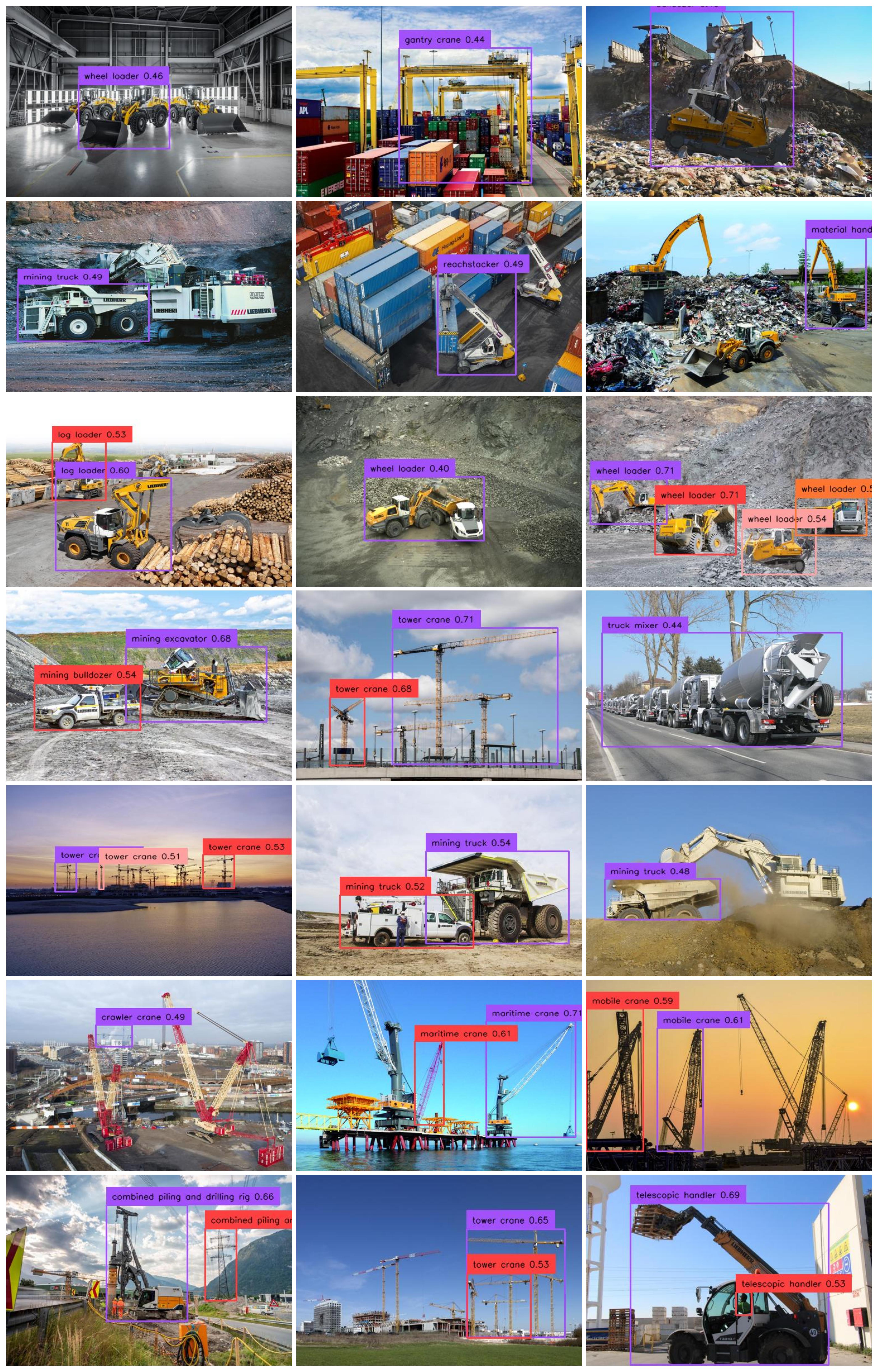}
    \caption{Original images annotated by Grounding DINO but disapproved by GPT-4o}
    \label{fig:image_grid_orig_da}
\end{figure}

\begin{figure}[!htbp]
    \centering
    \includegraphics[width=1\columnwidth]{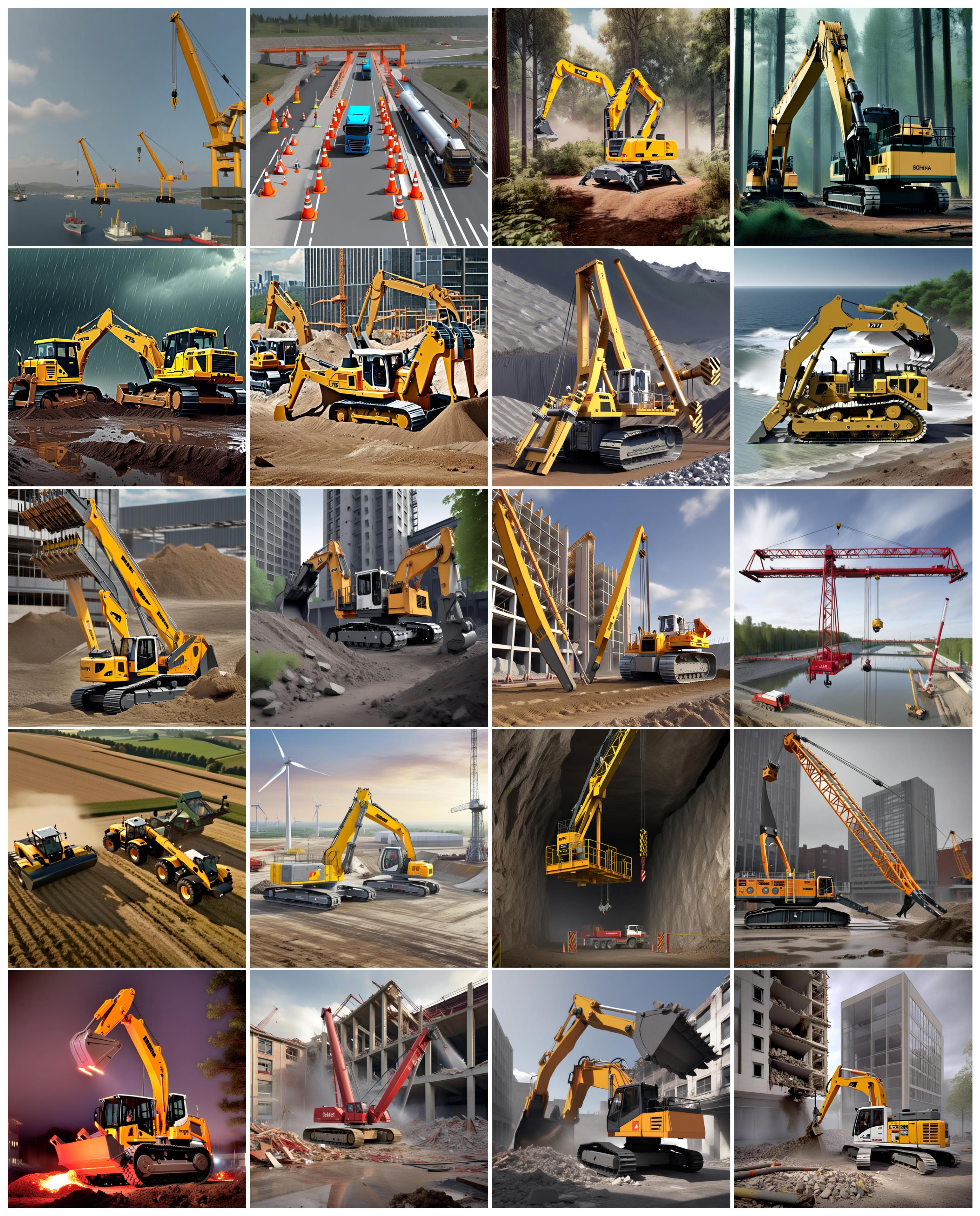}
    \caption{Generated images by DreamBooth with SDXL but disapproved by InternVL-1.5}
    \label{fig:image_grid_gen_d}
\end{figure}

\autoref{fig:image_grid_orig_aa}, \autoref{fig:image_grid_orig_da}, and \autoref{fig:image_grid_gen_d} showcase the effectiveness of LMM-based review. Specifically, \autoref{fig:image_grid_orig_aa} and \autoref{fig:image_grid_orig_da} illustrate annotated images approved and disapproved by GPT-4o during pseudo-label review, respectively. GPT-4o not only accurately assesses bounding box annotations in complex scenarios, including multiple objects and occlusion, but also identifies common annotation errors, such as missing objects, wrong labels, and inaccurate bounding box coordinates. On the other hand, the results of the photorealism review of generated images are demonstrated in \autoref{fig:image_grid_gen_d}. InternVL-1.5 can detect unrealistic objects and scenes. Preventing these inferiors from being included in the dataset is proven (in \autoref{sec:ablation} and \autoref{sec:lmm_exp}) to help improve the performance. The cases provided by these figures further validate the efficacy and essential role of the LMM-based review in the DART pipeline.

\begin{figure}[!h]
    \centering
    \setlength{\tabcolsep}{1pt}
    {\scriptsize
    \begin{tabular}{c@{\hskip 5pt} c@{\hskip 5pt} c c c c}
    
        \begin{tabular}{c c}
            \includegraphics[width=0.092\linewidth,height=0.092\linewidth]{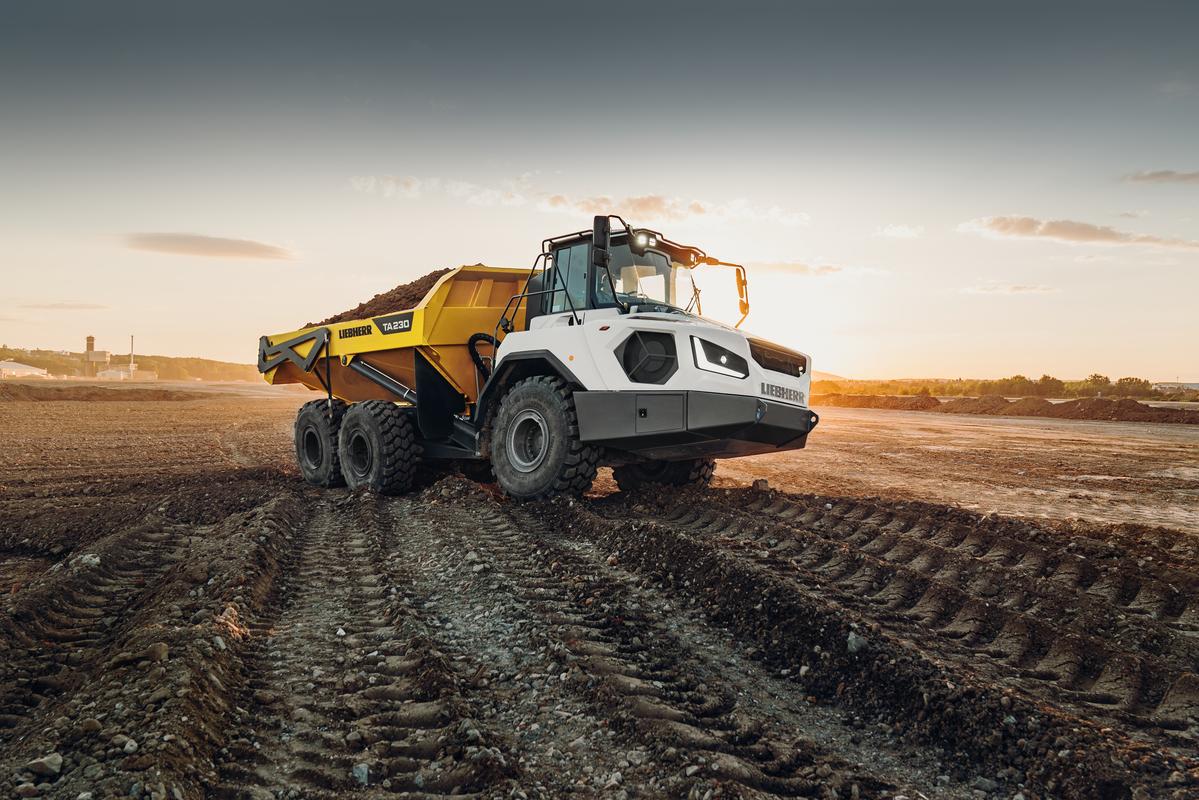} & 
            \includegraphics[width=0.092\linewidth,height=0.092\linewidth]{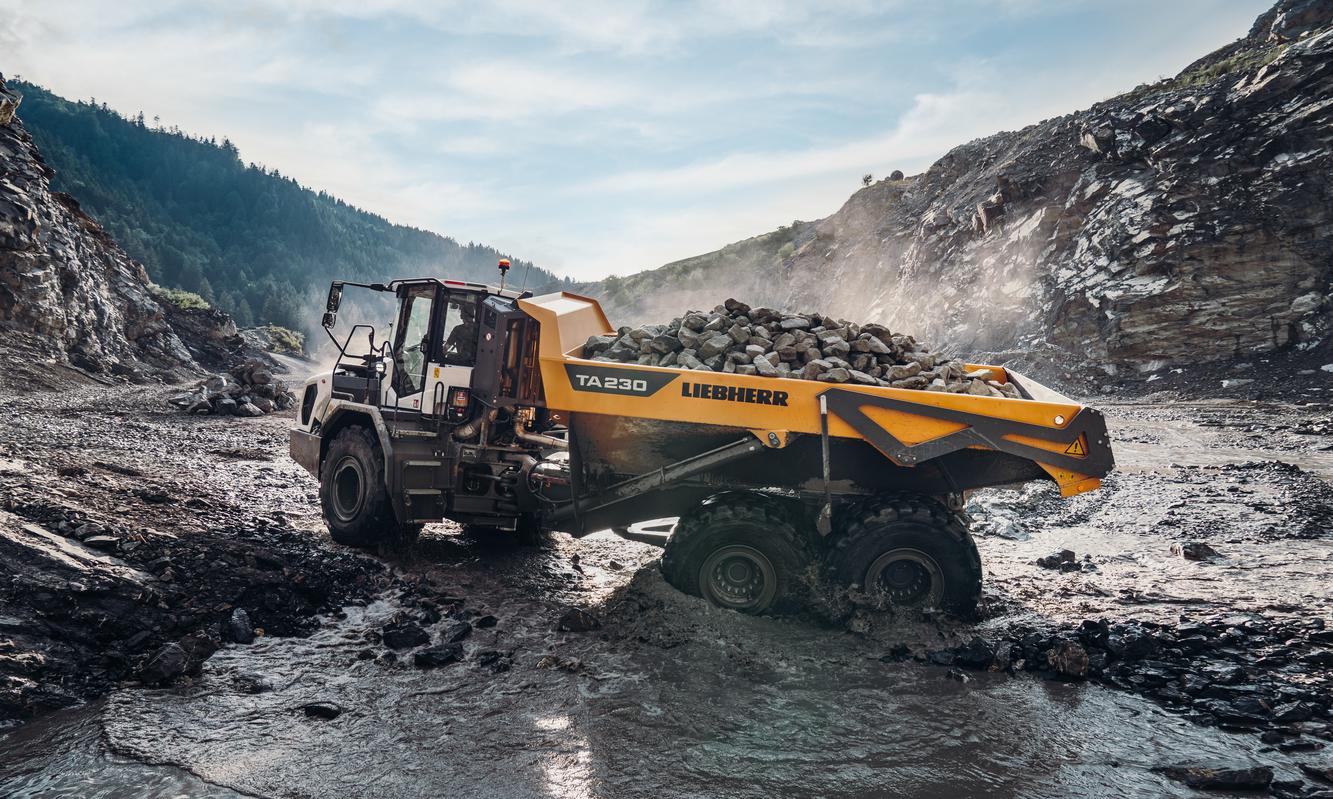} \\
            \includegraphics[width=0.092\linewidth,height=0.092\linewidth]{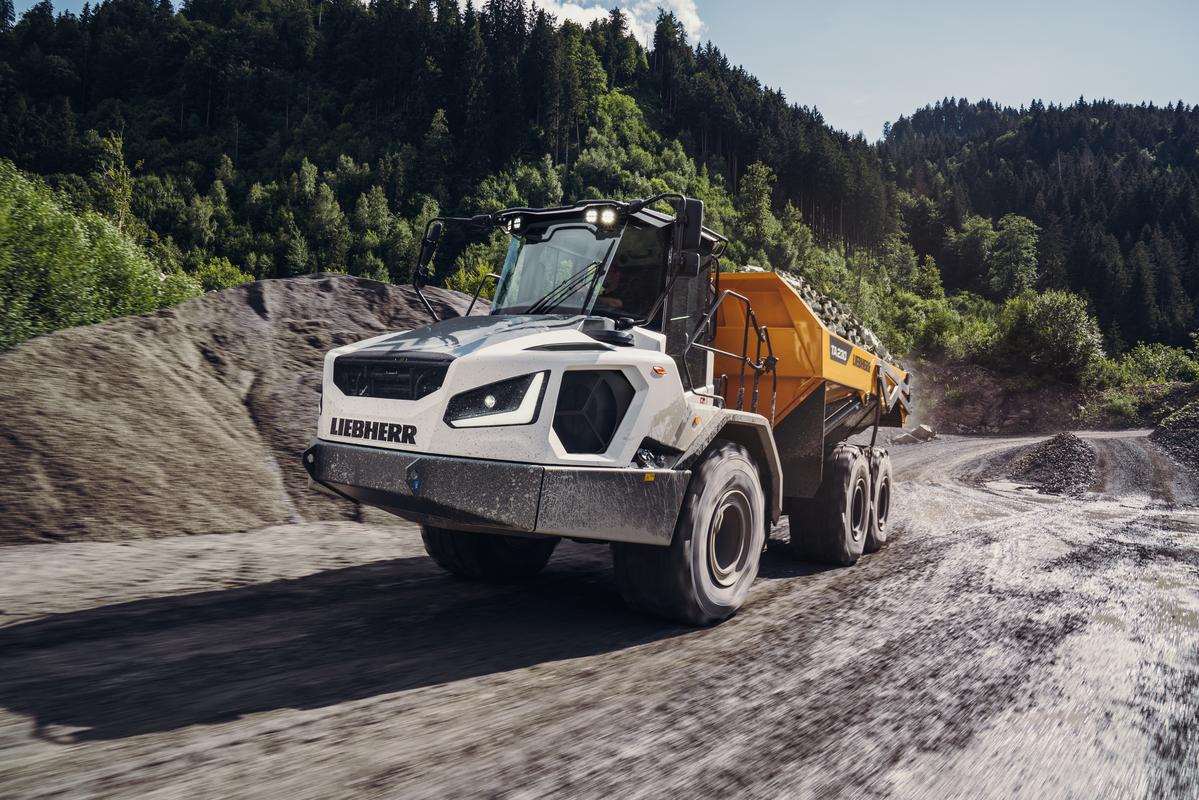} & 
            \includegraphics[width=0.092\linewidth,height=0.092\linewidth]{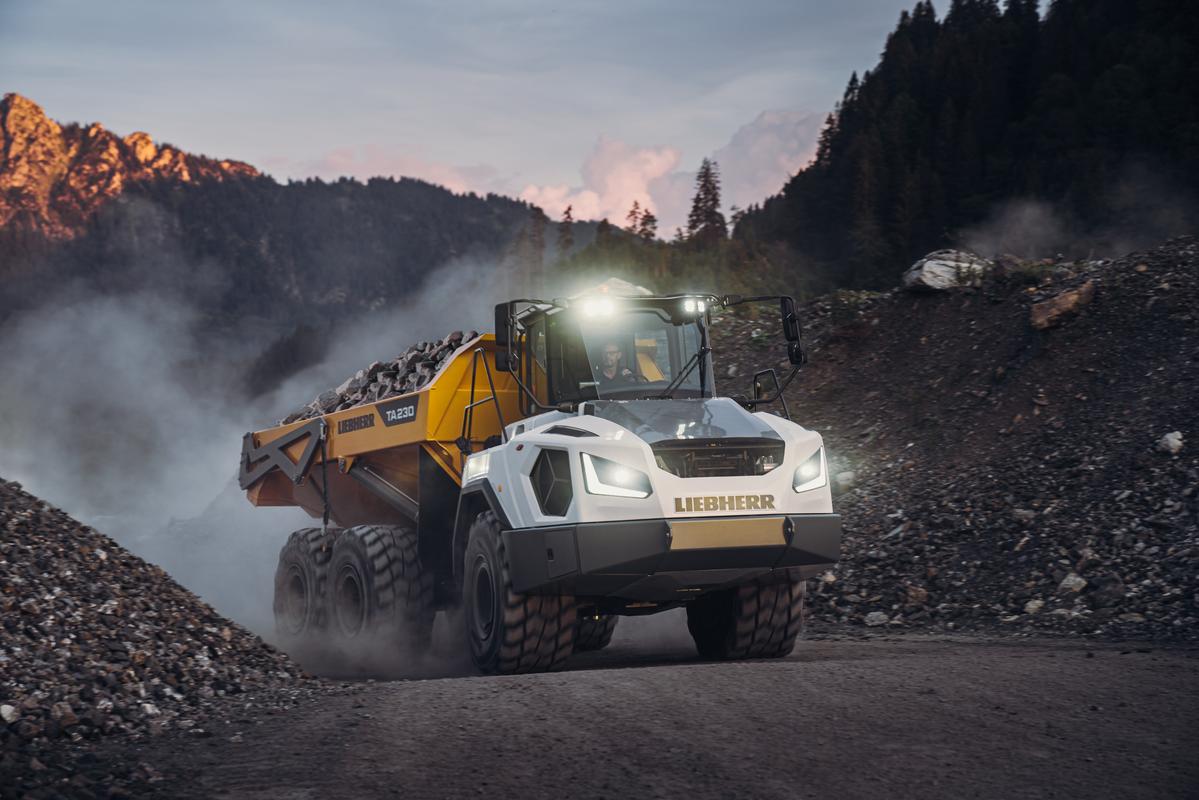}
        \end{tabular}
        &
        $\rightarrow$
        &
        \begin{tabular}{c}
        \includegraphics[width=0.184\linewidth]{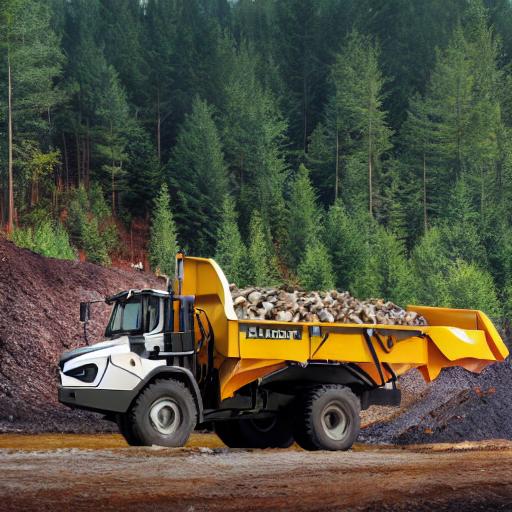}
        \end{tabular} &
        \begin{tabular}{c}
        \includegraphics[width=0.184\linewidth]{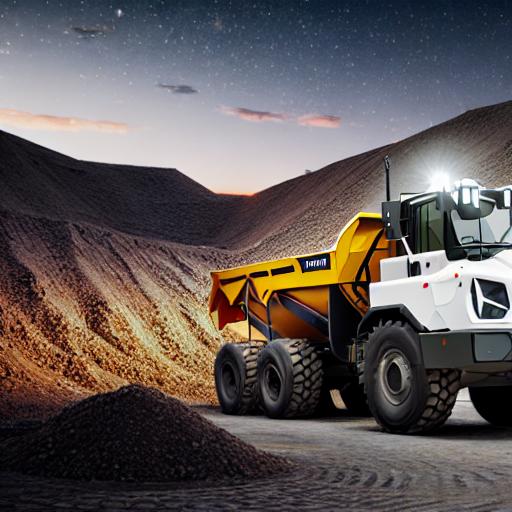} 
        \end{tabular} &
        \begin{tabular}{c}
        \includegraphics[width=0.184\linewidth]{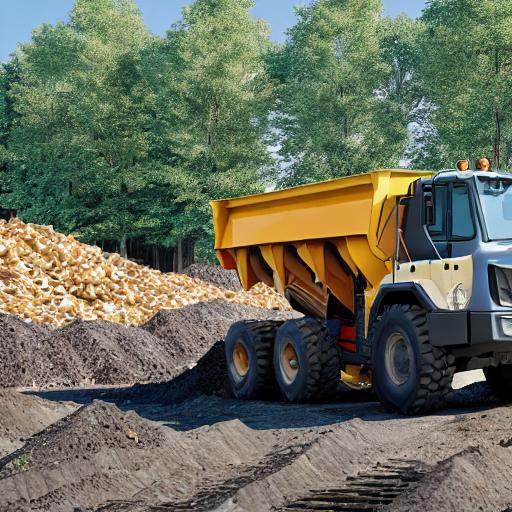} 
        \end{tabular} &
        \begin{tabular}{c}
        \includegraphics[width=0.184\linewidth]{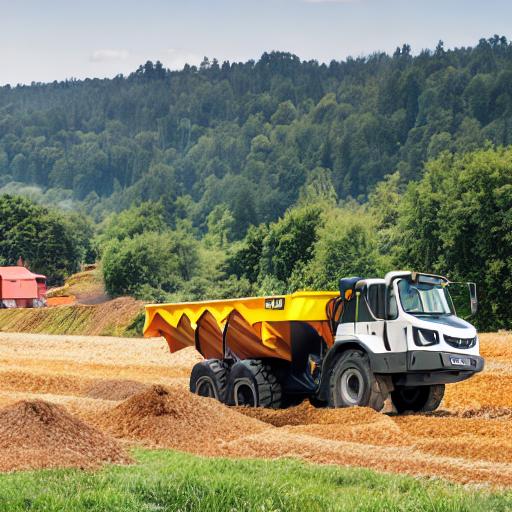}
        \end{tabular} \\
        
        {\footnotesize Instance data} & SD-1.5& {\begin{tabular}{c@{}c@{}c@{}c@{}} 32. Forest\\construction \end{tabular}} & {\begin{tabular}{c@{}c@{}c@{}c@{}} 15.Nighttime\\ construction \end{tabular}} & {\begin{tabular}{c@{}c@{}c@{}c@{}} 37. Suburban\\construction\end{tabular}} & {\begin{tabular}{c@{}c@{}c@{}c@{}} 44. Multiple machines\\rural area\end{tabular}} \\ \\

        \begin{tabular}{c c}
            \includegraphics[width=0.092\linewidth,height=0.092\linewidth]{images/SD/TA230/00012.jpg} & 
            \includegraphics[width=0.092\linewidth,height=0.092\linewidth]{images/SD/TA230/00017.jpg} \\
            \includegraphics[width=0.092\linewidth,height=0.092\linewidth]{images/SD/TA230/00018.jpg} & 
            \includegraphics[width=0.092\linewidth,height=0.092\linewidth]{images/SD/TA230/00019.jpg}
        \end{tabular}
        &
        $\rightarrow$
        &
        \begin{tabular}{c}
        \includegraphics[width=0.184\linewidth]{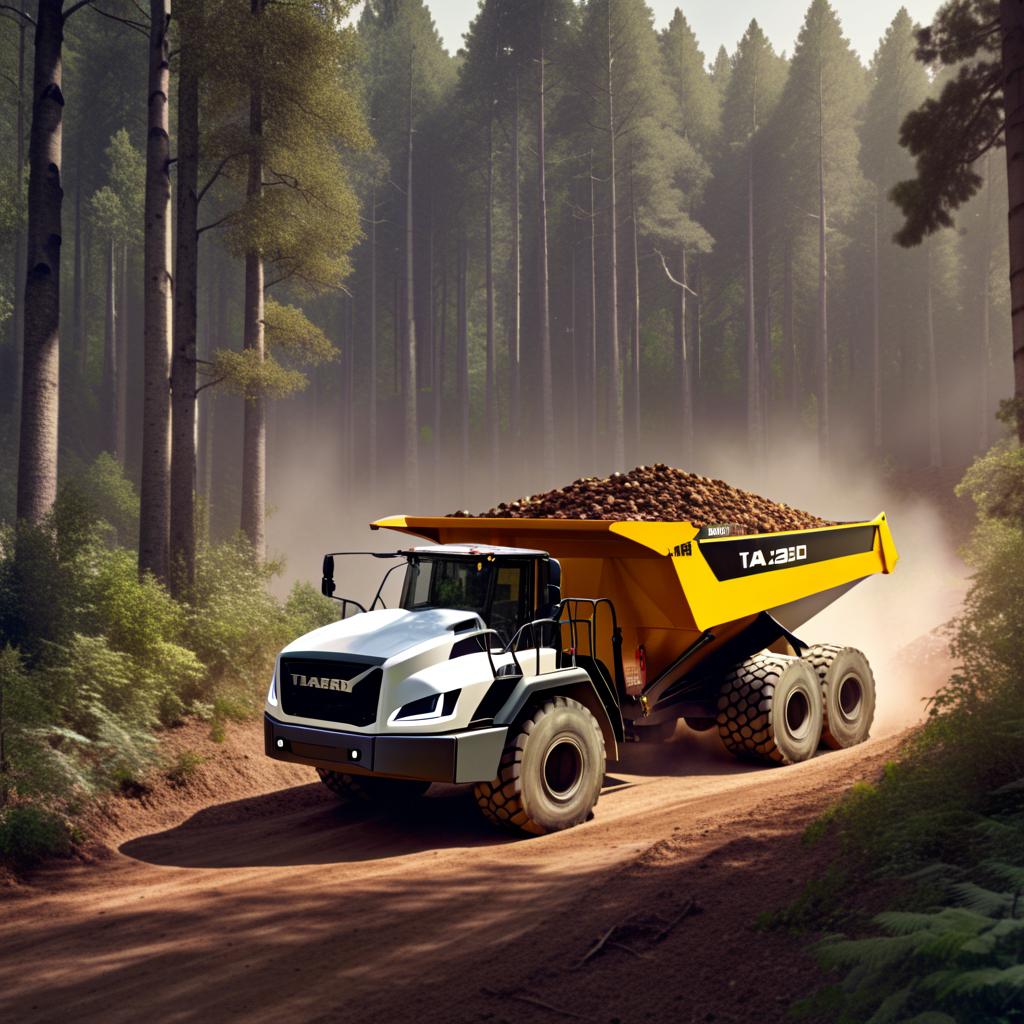}
        \end{tabular} &
        \begin{tabular}{c}
        \includegraphics[width=0.184\linewidth]{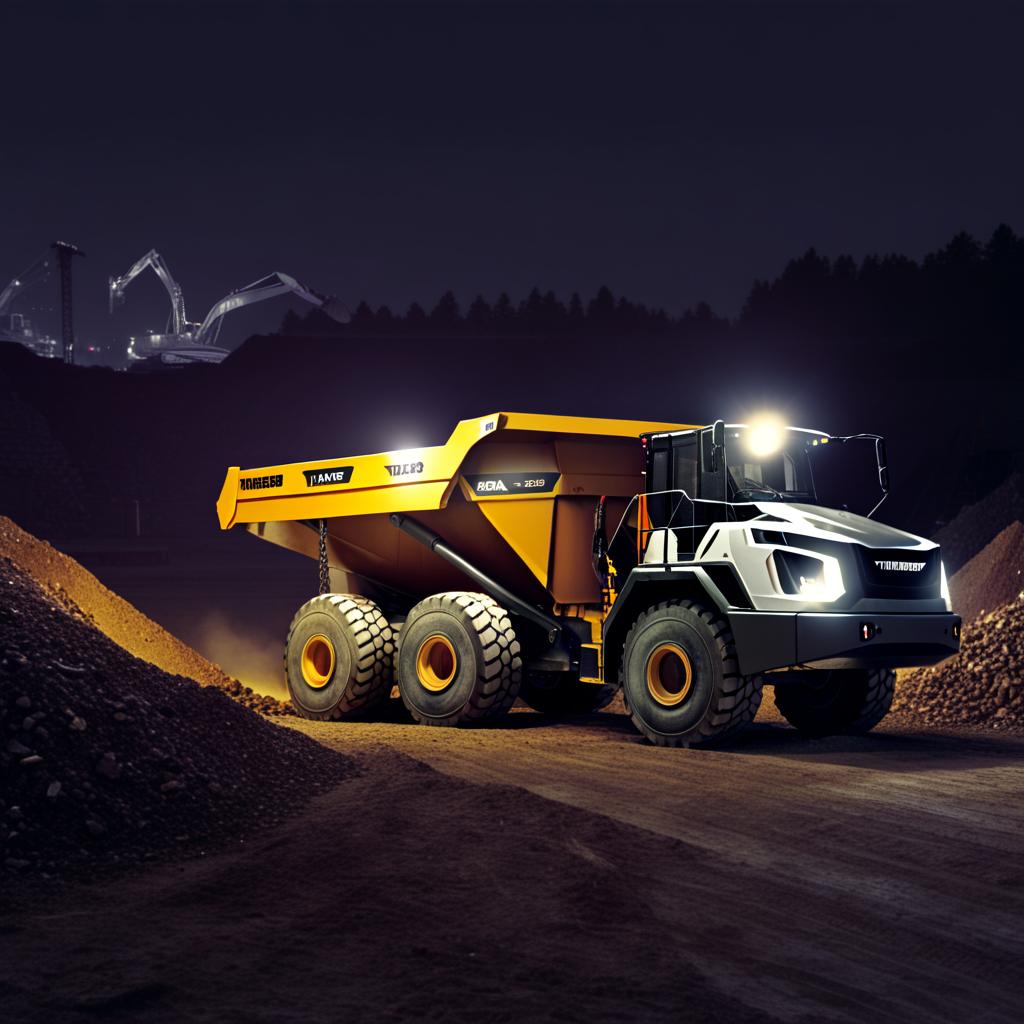}
        \end{tabular} &
        \begin{tabular}{c}
        \includegraphics[width=0.184\linewidth]{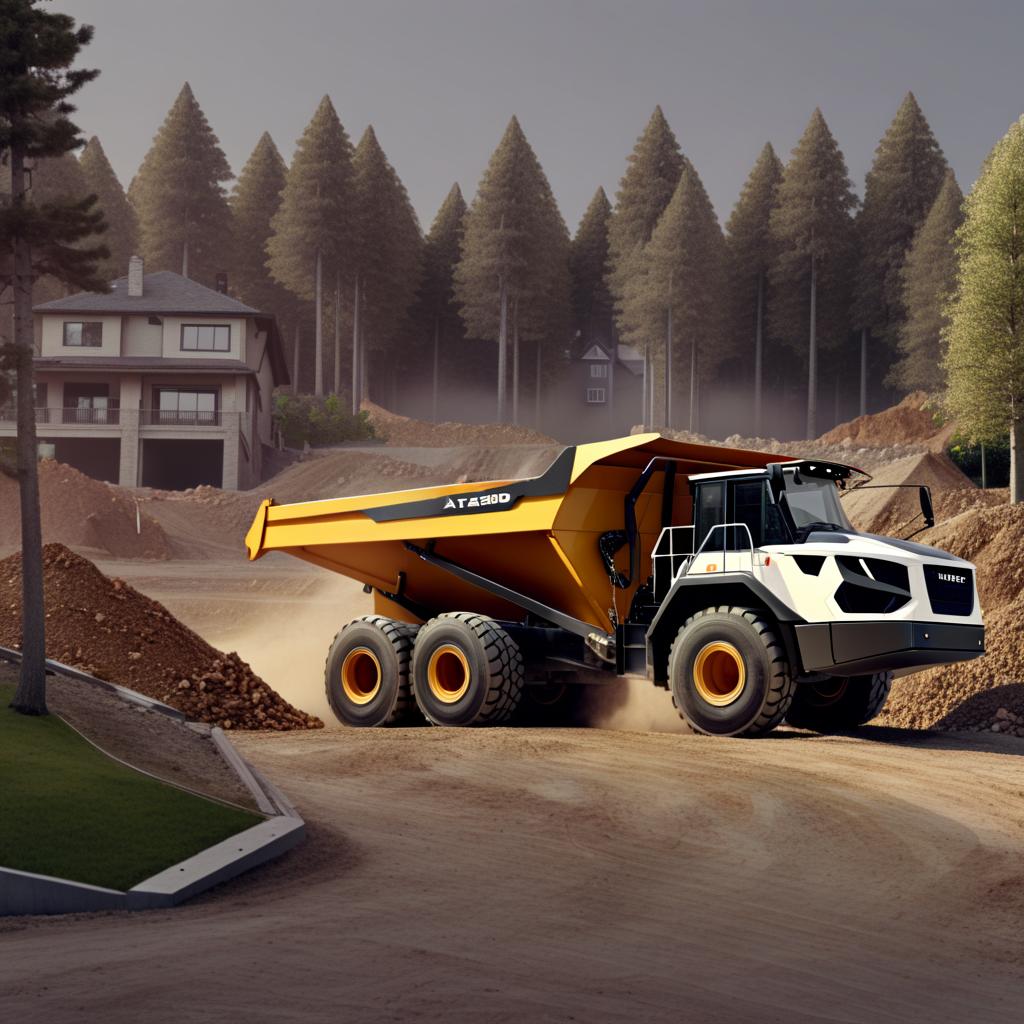}
        \end{tabular} &
        \begin{tabular}{c}
        \includegraphics[width=0.184\linewidth]{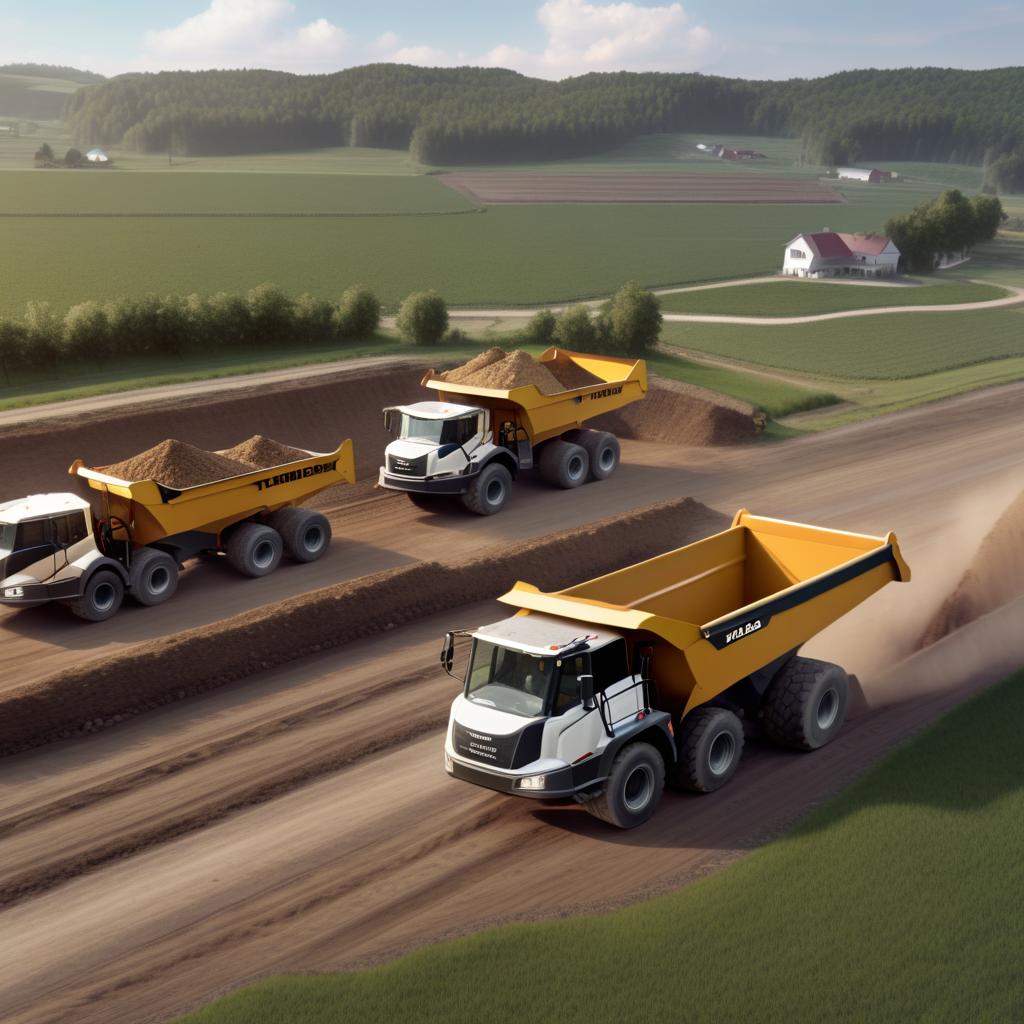}
        \end{tabular} \\
        
        {\footnotesize Instance data} & SDXL & {\begin{tabular}{c@{}c@{}c@{}c@{}} 32. Forest\\construction \end{tabular}} & {\begin{tabular}{c@{}c@{}c@{}c@{}} 15.Nighttime\\ construction \end{tabular}} & {\begin{tabular}{c@{}c@{}c@{}c@{}} 37. Suburban\\construction\end{tabular}} & {\begin{tabular}{c@{}c@{}c@{}c@{}} 44. Multiple machines\\rural area\end{tabular}} \\ \\
    \end{tabular}}
    \caption{SDXL vs SD-1.5}
    \label{fig:sdxl_vs_sd15} 
\end{figure}

Finally, \autoref{fig:sdxl_vs_sd15} illustrates comparisons between SDXL and SD-1.5 for data diversification. The two rows of generated images on the right side of the figure are produced by SD-1.5 and SDXL, respectively. To ensure a fair comparison, both models are trained under the same DreamBooth framework by identical instance data, and the images are induced by the exact same prompts (with ID and description given as captions underneath; refer to \autoref{tab:prompts_dreambooth} for the exact text prompt) during inference. It is clear from the figure that images generated by SDXL are not only more detailed and realistic but also better at interpreting and following complex text prompts. In contrast, SD-1.5 images tend to have monotonous poses and simple scenes. More importantly, SD-1.5 fails to capture the diverse scenarios described in the text prompts, especially for unusual scenes. Therefore, we only adopt DreamBooth with SDXL for our data diversification phase.

\FloatBarrier
\bibliographystyle{elsarticle-num}
\bibliography{DART}

\begin{thebibliography}{10}
\expandafter\ifx\csname url\endcsname\relax
  \def\url#1{\texttt{#1}}\fi
\expandafter\ifx\csname urlprefix\endcsname\relax\def\urlprefix{URL }\fi
\expandafter\ifx\csname href\endcsname\relax
  \def\href#1#2{#2} \def\path#1{#1}\fi

\bibitem{rombachHighResolutionImageSynthesis2022}
R.~Rombach, A.~Blattmann, D.~Lorenz, P.~Esser, B.~Ommer, \href{http://arxiv.org/abs/2112.10752}{High-{{Resolution Image Synthesis}} with {{Latent Diffusion Models}}} (Apr. 2022).
\newblock \href {http://arxiv.org/abs/2112.10752} {\path{arXiv:2112.10752}}.
\newline\urlprefix\url{http://arxiv.org/abs/2112.10752}

\bibitem{rameshHierarchicalTextConditionalImage2022}
A.~Ramesh, P.~Dhariwal, A.~Nichol, C.~Chu, M.~Chen, \href{http://arxiv.org/abs/2204.06125}{Hierarchical {{Text-Conditional Image Generation}} with {{CLIP Latents}}} (Apr. 2022).
\newblock \href {http://arxiv.org/abs/2204.06125} {\path{arXiv:2204.06125}}.
\newline\urlprefix\url{http://arxiv.org/abs/2204.06125}

\bibitem{nicholGLIDEPhotorealisticImage2022}
A.~Nichol, P.~Dhariwal, A.~Ramesh, P.~Shyam, P.~Mishkin, B.~McGrew, I.~Sutskever, M.~Chen, {{GLIDE}}: {{Towards Photorealistic Image Generation}} and {{Editing}} with {{Text-Guided Diffusion Models}} (Mar. 2022).
\newblock \href {http://arxiv.org/abs/2112.10741} {\path{arXiv:2112.10741}}, \href {https://doi.org/10.48550/arXiv.2112.10741} {\path{doi:10.48550/arXiv.2112.10741}}.

\bibitem{sahariaPhotorealisticTexttoImageDiffusion2022}
C.~Saharia, W.~Chan, S.~Saxena, L.~Li, J.~Whang, E.~Denton, S.~K.~S. Ghasemipour, B.~K. Ayan, S.~S. Mahdavi, R.~G. Lopes, T.~Salimans, J.~Ho, D.~J. Fleet, M.~Norouzi, Photorealistic {{Text-to-Image Diffusion Models}} with {{Deep Language Understanding}} (May 2022).
\newblock \href {http://arxiv.org/abs/2205.11487} {\path{arXiv:2205.11487}}, \href {https://doi.org/10.48550/arXiv.2205.11487} {\path{doi:10.48550/arXiv.2205.11487}}.

\bibitem{podellSDXLImprovingLatent2023}
D.~Podell, Z.~English, K.~Lacey, A.~Blattmann, T.~Dockhorn, J.~M{\"u}ller, J.~Penna, R.~Rombach, {{SDXL}}: {{Improving Latent Diffusion Models}} for {{High-Resolution Image Synthesis}} (Jul. 2023).
\newblock \href {http://arxiv.org/abs/2307.01952} {\path{arXiv:2307.01952}}, \href {https://doi.org/10.48550/arXiv.2307.01952} {\path{doi:10.48550/arXiv.2307.01952}}.

\bibitem{mouT2IAdapterLearningAdapters2023}
C.~Mou, X.~Wang, L.~Xie, Y.~Wu, J.~Zhang, Z.~Qi, Y.~Shan, X.~Qie, {{T2I-Adapter}}: {{Learning Adapters}} to {{Dig}} out {{More Controllable Ability}} for {{Text-to-Image Diffusion Models}} (Mar. 2023).
\newblock \href {http://arxiv.org/abs/2302.08453} {\path{arXiv:2302.08453}}, \href {https://doi.org/10.48550/arXiv.2302.08453} {\path{doi:10.48550/arXiv.2302.08453}}.

\bibitem{yeIPAdapterTextCompatible2023}
H.~Ye, J.~Zhang, S.~Liu, X.~Han, W.~Yang, {{IP-Adapter}}: {{Text Compatible Image Prompt Adapter}} for {{Text-to-Image Diffusion Models}} (Aug. 2023).
\newblock \href {http://arxiv.org/abs/2308.06721} {\path{arXiv:2308.06721}}, \href {https://doi.org/10.48550/arXiv.2308.06721} {\path{doi:10.48550/arXiv.2308.06721}}.

\bibitem{galImageWorthOne2022}
R.~Gal, Y.~Alaluf, Y.~Atzmon, O.~Patashnik, A.~H. Bermano, G.~Chechik, D.~{Cohen-Or}, An {{Image}} is {{Worth One Word}}: {{Personalizing Text-to-Image Generation}} using {{Textual Inversion}} (Aug. 2022).
\newblock \href {http://arxiv.org/abs/2208.01618} {\path{arXiv:2208.01618}}, \href {https://doi.org/10.48550/arXiv.2208.01618} {\path{doi:10.48550/arXiv.2208.01618}}.

\bibitem{ruizDreamBoothFineTuning2023}
N.~Ruiz, Y.~Li, V.~Jampani, Y.~Pritch, M.~Rubinstein, K.~Aberman, {{DreamBooth}}: {{Fine Tuning Text-to-Image Diffusion Models}} for {{Subject-Driven Generation}} (Mar. 2023).
\newblock \href {http://arxiv.org/abs/2208.12242} {\path{arXiv:2208.12242}}, \href {https://doi.org/10.48550/arXiv.2208.12242} {\path{doi:10.48550/arXiv.2208.12242}}.

\bibitem{openaiGPT4TechnicalReport2023}
OpenAI, {{GPT-4 Technical Report}} (Mar. 2023).
\newblock \href {http://arxiv.org/abs/2303.08774} {\path{arXiv:2303.08774}}, \href {https://doi.org/10.48550/arXiv.2303.08774} {\path{doi:10.48550/arXiv.2303.08774}}.

\bibitem{openaiGPT4o2024}
OpenAI, \href{https://openai.com/index/hello-gpt-4o/}{{{GPT-4o}}} (2024).
\newline\urlprefix\url{https://openai.com/index/hello-gpt-4o/}

\bibitem{liuGroundingDINOMarrying2023}
S.~Liu, Z.~Zeng, T.~Ren, F.~Li, H.~Zhang, J.~Yang, C.~Li, J.~Yang, H.~Su, J.~Zhu, L.~Zhang, Grounding {{DINO}}: {{Marrying DINO}} with {{Grounded Pre-Training}} for {{Open-Set Object Detection}} (Mar. 2023).
\newblock \href {http://arxiv.org/abs/2303.05499} {\path{arXiv:2303.05499}}, \href {https://doi.org/10.48550/arXiv.2303.05499} {\path{doi:10.48550/arXiv.2303.05499}}.

\bibitem{redmonYouOnlyLook2016}
J.~Redmon, S.~Divvala, R.~Girshick, A.~Farhadi, \href{http://arxiv.org/abs/1506.02640}{You {{Only Look Once}}: {{Unified}}, {{Real-Time Object Detection}}} (May 2016).
\newblock \href {http://arxiv.org/abs/1506.02640} {\path{arXiv:1506.02640}}.
\newline\urlprefix\url{http://arxiv.org/abs/1506.02640}

\bibitem{redmonYOLO9000BetterFaster2016}
J.~Redmon, A.~Farhadi, {{YOLO9000}}: {{Better}}, {{Faster}}, {{Stronger}} (Dec. 2016).
\newblock \href {http://arxiv.org/abs/1612.08242} {\path{arXiv:1612.08242}}, \href {https://doi.org/10.48550/arXiv.1612.08242} {\path{doi:10.48550/arXiv.1612.08242}}.

\bibitem{redmonYOLOv3IncrementalImprovement2018}
J.~Redmon, A.~Farhadi, {{YOLOv3}}: {{An Incremental Improvement}} (Apr. 2018).
\newblock \href {http://arxiv.org/abs/1804.02767} {\path{arXiv:1804.02767}}, \href {https://doi.org/10.48550/arXiv.1804.02767} {\path{doi:10.48550/arXiv.1804.02767}}.

\bibitem{bochkovskiyYOLOv4OptimalSpeed2020}
A.~Bochkovskiy, C.-Y. Wang, H.-Y.~M. Liao, {{YOLOv4}}: {{Optimal Speed}} and {{Accuracy}} of {{Object Detection}} (Apr. 2020).
\newblock \href {http://arxiv.org/abs/2004.10934} {\path{arXiv:2004.10934}}, \href {https://doi.org/10.48550/arXiv.2004.10934} {\path{doi:10.48550/arXiv.2004.10934}}.

\bibitem{YOLOv5Ultralytics2020}
\href{https://github.com/ultralytics/yolov5}{{{YOLOv5}} by {{Ultralytics}}} (May 2020).
\newline\urlprefix\url{https://github.com/ultralytics/yolov5}

\bibitem{liYOLOv6SingleStageObject2022}
C.~Li, L.~Li, H.~Jiang, K.~Weng, Y.~Geng, L.~Li, Z.~Ke, Q.~Li, M.~Cheng, W.~Nie, Y.~Li, B.~Zhang, Y.~Liang, L.~Zhou, X.~Xu, X.~Chu, X.~Wei, X.~Wei, \href{http://arxiv.org/abs/2209.02976}{{{YOLOv6}}: {{A Single-Stage Object Detection Framework}} for {{Industrial Applications}}} (Sep. 2022).
\newblock \href {http://arxiv.org/abs/2209.02976} {\path{arXiv:2209.02976}}.
\newline\urlprefix\url{http://arxiv.org/abs/2209.02976}

\bibitem{wangYOLOv7TrainableBagoffreebies2022}
C.-Y. Wang, A.~Bochkovskiy, H.-Y.~M. Liao, {{YOLOv7}}: {{Trainable}} bag-of-freebies sets new state-of-the-art for real-time object detectors (Jul. 2022).
\newblock \href {http://arxiv.org/abs/2207.02696} {\path{arXiv:2207.02696}}, \href {https://doi.org/10.48550/arXiv.2207.02696} {\path{doi:10.48550/arXiv.2207.02696}}.

\bibitem{liYOLOv6V3FullScale2023}
C.~Li, L.~Li, Y.~Geng, H.~Jiang, M.~Cheng, B.~Zhang, Z.~Ke, X.~Xu, X.~Chu, {{YOLOv6}} v3.0: {{A Full-Scale Reloading}} (Jan. 2023).
\newblock \href {http://arxiv.org/abs/2301.05586} {\path{arXiv:2301.05586}}, \href {https://doi.org/10.48550/arXiv.2301.05586} {\path{doi:10.48550/arXiv.2301.05586}}.

\bibitem{jocherUltralyticsYOLOv82023}
G.~Jocher, \href{https://github.com/ultralytics/ultralytics}{Ultralytics {{YOLOv8}}} (2023).
\newline\urlprefix\url{https://github.com/ultralytics/ultralytics}

\bibitem{wangYOLOv9LearningWhat2024}
C.-Y. Wang, I.-H. Yeh, H.-Y.~M. Liao, {{YOLOv9}}: {{Learning What You Want}} to {{Learn Using Programmable Gradient Information}} (Feb. 2024).
\newblock \href {http://arxiv.org/abs/2402.13616} {\path{arXiv:2402.13616}}, \href {https://doi.org/10.48550/arXiv.2402.13616} {\path{doi:10.48550/arXiv.2402.13616}}.

\bibitem{wangYOLOv10RealTimeEndtoEnd2024}
A.~Wang, H.~Chen, L.~Liu, K.~Chen, Z.~Lin, J.~Han, G.~Ding, {{YOLOv10}}: {{Real-Time End-to-End Object Detection}} (May 2024).
\newblock \href {http://arxiv.org/abs/2405.14458} {\path{arXiv:2405.14458}}, \href {https://doi.org/10.48550/arXiv.2405.14458} {\path{doi:10.48550/arXiv.2405.14458}}.

\bibitem{xuMultiscaleObjectDetection2024}
Z.~Xu, N.~Jha, S.~Mehadi, M.~Mandal, Multiscale object detection on complex architectural floor plans, Automation in Construction 165 (2024) 105486.
\newblock \href {https://doi.org/10.1016/j.autcon.2024.105486} {\path{doi:10.1016/j.autcon.2024.105486}}.

\bibitem{dongELNetEfficientLightweight2024}
C.~Dong, X.~Jiang, Y.~Hu, Y.~Du, L.~Pan, {{EL-Net}}: {{An}} efficient and lightweight optimized network for object detection in remote sensing images, Expert Systems with Applications 255 (2024) 124661.
\newblock \href {https://doi.org/10.1016/j.eswa.2024.124661} {\path{doi:10.1016/j.eswa.2024.124661}}.

\bibitem{gaoPETransformerPathEnhanced2024}
J.~Gao, Y.~Zhang, X.~Geng, H.~Tang, U.~A. Bhatti, {{PE-Transformer}}: {{Path}} enhanced transformer for improving underwater object detection, Expert Systems with Applications 246 (2024) 123253.
\newblock \href {https://doi.org/10.1016/j.eswa.2024.123253} {\path{doi:10.1016/j.eswa.2024.123253}}.

\bibitem{choDetectionMovingObjects2023}
J.~Cho, K.~Kim, Detection of moving objects in multi-complex environments using selective attention networks ({{SANet}}), Automation in Construction 155 (2023) 105066.
\newblock \href {https://doi.org/10.1016/j.autcon.2023.105066} {\path{doi:10.1016/j.autcon.2023.105066}}.

\bibitem{guptaLVISDatasetLarge2019}
A.~Gupta, P.~Doll{\'a}r, R.~Girshick, {{LVIS}}: {{A Dataset}} for {{Large Vocabulary Instance Segmentation}} (Sep. 2019).
\newblock \href {http://arxiv.org/abs/1908.03195} {\path{arXiv:1908.03195}}, \href {https://doi.org/10.48550/arXiv.1908.03195} {\path{doi:10.48550/arXiv.1908.03195}}.

\bibitem{chenHowFarAre2024}
Z.~Chen, W.~Wang, H.~Tian, S.~Ye, Z.~Gao, E.~Cui, W.~Tong, K.~Hu, J.~Luo, Z.~Ma, J.~Ma, J.~Wang, X.~Dong, H.~Yan, H.~Guo, C.~He, B.~Shi, Z.~Jin, C.~Xu, B.~Wang, X.~Wei, W.~Li, W.~Zhang, B.~Zhang, P.~Cai, L.~Wen, X.~Yan, M.~Dou, L.~Lu, X.~Zhu, T.~Lu, D.~Lin, Y.~Qiao, J.~Dai, W.~Wang, How {{Far Are We}} to {{GPT-4V}}? {{Closing}} the {{Gap}} to {{Commercial Multimodal Models}} with {{Open-Source Suites}} (Apr. 2024).
\newblock \href {http://arxiv.org/abs/2404.16821} {\path{arXiv:2404.16821}}, \href {https://doi.org/10.48550/arXiv.2404.16821} {\path{doi:10.48550/arXiv.2404.16821}}.

\bibitem{wangInstantIDZeroshotIdentityPreserving2024}
Q.~Wang, X.~Bai, H.~Wang, Z.~Qin, A.~Chen, H.~Li, X.~Tang, Y.~Hu, \href{http://arxiv.org/abs/2401.07519}{{{InstantID}}: {{Zero-shot Identity-Preserving Generation}} in {{Seconds}}} (Feb. 2024).
\newblock \href {http://arxiv.org/abs/2401.07519} {\path{arXiv:2401.07519}}.
\newline\urlprefix\url{http://arxiv.org/abs/2401.07519}

\bibitem{guOpenvocabularyObjectDetection2022}
X.~Gu, T.-Y. Lin, W.~Kuo, Y.~Cui, Open-vocabulary {{Object Detection}} via {{Vision}} and {{Language Knowledge Distillation}} (May 2022).
\newblock \href {http://arxiv.org/abs/2104.13921} {\path{arXiv:2104.13921}}, \href {https://doi.org/10.48550/arXiv.2104.13921} {\path{doi:10.48550/arXiv.2104.13921}}.

\bibitem{radfordLearningTransferableVisual2021}
A.~Radford, J.~W. Kim, C.~Hallacy, A.~Ramesh, G.~Goh, S.~Agarwal, G.~Sastry, A.~Askell, P.~Mishkin, J.~Clark, G.~Krueger, I.~Sutskever, Learning {{Transferable Visual Models From Natural Language Supervision}}, arXiv:2103.00020 [cs] (Feb. 2021).
\newblock \href {http://arxiv.org/abs/2103.00020} {\path{arXiv:2103.00020}}, \href {https://doi.org/10.48550/arxiv.2103.00020} {\path{doi:10.48550/arxiv.2103.00020}}.

\bibitem{renFasterRCNNRealTime2016}
S.~Ren, K.~He, R.~Girshick, J.~Sun, Faster {{R-CNN}}: {{Towards Real-Time Object Detection}} with {{Region Proposal Networks}} (Jan. 2016).
\newblock \href {http://arxiv.org/abs/1506.01497} {\path{arXiv:1506.01497}}, \href {https://doi.org/10.48550/arXiv.1506.01497} {\path{doi:10.48550/arXiv.1506.01497}}.

\bibitem{liGroundedLanguageImagePretraining2022}
L.~H. Li, P.~Zhang, H.~Zhang, J.~Yang, C.~Li, Y.~Zhong, L.~Wang, L.~Yuan, L.~Zhang, J.-N. Hwang, K.-W. Chang, J.~Gao, Grounded {{Language-Image Pre-training}} (Jun. 2022).
\newblock \href {http://arxiv.org/abs/2112.03857} {\path{arXiv:2112.03857}}, \href {https://doi.org/10.48550/arXiv.2112.03857} {\path{doi:10.48550/arXiv.2112.03857}}.

\bibitem{chengYOLOWorldRealTimeOpenVocabulary2024}
T.~Cheng, L.~Song, Y.~Ge, W.~Liu, X.~Wang, Y.~Shan, {{YOLO-World}}: {{Real-Time Open-Vocabulary Object Detection}} (Feb. 2024).
\newblock \href {http://arxiv.org/abs/2401.17270} {\path{arXiv:2401.17270}}, \href {https://doi.org/10.48550/arXiv.2401.17270} {\path{doi:10.48550/arXiv.2401.17270}}.

\bibitem{anthropicClaudeFamily2024}
Anthropic, \href{https://www.anthropic.com/news/claude-3-family}{Claude 3 family} (2024).
\newline\urlprefix\url{https://www.anthropic.com/news/claude-3-family}

\bibitem{googleGeminiFamilyHighly2024}
G.~T. Google, \href{http://arxiv.org/abs/2312.11805}{Gemini: {{A Family}} of {{Highly Capable Multimodal Models}}} (Jun. 2024).
\newblock \href {http://arxiv.org/abs/2312.11805} {\path{arXiv:2312.11805}}.
\newline\urlprefix\url{http://arxiv.org/abs/2312.11805}

\bibitem{alayracFlamingoVisualLanguage2022}
J.-B. Alayrac, J.~Donahue, P.~Luc, A.~Miech, I.~Barr, Y.~Hasson, K.~Lenc, A.~Mensch, K.~Millican, M.~Reynolds, R.~Ring, E.~Rutherford, S.~Cabi, T.~Han, Z.~Gong, S.~Samangooei, M.~Monteiro, J.~Menick, S.~Borgeaud, A.~Brock, A.~Nematzadeh, S.~Sharifzadeh, M.~Binkowski, R.~Barreira, O.~Vinyals, A.~Zisserman, K.~Simonyan, Flamingo: A {{Visual Language Model}} for {{Few-Shot Learning}} (Nov. 2022).
\newblock \href {http://arxiv.org/abs/2204.14198} {\path{arXiv:2204.14198}}, \href {https://doi.org/10.48550/arXiv.2204.14198} {\path{doi:10.48550/arXiv.2204.14198}}.

\bibitem{liuVisualInstructionTuning2023}
H.~Liu, C.~Li, Q.~Wu, Y.~J. Lee, Visual {{Instruction Tuning}} (Dec. 2023).
\newblock \href {http://arxiv.org/abs/2304.08485} {\path{arXiv:2304.08485}}, \href {https://doi.org/10.48550/arXiv.2304.08485} {\path{doi:10.48550/arXiv.2304.08485}}.

\bibitem{liuImprovedBaselinesVisual2024}
H.~Liu, C.~Li, Y.~Li, Y.~J. Lee, Improved {{Baselines}} with {{Visual Instruction Tuning}} (May 2024).
\newblock \href {http://arxiv.org/abs/2310.03744} {\path{arXiv:2310.03744}}, \href {https://doi.org/10.48550/arXiv.2310.03744} {\path{doi:10.48550/arXiv.2310.03744}}.

\bibitem{xuLLaVAUHDLMMPerceiving2024}
R.~Xu, Y.~Yao, Z.~Guo, J.~Cui, Z.~Ni, C.~Ge, T.-S. Chua, Z.~Liu, M.~Sun, G.~Huang, {{LLaVA-UHD}}: An {{LMM Perceiving Any Aspect Ratio}} and {{High-Resolution Images}} (Mar. 2024).
\newblock \href {http://arxiv.org/abs/2403.11703} {\path{arXiv:2403.11703}}, \href {https://doi.org/10.48550/arXiv.2403.11703} {\path{doi:10.48550/arXiv.2403.11703}}.

\bibitem{zhuMiniGPT4EnhancingVisionLanguage2023}
D.~Zhu, J.~Chen, X.~Shen, X.~Li, M.~Elhoseiny, {{MiniGPT-4}}: {{Enhancing Vision-Language Understanding}} with {{Advanced Large Language Models}} (Apr. 2023).
\newblock \href {http://arxiv.org/abs/2304.10592} {\path{arXiv:2304.10592}}, \href {https://doi.org/10.48550/arXiv.2304.10592} {\path{doi:10.48550/arXiv.2304.10592}}.

\bibitem{chenInternVLScalingVision2024}
Z.~Chen, J.~Wu, W.~Wang, W.~Su, G.~Chen, S.~Xing, M.~Zhong, Q.~Zhang, X.~Zhu, L.~Lu, B.~Li, P.~Luo, T.~Lu, Y.~Qiao, J.~Dai, {{InternVL}}: {{Scaling}} up {{Vision Foundation Models}} and {{Aligning}} for {{Generic Visual-Linguistic Tasks}} (Jan. 2024).
\newblock \href {http://arxiv.org/abs/2312.14238} {\path{arXiv:2312.14238}}, \href {https://doi.org/10.48550/arXiv.2312.14238} {\path{doi:10.48550/arXiv.2312.14238}}.

\bibitem{liLLaVAMedTrainingLarge2023}
C.~Li, C.~Wong, S.~Zhang, N.~Usuyama, H.~Liu, J.~Yang, T.~Naumann, H.~Poon, J.~Gao, {{LLaVA-Med}}: {{Training}} a {{Large Language-and-Vision Assistant}} for {{Biomedicine}} in {{One Day}} (Jun. 2023).
\newblock \href {http://arxiv.org/abs/2306.00890} {\path{arXiv:2306.00890}}, \href {https://doi.org/10.48550/arXiv.2306.00890} {\path{doi:10.48550/arXiv.2306.00890}}.

\bibitem{leeNERIFGPT4vAutomatic2023}
G.-G. Lee, X.~Zhai, {{NERIF}}: {{GPT-4V}} for {{Automatic Scoring}} of {{Drawn Models}} (Dec. 2023).
\newblock \href {http://arxiv.org/abs/2311.12990} {\path{arXiv:2311.12990}}, \href {https://doi.org/10.48550/arXiv.2311.12990} {\path{doi:10.48550/arXiv.2311.12990}}.

\bibitem{yangSetofMarkPromptingUnleashes2023}
J.~Yang, H.~Zhang, F.~Li, X.~Zou, C.~Li, J.~Gao, Set-of-{{Mark Prompting Unleashes Extraordinary Visual Grounding}} in {{GPT-4V}} (Nov. 2023).
\newblock \href {http://arxiv.org/abs/2310.11441} {\path{arXiv:2310.11441}}, \href {https://doi.org/10.48550/arXiv.2310.11441} {\path{doi:10.48550/arXiv.2310.11441}}.

\bibitem{yangDawnLMMsPreliminary2023}
Z.~Yang, L.~Li, K.~Lin, J.~Wang, C.-C. Lin, Z.~Liu, L.~Wang, The {{Dawn}} of {{LMMs}}: {{Preliminary Explorations}} with {{GPT-4V}}(ision) (Sep. 2023).
\newblock \href {http://arxiv.org/abs/2309.17421} {\path{arXiv:2309.17421}}, \href {https://doi.org/10.48550/arXiv.2309.17421} {\path{doi:10.48550/arXiv.2309.17421}}.

\bibitem{wangCSPNetNewBackbone2019}
C.-Y. Wang, H.-Y.~M. Liao, I.-H. Yeh, Y.-H. Wu, P.-Y. Chen, J.-W. Hsieh, {{CSPNet}}: {{A New Backbone}} that can {{Enhance Learning Capability}} of {{CNN}} (Nov. 2019).
\newblock \href {http://arxiv.org/abs/1911.11929} {\path{arXiv:1911.11929}}, \href {https://doi.org/10.48550/arXiv.1911.11929} {\path{doi:10.48550/arXiv.1911.11929}}.

\bibitem{fengTOODTaskalignedOnestage2021}
C.~Feng, Y.~Zhong, Y.~Gao, M.~R. Scott, W.~Huang, {{TOOD}}: {{Task-aligned One-stage Object Detection}} (Aug. 2021).
\newblock \href {http://arxiv.org/abs/2108.07755} {\path{arXiv:2108.07755}}, \href {https://doi.org/10.48550/arXiv.2108.07755} {\path{doi:10.48550/arXiv.2108.07755}}.

\bibitem{zhengDistanceIoULossFaster2019}
Z.~Zheng, P.~Wang, W.~Liu, J.~Li, R.~Ye, D.~Ren, Distance-{{IoU Loss}}: {{Faster}} and {{Better Learning}} for {{Bounding Box Regression}} (Nov. 2019).
\newblock \href {http://arxiv.org/abs/1911.08287} {\path{arXiv:1911.08287}}, \href {https://doi.org/10.48550/arXiv.1911.08287} {\path{doi:10.48550/arXiv.1911.08287}}.

\bibitem{gevorgyanSIoULossMore2022}
Z.~Gevorgyan, \href{http://arxiv.org/abs/2205.12740}{{{SIoU Loss}}: {{More Powerful Learning}} for {{Bounding Box Regression}}} (May 2022).
\newblock \href {http://arxiv.org/abs/2205.12740} {\path{arXiv:2205.12740}}.
\newline\urlprefix\url{http://arxiv.org/abs/2205.12740}

\bibitem{rezatofighiGeneralizedIntersectionUnion2019}
H.~Rezatofighi, N.~Tsoi, J.~Gwak, A.~Sadeghian, I.~Reid, S.~Savarese, Generalized {{Intersection}} over {{Union}}: {{A Metric}} and {{A Loss}} for {{Bounding Box Regression}} (Apr. 2019).
\newblock \href {http://arxiv.org/abs/1902.09630} {\path{arXiv:1902.09630}}, \href {https://doi.org/10.48550/arXiv.1902.09630} {\path{doi:10.48550/arXiv.1902.09630}}.

\bibitem{dingRepVGGMakingVGGstyle2021}
X.~Ding, X.~Zhang, N.~Ma, J.~Han, G.~Ding, J.~Sun, {{RepVGG}}: {{Making VGG-style ConvNets Great Again}} (Mar. 2021).
\newblock \href {http://arxiv.org/abs/2101.03697} {\path{arXiv:2101.03697}}, \href {https://doi.org/10.48550/arXiv.2101.03697} {\path{doi:10.48550/arXiv.2101.03697}}.

\bibitem{wangDesigningNetworkDesign2022}
C.-Y. Wang, H.-Y.~M. Liao, I.-H. Yeh, \href{http://arxiv.org/abs/2211.04800}{Designing {{Network Design Strategies Through Gradient Path Analysis}}} (Nov. 2022).
\newblock \href {http://arxiv.org/abs/2211.04800} {\path{arXiv:2211.04800}}.
\newline\urlprefix\url{http://arxiv.org/abs/2211.04800}

\bibitem{klingerPHashOrgHome2008}
E.~Klinger, D.~Starkweather, \href{http://www.phash.org/}{{{pHash}}.org: {{Home}} of {{pHash}}, the open source perceptual hash library} (2008).
\newline\urlprefix\url{http://www.phash.org/}

\bibitem{ronnebergerUNetConvolutionalNetworks2015}
O.~Ronneberger, P.~Fischer, T.~Brox, U-{{Net}}: {{Convolutional Networks}} for {{Biomedical Image Segmentation}} (May 2015).
\newblock \href {http://arxiv.org/abs/1505.04597} {\path{arXiv:1505.04597}}, \href {https://doi.org/10.48550/arXiv.1505.04597} {\path{doi:10.48550/arXiv.1505.04597}}.

\bibitem{huLoRALowRankAdaptation2021}
E.~J. Hu, Y.~Shen, P.~Wallis, Z.~{Allen-Zhu}, Y.~Li, S.~Wang, L.~Wang, W.~Chen, {{LoRA}}: {{Low-Rank Adaptation}} of {{Large Language Models}} (Oct. 2021).
\newblock \href {http://arxiv.org/abs/2106.09685} {\path{arXiv:2106.09685}}, \href {https://doi.org/10.48550/arXiv.2106.09685} {\path{doi:10.48550/arXiv.2106.09685}}.

\bibitem{hangEfficientDiffusionTraining2024}
T.~Hang, S.~Gu, C.~Li, J.~Bao, D.~Chen, H.~Hu, X.~Geng, B.~Guo, Efficient {{Diffusion Training}} via {{Min-SNR Weighting Strategy}} (Mar. 2024).
\newblock \href {http://arxiv.org/abs/2303.09556} {\path{arXiv:2303.09556}}, \href {https://doi.org/10.48550/arXiv.2303.09556} {\path{doi:10.48550/arXiv.2303.09556}}.

\bibitem{vaswaniAttentionAllYou2017}
A.~Vaswani, N.~Shazeer, N.~Parmar, J.~Uszkoreit, L.~Jones, A.~N. Gomez, {\L}.~Kaiser, I.~Polosukhin, Attention is all you need, Advances in neural information processing systems 30 (2017).
\newblock \href {https://doi.org/10.48550/arxiv.1706.03762} {\path{doi:10.48550/arxiv.1706.03762}}.

\bibitem{dosovitskiyImageWorth16x162021}
A.~Dosovitskiy, L.~Beyer, A.~Kolesnikov, D.~Weissenborn, X.~Zhai, T.~Unterthiner, M.~Dehghani, M.~Minderer, G.~Heigold, S.~Gelly, J.~Uszkoreit, N.~Houlsby, An {{Image}} is {{Worth}} 16x16 {{Words}}: {{Transformers}} for {{Image Recognition}} at {{Scale}}, arXiv:2010.11929 [cs] (Jun. 2021).
\newblock \href {http://arxiv.org/abs/2010.11929} {\path{arXiv:2010.11929}}, \href {https://doi.org/10.48550/arxiv.2010.11929} {\path{doi:10.48550/arxiv.2010.11929}}.

\bibitem{weiChainofThoughtPromptingElicits2023}
J.~Wei, X.~Wang, D.~Schuurmans, M.~Bosma, B.~Ichter, F.~Xia, E.~Chi, Q.~Le, D.~Zhou, Chain-of-{{Thought Prompting Elicits Reasoning}} in {{Large Language Models}} (Jan. 2023).
\newblock \href {http://arxiv.org/abs/2201.11903} {\path{arXiv:2201.11903}}, \href {https://doi.org/10.48550/arXiv.2201.11903} {\path{doi:10.48550/arXiv.2201.11903}}.

\bibitem{heSpatialPyramidPooling2014}
K.~He, X.~Zhang, S.~Ren, J.~Sun, Spatial {{Pyramid Pooling}} in {{Deep Convolutional Networks}} for {{Visual Recognition}} (2014).
\newblock \href {http://arxiv.org/abs/1406.4729} {\path{arXiv:1406.4729}}, \href {https://doi.org/10.1007/978-3-319-10578-9_23} {\path{doi:10.1007/978-3-319-10578-9_23}}.

\bibitem{liuPathAggregationNetwork2018}
S.~Liu, L.~Qi, H.~Qin, J.~Shi, J.~Jia, Path {{Aggregation Network}} for {{Instance Segmentation}} (Sep. 2018).
\newblock \href {http://arxiv.org/abs/1803.01534} {\path{arXiv:1803.01534}}, \href {https://doi.org/10.48550/arXiv.1803.01534} {\path{doi:10.48550/arXiv.1803.01534}}.

\bibitem{liGeneralizedFocalLoss2020a}
X.~Li, W.~Wang, L.~Wu, S.~Chen, X.~Hu, J.~Li, J.~Tang, J.~Yang, Generalized {{Focal Loss}}: {{Learning Qualified}} and {{Distributed Bounding Boxes}} for {{Dense Object Detection}} (Jun. 2020).
\newblock \href {http://arxiv.org/abs/2006.04388} {\path{arXiv:2006.04388}}, \href {https://doi.org/10.48550/arXiv.2006.04388} {\path{doi:10.48550/arXiv.2006.04388}}.

\bibitem{rangekingBriefSummaryYOLOv82023}
RangeKing, \href{https://github.com/ultralytics/ultralytics/issues/189}{Brief summary of {{YOLOv8}} model structure} (2023).
\newline\urlprefix\url{https://github.com/ultralytics/ultralytics/issues/189}

\bibitem{linMicrosoftCOCOCommon2015}
T.-Y. Lin, M.~Maire, S.~Belongie, L.~Bourdev, R.~Girshick, J.~Hays, P.~Perona, D.~Ramanan, C.~L. Zitnick, P.~Doll{\'a}r, Microsoft {{COCO}}: {{Common Objects}} in {{Context}} (Feb. 2015).
\newblock \href {http://arxiv.org/abs/1405.0312} {\path{arXiv:1405.0312}}, \href {https://doi.org/10.48550/arXiv.1405.0312} {\path{doi:10.48550/arXiv.1405.0312}}.

\bibitem{loshchilovDecoupledWeightDecay2017}
I.~Loshchilov, F.~Hutter, Decoupled weight decay regularization, arXiv preprint arXiv:1711.05101 (2017).
\newblock \href {http://arxiv.org/abs/1711.05101} {\path{arXiv:1711.05101}}, \href {https://doi.org/10.48550/arxiv.1711.05101} {\path{doi:10.48550/arxiv.1711.05101}}.

\bibitem{reimersSentenceBERTSentenceEmbeddings2019}
N.~Reimers, I.~Gurevych, Sentence-{{BERT}}: {{Sentence Embeddings}} using {{Siamese BERT-Networks}} (Aug. 2019).
\newblock \href {http://arxiv.org/abs/1908.10084} {\path{arXiv:1908.10084}}, \href {https://doi.org/10.48550/arXiv.1908.10084} {\path{doi:10.48550/arXiv.1908.10084}}.

\bibitem{liuSwinTransformerHierarchical2021}
Z.~Liu, Y.~Lin, Y.~Cao, H.~Hu, Y.~Wei, Z.~Zhang, S.~Lin, B.~Guo, Swin {{Transformer}}: {{Hierarchical Vision Transformer}} using {{Shifted Windows}} (Aug. 2021).
\newblock \href {http://arxiv.org/abs/2103.14030} {\path{arXiv:2103.14030}}, \href {https://doi.org/10.48550/arXiv.2103.14030} {\path{doi:10.48550/arXiv.2103.14030}}.

\bibitem{zongDETRsCollaborativeHybrid2023}
Z.~Zong, G.~Song, Y.~Liu, \href{http://arxiv.org/abs/2211.12860}{{{DETRs}} with {{Collaborative Hybrid Assignments Training}}} (Aug. 2023).
\newblock \href {http://arxiv.org/abs/2211.12860} {\path{arXiv:2211.12860}}.
\newline\urlprefix\url{http://arxiv.org/abs/2211.12860}

\bibitem{wangInternImageExploringLargeScale2023}
W.~Wang, J.~Dai, Z.~Chen, Z.~Huang, Z.~Li, X.~Zhu, X.~Hu, T.~Lu, L.~Lu, H.~Li, X.~Wang, Y.~Qiao, {{InternImage}}: {{Exploring Large-Scale Vision Foundation Models}} with {{Deformable Convolutions}} (Apr. 2023).
\newblock \href {http://arxiv.org/abs/2211.05778} {\path{arXiv:2211.05778}}, \href {https://doi.org/10.48550/arXiv.2211.05778} {\path{doi:10.48550/arXiv.2211.05778}}.

\end{thebibliography}

\end{document}